\documentclass[journal,comsoc]{IEEEtran}
\usepackage[T1]{fontenc}

\ifCLASSINFOpdf
\else
\fi
\usepackage{cite}
\usepackage{amsmath,amssymb,amsfonts}
\usepackage{algorithm,algorithmic}
\usepackage{graphicx}
\usepackage{xcolor}
\usepackage{algorithmic}
\usepackage{graphicx}
\usepackage{caption}
\usepackage{amsmath}

\usepackage{tabu}
\usepackage{amsthm}
\usepackage{mathtools}
\usepackage{footnote}

\usepackage{subfigure}

\newtheorem{proposition}{Proposition}

\DeclareMathAlphabet{\mathcal}{OMS}{cmsy}{m}{n}

\DeclareMathOperator*{\argmax}{argmax}

\usepackage[cmintegrals]{newtxmath}
\begin{document}
%
\title{Multi-Agent Meta-Reinforcement Learning for Self-Powered and Sustainable Edge Computing Systems
}
%
%
%


\author{Md.~Shirajum~Munir,~\IEEEmembership{Graduate~Student~Member,~IEEE,}
	Nguyen~H.~Tran,~\IEEEmembership{Senior~Member,~IEEE,}
	Walid~Saad,~\IEEEmembership{Fellow,~IEEE,}
	and~Choong~Seon~Hong,~\IEEEmembership{Senior~Member,~IEEE}
	\thanks{This work was partially supported by the National Research Foundation of Korea(NRF) grant funded by the Korea government(MSIT) (No. 2020R1A4A1018607) and by Institute of Information \& communications Technology Planning \& Evaluation (IITP) grant funded by the Korea government(MSIT) (No.2019-0-01287, Evolvable Deep Learning Model Generation Platform for Edge Computing).}
	\thanks{Md. Shirajum Munir, and Choong Seon Hong are with the Department of Computer Science and Engineering, Kyung Hee University, Yongin-si 17104, Republic of Korea (e-mail: munir@khu.ac.kr; cshong@khu.ac.kr).}
	\thanks{Nguyen H. Tran is with the School of Computer Science, The University of Sydney, Sydney, 2006, NSW, Australia. (e-mail: nguyen.tran@sydney.edu.au).}
	\thanks{Walid Saad  is with the Wireless@VT Group, Bradley Department of Electrical
		and Computer Engineering, Virginia Tech, Blacksburg, VA 24061 USA, and
		also with the Department of Computer Science and Engineering, Kyung Hee
		University, Yongin-si 17104, Republic of Korea (e-mail: walids@vt.edu).}
	\thanks{Corresponding author: Choong Seon Hong (e-mail: cshong@khu.ac.kr).}
   \thanks{\textcopyright 2021 IEEE. Personal use of this material is permitted.  Permission from IEEE must be obtained for all other uses, in any current or future media, including reprinting/republishing this material for advertising or promotional purposes, creating new collective works, for resale or redistribution to servers or lists, or reuse of any copyrighted component of this work in other works.}}

%
%

\markboth{ACCEPTED ARTICLE BY IEEE Transactions on Network and Service Management, DOI: 10.1109/TNSM.2021.3057960}%
{Shell \MakeLowercase{\textit{et al.}}: Bare Demo of IEEEtran.cls for IEEE Communications Society Journals}
%



\maketitle

\begin{abstract}
The stringent requirements of mobile edge computing (MEC) applications and functions fathom the high capacity and dense deployment of MEC hosts to the upcoming wireless networks. However, operating such high capacity MEC hosts can significantly increase energy consumption. Thus, a base station (BS) unit can act as a self-powered BS. In this paper, an effective energy dispatch mechanism for self-powered wireless networks with edge computing capabilities is studied. First, a two-stage linear stochastic programming problem is formulated with the goal of minimizing the total energy consumption cost of the system while fulfilling the energy demand. Second, a semi-distributed data-driven solution is proposed by developing a novel multi-agent meta-reinforcement learning (MAMRL) framework to solve the formulated problem. In particular, each BS plays the role of a local agent that explores a Markovian behavior for both energy consumption and generation while each BS transfers time-varying features to a meta-agent. Sequentially, the meta-agent optimizes (i.e., exploits) the energy dispatch decision by accepting only the observations from each local agent with its own state information. Meanwhile, each BS agent estimates its own energy dispatch policy by applying the learned parameters from meta-agent. Finally, the proposed MAMRL framework is benchmarked by analyzing deterministic, asymmetric, and stochastic environments in terms of non-renewable energy usages, energy cost, and accuracy. Experimental results show that the proposed MAMRL model can reduce up to $11\%$ non-renewable energy usage and by $22.4\%$ the energy cost (with $95.8\%$ prediction accuracy), compared to other baseline methods.
\end{abstract}


%
\IEEEpeerreviewmaketitle

\section{Introduction}
Next-generation wireless networks are expected to significantly rely on \emph{edge} applications and functions that include edge computing and edge artificial intelligence (edge AI) \cite{IEEEhowto:Saad_6G, IEEEhowto:Park_Edge_AI, IEEEhowto:Dahlman_AI_in_Moblile_Com, IEEEhowto:Munir_Edge_AI,IEEEhowto:Chen_AI_Tutorial, IEEEhowto:Chen_Com_C_Framework_FL,IEEEhowto:Tran_FL_Infocom}. To successfully support such edge services within a wireless network with mobile edge computing (MEC) capabilities, energy management (i.e., demand and supply) is one of the most critical design challenges. In particular, it is imperative to equip next-generation wireless networks with alternative energy sources, such as renewable energy, in order to provide extremely reliable energy dispatch with less energy consumption cost \cite{IEEEhowto:Lee_self_powered, IEEEhowto:Wei_A3C_Hybrid_Power, IEEEhowto:Munir_Edge_Microgrid, IEEEhowto:Xu_Offloading_EH, IEEEhowto:Munir_GC_Multi_Agent,IEEEhowto:Piovesan_microgrid_def,IEEEhowto:Huang_Jointly_Optimizing_BS, IEEEhowto:Li_Egde_Renewable}. 
An efficient energy dispatch design requires energy sustainability, which not only saves energy consumption cost, but also fulfills the energy demand of the edge computing by enabling its own renewable energy sources. Specifically, sustainable energy is the practice of seamless energy flow to the MEC system that emerges to meet the energy demand without compromising the ability of future energy generation. Furthermore, to ensure a sustainable MEC operation, the retrogressive penetration of uncertainty for energy consumption and generation is essential. A summary of the challenges that are solved by the literature to enable renewable energy sources for the wireless network is presented in Table \ref{tab:renewable_wireless_network}.

To provide sustainable edge computing for next-generation wireless systems, each base station (BS) with MEC capabilities unit can be equipped with renewable energy sources.  Thus, the energy source of such a BS unit not only relies solely on the power grid, but also on the equipped renewable energy sources. In particular, in a self-powered network, wireless BSs with MEC capabilities is equipped with its own renewable energy sources that can generate renewable energy, consume, store, and share energy with other BS units.

Delivering seamless energy flow with a low energy consumption cost in a self-powered wireless network with MEC capabilities can lead to uncertainty in both energy demand and generation. In particular, the randomness of the energy demand is induced by the uncertain resources (i.e., computation and communication) request by the edge services and applications. Meanwhile, the energy generation of a renewable source (i.e., a solar panel) at each self-powered BS unit varies on the time of a day. In other words, the pattern of energy demand and generation will differ from one self-powered BS unit to another. Thus, such fluctuating energy demand and generation pattern induces a non-independent and identically distributed (non-i.i.d.) of energy dispatch at each BS over time. 
To overcome this non-i.i.d. energy demand and generation, characterizing the expected amount of uncertainty is crucial to ensure a seamless energy flow to the self-powered wireless network.
As such, when designing self-powered wireless networks, it is necessary to take into account this uncertainty in the energy patterns.

\begin{table*}[!t]
	\caption{Summary of the challenges that are solved by the literature for enabling renewable energy sources in the wireless network.}
	\centering
	\begin{tabular}{|p{1.2cm}|p{2.2cm}|p{1.2cm}|p{1.2cm}|p{1.2cm}|p{1.2cm}|p{6.8cm}|}
		\hline
		\textbf{Ref.} & \textbf{Energy sources} & \textbf{MEC capabilities} & \textbf{Non-i.i.d. dataset} & \textbf{Energy dispatch}  & \textbf{Energy cost} &  \textbf{Remarks} \\ 
		\hline
		\cite{IEEEhowto:Lee_self_powered} & Renewable & No & No  & No & No & Activation and deactivation of BSs in a self-powered network\\ 
		\hline
		\cite{IEEEhowto:Wei_A3C_Hybrid_Power} & Hybrid
		energy & No & No & No & No & User scheduling and network
		resource management\\ 
		\hline
		\cite{IEEEhowto:Xu_Offloading_EH} & Hybrid energy & Yes & No & No & No & Load balancing between the centralized cloud and edge server\\ 
		\hline
		\cite{IEEEhowto:Munir_Edge_Microgrid} & Microgrid & Yes & No & Yes & No & MEC task assignment and energy demand-response (DR) management\\ 
		\hline
		\cite{IEEEhowto:Munir_GC_Multi_Agent} & Microgrid & Yes & No & Yes & No & Risk-sensitive energy profiling for microgrid-powered MEC network\\ 
		\hline
		\cite{IEEEhowto:Piovesan_microgrid_def} & Renewable & No & No & Yes & No & Energy load balancing among the SBSs with a microgrid\\ 
		\hline
		\cite{IEEEhowto:Huang_Jointly_Optimizing_BS} & Smart grid enabled hybrid energy & No & No & Yes & No & Joint network resource allocation and energy sharing among the BSs \\ 
		\hline
		\cite{IEEEhowto:Li_Egde_Renewable} & Hybrid
		energy & No & No & Yes & No & Overall system architecture for edge computing and renewable energy resources\\ 
		\hline
		This work & Smart grid enabled self-powered renewable energy & Yes & Yes & Yes & Yes & An effective energy dispatch mechanism for self-powered wireless networks with edge computing capabilities\\ 
		\hline
	\end{tabular}
	\label{tab:renewable_wireless_network}
\end{table*}

\subsection{Related Works}
The problem of energy management for MEC-enabled wireless networks has been studied  in \cite{IEEEhowto:Mao_Stochastic_Joint_Radio_Sim_para, IEEEhowto:Tran_energy_MEC_no_downlink, IEEEhowto:Chang_MEC_Power_Model, IEEEhowto:Sun_EMM, IEEEhowto:Abedin_Green_IoT, IEEEhowto:Zhang_Calibrated_Lr_Power_SBS, IEEEhowto:AKIN_EH_Store_Markov} (summary in Table \ref{tab:Related_Works}). In \cite{IEEEhowto:Mao_Stochastic_Joint_Radio_Sim_para}, the authors proposed a joint mechanism for radio resource management and users task offloading with the goal of minimizing the long-term power consumption for both mobile devices and the MEC server. The authors in \cite{IEEEhowto:Tran_energy_MEC_no_downlink} proposed a heuristic to solve the joint problem of computational resource allocation, uplink transmission power, and user task offloading problem. The work in \cite{IEEEhowto:Chang_MEC_Power_Model} studied the tradeoff between communication and computation for a MEC system and the authors proposed a MEC server CPU scaling mechanism for reducing the energy consumption. Further, the work in \cite{IEEEhowto:Sun_EMM} proposed an energy-aware mobility management scheme for MEC in ultra-dense networks, and they addressed the problem using Lyapunov optimization and multi-armed bandits. Recently, the authors in \cite{IEEEhowto:Zhang_Calibrated_Lr_Power_SBS} proposed a distributed power control scheme for a small cell network by using the concept of a multi-agent calibrate learning. Further, the authors in \cite{IEEEhowto:AKIN_EH_Store_Markov} studied the problem of energy storage and energy harvesting (EH) for a wireless network using deviation theory and Markov processes. However, all of these existing works assume that the consumed energy is available from the energy utility source to the wireless network system \cite{IEEEhowto:Mao_Stochastic_Joint_Radio_Sim_para, IEEEhowto:Tran_energy_MEC_no_downlink, IEEEhowto:Chang_MEC_Power_Model, IEEEhowto:Sun_EMM, IEEEhowto:Abedin_Green_IoT, IEEEhowto:Zhang_Calibrated_Lr_Power_SBS, IEEEhowto:AKIN_EH_Store_Markov}. Since the assumed models are often focused on energy management and user task offloading on network resource allocations, the random demand for computational (e.g., CPU computation, memory, etc.) and communication requirements of the edge applications and services are not considered. In fact, even if enough energy supply is available, the energy cost related to network operation can be significant because of the usage of non-renewable (e.g., coal, petroleum, natural gas). Indeed, it is necessary to include renewable energy sources towards the next-generation wireless networking infrastructure.

\begin{table}[!t]
	\caption{Summary of the related works \cite{IEEEhowto:Mao_Stochastic_Joint_Radio_Sim_para, IEEEhowto:Tran_energy_MEC_no_downlink, IEEEhowto:Chang_MEC_Power_Model, IEEEhowto:Sun_EMM, IEEEhowto:Abedin_Green_IoT, IEEEhowto:Zhang_Calibrated_Lr_Power_SBS, IEEEhowto:AKIN_EH_Store_Markov, IEEEhowto:Incentivizing_Tran, IEEEhowto:Wang_Meta_RL, IEEEhowto:Schweighofera_Meta_RL, IEEEhowto:Andrychowicz_L2L_by_GD, IEEEhowto:Mnih_async_a3c, IEEEhowto:Lowe_Multi_agent_actor_critic}.}
	\centering
	\begin{tabular}{|p{0.6cm}|p{2.5cm}|p{1.5cm}|p{2.2cm}|}
		\hline
		\small\textbf{Ref.} & \small\textbf{Contributions} & \small\textbf{Method} & \small\textbf{Limitation} \\ 
		\hline
		\small \cite{IEEEhowto:Mao_Stochastic_Joint_Radio_Sim_para} & \small Radio resource management and users task offloading  & \small Optimization & \small Usage of non-renewable, deterministic environment   \\ 
		\hline
		\small \cite{IEEEhowto:Tran_energy_MEC_no_downlink} & \small Computational resource allocation, uplink transmission power, and user task offloading & \small Heuristic & \small Usage of non-renewable, energy dispatch, performance guarantee  \\ 
		\hline
		\small \cite{IEEEhowto:Chang_MEC_Power_Model} & \small MEC server CPU scaling mechanism for reducing the energy consumption & \small Optimization & \small Usage of non-renewable, energy dispatch  \\ 
		\hline
		\small \cite{IEEEhowto:Sun_EMM} & \small Energy-aware mobility management scheme for MEC & \small Lyapunov and multi-armed bandits & \small Energy dispatch, i.i.d. energy demand-response  \\ 
		\hline
		\small \cite{IEEEhowto:Abedin_Green_IoT} & \small Energy efficient green-IoT network & \small Heuristic & \small Edge computing, Energy dispatch, deterministic environment  \\ 
		\hline
		\small \cite{IEEEhowto:Zhang_Calibrated_Lr_Power_SBS} &  \small Distributed power control scheme for a small cell network & \small Multi-agent calibrate learning &  \small Usage of non-renewable, energy dispatch  \\ 
		\hline
		\small \cite{IEEEhowto:AKIN_EH_Store_Markov} & \small Energy storage and energy harvesting (EH) for a wireless network  & \small Deviation theory and Markov processes & \small MEC capabilities, i.i.d. energy demand-response  \\ 
		\hline
		\small \cite{IEEEhowto:Incentivizing_Tran} & \small Non-coordinated energy shedding and mis-aligned incentives for mixed-use building & \small Auction theory & \small MEC capabilities, i.i.d. energy demand-response  \\ 
		\hline
		\small \cite{IEEEhowto:Wang_Meta_RL} & \small Tradeoff between effectiveness and available amounts of training data & \small Deep meta-RL & \small Stochastic environment and a multi-agent scenario  \\ 
		\hline
		\small \cite{IEEEhowto:Schweighofera_Meta_RL} & \small Controling the meta-parameter in both static and dynamic environments & \small SGD-based meta-parameter learning & \small Single-agent, same environment  \\ 
		\hline
		\small \cite{IEEEhowto:Andrychowicz_L2L_by_GD} & \small Learning to learn mechanism with the recurrent neural networks & \small Generalized transfer learning & \small Deterministic environment, single-agent  \\ 
		\hline
		\small \cite{IEEEhowto:Mnih_async_a3c} & \small Asynchronous multi-agent RL framework & \small One-step Q-learning, one-step Sarsa, and	n-step Q-learning & \small Deterministic environment  \\ 
		\hline
		\small \cite{IEEEhowto:Lowe_Multi_agent_actor_critic} & \small General-purpose multi-agent scheme & \small Extension of the actor-critic policy gradient  & \small Same environment for all of the local actors  \\ 
		\hline
	\end{tabular}
	\label{tab:Related_Works}
\end{table}

Recently, some of the challenges of renewable energy powered wireless networks have been studied in \cite{IEEEhowto:Lee_self_powered, IEEEhowto:Wei_A3C_Hybrid_Power, IEEEhowto:Munir_Edge_Microgrid, IEEEhowto:Xu_Offloading_EH, IEEEhowto:Munir_GC_Multi_Agent,IEEEhowto:Piovesan_microgrid_def,IEEEhowto:Huang_Jointly_Optimizing_BS, IEEEhowto:Incentivizing_Tran}. In \cite{IEEEhowto:Lee_self_powered}, the authors proposed an online optimization framework to analyze the activation and deactivation of BSs in a self-powered network. In \cite{IEEEhowto:Wei_A3C_Hybrid_Power}, proposed a hybrid power source infrastructure to support heterogeneous networks (HetNets), a model-free deep reinforcement learning (RL) mechanism was proposed for user scheduling and network resource management. In \cite{IEEEhowto:Xu_Offloading_EH}, the authors developed an RL scheme for edge resource management while incorporating renewable energy in the edge network. In particular, the goal of \cite{IEEEhowto:Xu_Offloading_EH} is to minimize a long-term system cost by load balancing between the centralized cloud and edge server. The authors in \cite{IEEEhowto:Munir_Edge_Microgrid} introduced a microgrid enabled edge computing system. A joint optimization problem is studied for MEC task assignment and energy demand-response (DR) management. The authors in \cite{IEEEhowto:Munir_Edge_Microgrid} developed a model-based deep RL framework to tackle the joint problem. In \cite{IEEEhowto:Munir_GC_Multi_Agent}, the authors proposed a risk-sensitive energy profiling for microgrid-powered MEC network to ensure a sustainable energy supply for green edge computing by capturing the conditional value at risk (CVaR) tail distribution of the energy shortfall. The authors in \cite{IEEEhowto:Munir_GC_Multi_Agent} proposed a multi-agent RL system to solve the energy scheduling problem. In \cite{IEEEhowto:Piovesan_microgrid_def}, the authors proposed a  self-sustainable mobile networks, using graph-based approach for intelligent energy management with a microgrid. The authors in \cite{IEEEhowto:Huang_Jointly_Optimizing_BS} proposed a smart grid-enabled wireless network and minimized grid energy consumption by applying energy sharing among the BSs. Furthermore, in \cite{IEEEhowto:Incentivizing_Tran}, the authors addressed challenges of non-coordinated energy shedding and mis-aligned incentives for mixed-use building (i.e., buildings and data centers) using auction theory to reduce energy usage. However, these works \cite{IEEEhowto:Wei_A3C_Hybrid_Power, IEEEhowto:Munir_Edge_Microgrid, IEEEhowto:Xu_Offloading_EH, IEEEhowto:Munir_GC_Multi_Agent,IEEEhowto:Piovesan_microgrid_def,IEEEhowto:Huang_Jointly_Optimizing_BS,IEEEhowto:Incentivizing_Tran} do not investigate the problem of energy dispatch nor do they account for the energy cost of MEC-enabled, self-powered networks when the demand and generation of each self-powered BS are non-i.i.d.. Dealing with non-i.i.d. energy demand and generation among self-powered BSs is challenging due to the intrinsic energy requirements of each BS evolve the uncertainty. In order to overcome this unique \emph{energy dispatch} challenge, we propose to develop a \emph{multi-agent meta-reinforcement learning framework} that can adapt new uncertain environment without considering the entire past experience.

Some interesting problems related to meta-RL and multi-agent deep RL are studied in \cite{IEEEhowto:Wang_Meta_RL, IEEEhowto:Schweighofera_Meta_RL, IEEEhowto:Andrychowicz_L2L_by_GD, IEEEhowto:Mnih_async_a3c, IEEEhowto:Lowe_Multi_agent_actor_critic} (summary in Table \ref{tab:Related_Works}). In \cite{IEEEhowto:Wang_Meta_RL}, the authors focused on studying the challenges of the tradeoff between effectiveness and available amounts of training data for a deep-RL based learning system. To this end, the authors in \cite{IEEEhowto:Wang_Meta_RL} tackled those challenges by exploring a deep meta-reinforcement learning architecture. This learning architecture comprises of two learning systems: 1) lower-level system that can learn each new task very quickly, 2) higher-level system is responsible to improve the performance of each lower-level system task. In particular, this learning mechanism is involved with one lower-level system that can learn relatively quickly as compared with a higher-level system. This lower-level system can adapt to a new task while a higher-level system performs fine-tuning so as to improve the performance of the lower-level system. In particular, in deep meta-reinforcement learning, a lower-level system quantifies a reward based on the desired action and feeds back that reward to a higher-level system to tune the weights of a recurrent network. However, the authors in \cite{IEEEhowto:Wang_Meta_RL} do not consider a stochastic environment nor do they extend their work for a multi-agent scenario. 
The authors in \cite{IEEEhowto:Schweighofera_Meta_RL} proposed a stochastic gradient-based meta-parameter learning scheme for tuning reinforcement learning parameters to the physical environmental dynamics. Particularly, the experiment in \cite{IEEEhowto:Schweighofera_Meta_RL} performed in both animal and robot environments, where an animal must recognize food before it starves and a robot must recharge before the battery is empty. Thus, the proposed scheme can effectively find meta-parameter values and controls the meta-parameter in both static and dynamic environments.  
In \cite{IEEEhowto:Andrychowicz_L2L_by_GD}, the authors investigated a learning to learn (i.e., meta-learning) mechanism with the recurrent neural networks, where the meta-learning problem was designed as a generalized transfer learning scheme. In particular, the authors in \cite{IEEEhowto:Andrychowicz_L2L_by_GD} considered a parametrized optimizer that can transfer the neural network parameters update to an optimizee. Meanwhile, the optimizee can determine the gradients without relying on the optimizer parameters. Moreover, the optimizee sends the error to the optimizer, and updates its own parameters based on the transferred parameters. This mechanism allows an agent to learn new tasks for a similar structure.
An asynchronous multi-agent RL framework was studied in  \cite{IEEEhowto:Mnih_async_a3c}, where the authors investigated how parallel actor learners of asynchronous advantage actor-critic (A3C) can achieve better stability during the neural network training comparted to asynchronous RL schemes. Such schemes include asynchronous one-step Q-learning, one-step Sarsa, and n-step Q-learning.
The authors in \cite{IEEEhowto:Lowe_Multi_agent_actor_critic} proposed a general-purpose multi-agent scheme by adopting the framework of centralized training with decentralized execution. In particular, the authors in \cite{IEEEhowto:Lowe_Multi_agent_actor_critic} proposed an extension of the actor-critic policy gradient mechanism by modifying the role of the critic. This critic is augmented with an additional policy information from the other actors (agents). Sequentially, each local actor executes in a decentralized manner and sends its own policy to the centralized critic for further investigation. However, the environment (i.e., state information) of this model remains the same for all of the local actors while in our setting the environment of each BS agent is deferent from others based on its own energy demand and generation.
Moreover, the works in \cite{IEEEhowto:Wang_Meta_RL, IEEEhowto:Schweighofera_Meta_RL, IEEEhowto:Andrychowicz_L2L_by_GD, IEEEhowto:Mnih_async_a3c, IEEEhowto:Lowe_Multi_agent_actor_critic}, do not consider a multi-agent environment in which the policy of each agent relies on its own state information. In particular, such state information belongs to a non-i.i.d. learning environment when environmental dynamics become distinct among the agents.



\subsection{Contributions}
\begin{figure}[!t]
	\centerline{\includegraphics[width=8.6cm]{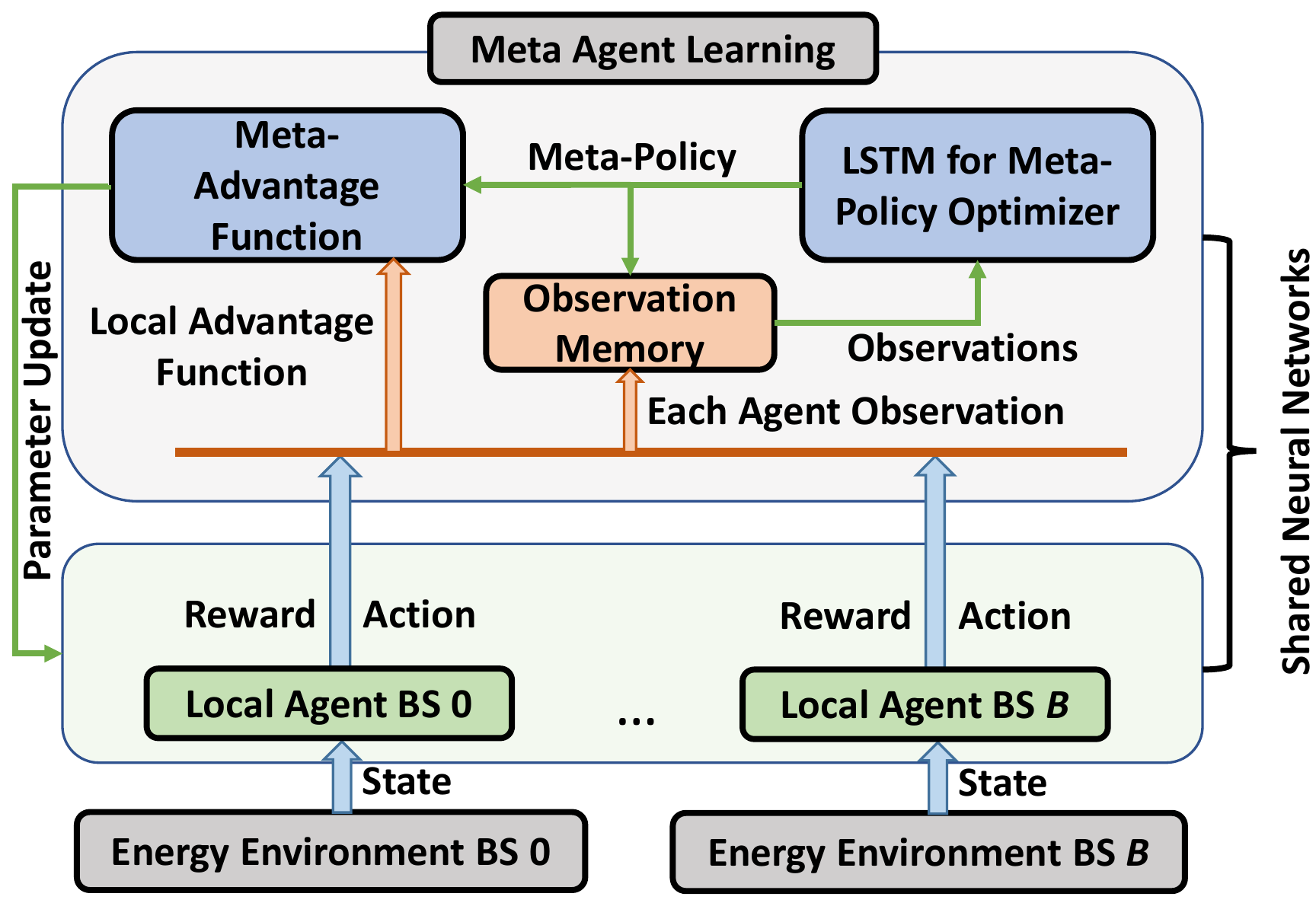}}
	\caption{Multi-agent meta-reinforcement learning framework of self-powered energy dispatch for sustainable edge computing.}
	\label{MAMRL_Training_Framework}
	\vspace{-4mm}
\end{figure}
The main contribution of this paper is a novel energy management framework for next-generation MEC in self-powered wireless network that is reliable against extreme uncertain energy demand and generation. We formulate a two-stage stochastic energy cost minimization problem that can balance renewable, non-renewable, and storage energy without knowing the actual demand. In fact, the formulated problem also investigates the realization of renewable energy generation after receiving the uncertain energy demand from the MEC applications and service requests. To solve this problem, we propose a multi-agent meta-reinforcement learning (MAMRL) framework that dynamically observes the non-i.i.d. behavior of time-varying features in both energy demand and generation at each BS and, then transfers those observations to obtain an energy dispatch decision and execute the energy dispatch policy to the self-powered BS. Fig. \ref{MAMRL_Training_Framework} illustrates how we propose to dispatch energy to ensure sustainable edge computing over a self-powered network using MAMRL framework. As we can see, each BS that includes small cell base stations (SBSs) and a macro base station (MBS) will act as a local agent and transfer their own decision (reward and action) to the meta-agent. Then, the meta-agent accumulates all of the non-i.i.d. observations from each local agent (i.e., SBSs and MBS) and optimizes the energy dispatch policy. The proposed MAMRL framework then provides feedback to each BS agent for exploring efficiently that acquire the right decision more quickly. Thus, the proposed MAMRL framework ensures autonomous decision making under an uncertain and unknown environment. Our key contributions include: 
\begin{itemize}
\item We formulate a self-powered energy dispatch problem for MEC-supported wireless network, in which the objective is to minimize the total energy consumption cost of network while considering the uncertainty of both energy consumption and generation. The formulated problem is, thus, a two-stage linear stochastic programming. In particular, the first stage makes a decision when energy demand is unknown, and the second stage discretizes the realization of renewable energy generation after knowing energy demand of the network.
	
\item To solve the formulated problem, we propose a new multi-agent meta-reinforcement learning framework by considering the skill transfer mechanism \cite{IEEEhowto:Wang_Meta_RL, IEEEhowto:Schweighofera_Meta_RL} between each local agent (i.e., self-powered BS) and meta-agent. In this MAMRL scheme, each local agent explores its own energy dispatch decision using Markovian properties for capturing the time-varying features of both energy demand and generation. Meanwhile, the meta-agent evaluates (exploits) that decision for each local agent and optimizes the energy dispatch decision. In particular, we design a long short-term memory (LSTM) as a meta-agent (i.e., run at MBS) that is capable of avoiding the incompetent decision from each local agent and learns the right features more quickly by maintaining its own state information.
		
\item We develop the proposed MAMRL energy dispatch framework in a semi-distributed manner. Each local agent (i.e., self-powered BS) estimates its own energy dispatch decision using local energy data (i.e., demand and generation), and provides observations to the meta-agent individually. Consequently, the meta-agent optimizes the decision centrally and assists the local agent toward a globally optimized decision. Thus, this approach not only reduces the computational complexity and communication overhead but it also mitigates the curse of dimensionality under the uncertainty by utilizing non-i.i.d. energy demand and generation from each local agent.

\item Experimental results using real datasets establish a significant performance gain of the energy dispatch under the deterministic, asymmetric, and stochastic environments. Particularly, the results show that the proposed MAMRL model saves up to $22.44\%$ of energy consumption cost over a baseline approach while achieving an average accuracy of around $95.8\%$ in a stochastic environment. Our approach also decreases the usage of non-renewable energy up to $11\%$ of total consumed energy. 
\end{itemize}

The rest of the paper is organized as follows. Section II presents the system model of self-powered edge computing. The problem formulation is described in Section III. Section IV provides MAMRL framework for solving energy dispatch problem. Experimental results are analyzed in Section V. Finally, conclusions are drawn in Section VI.

\section{System Model of Self-Powered Edge Computing}
\begin{figure}[!t]
	\centerline{\includegraphics[width=8.9cm]{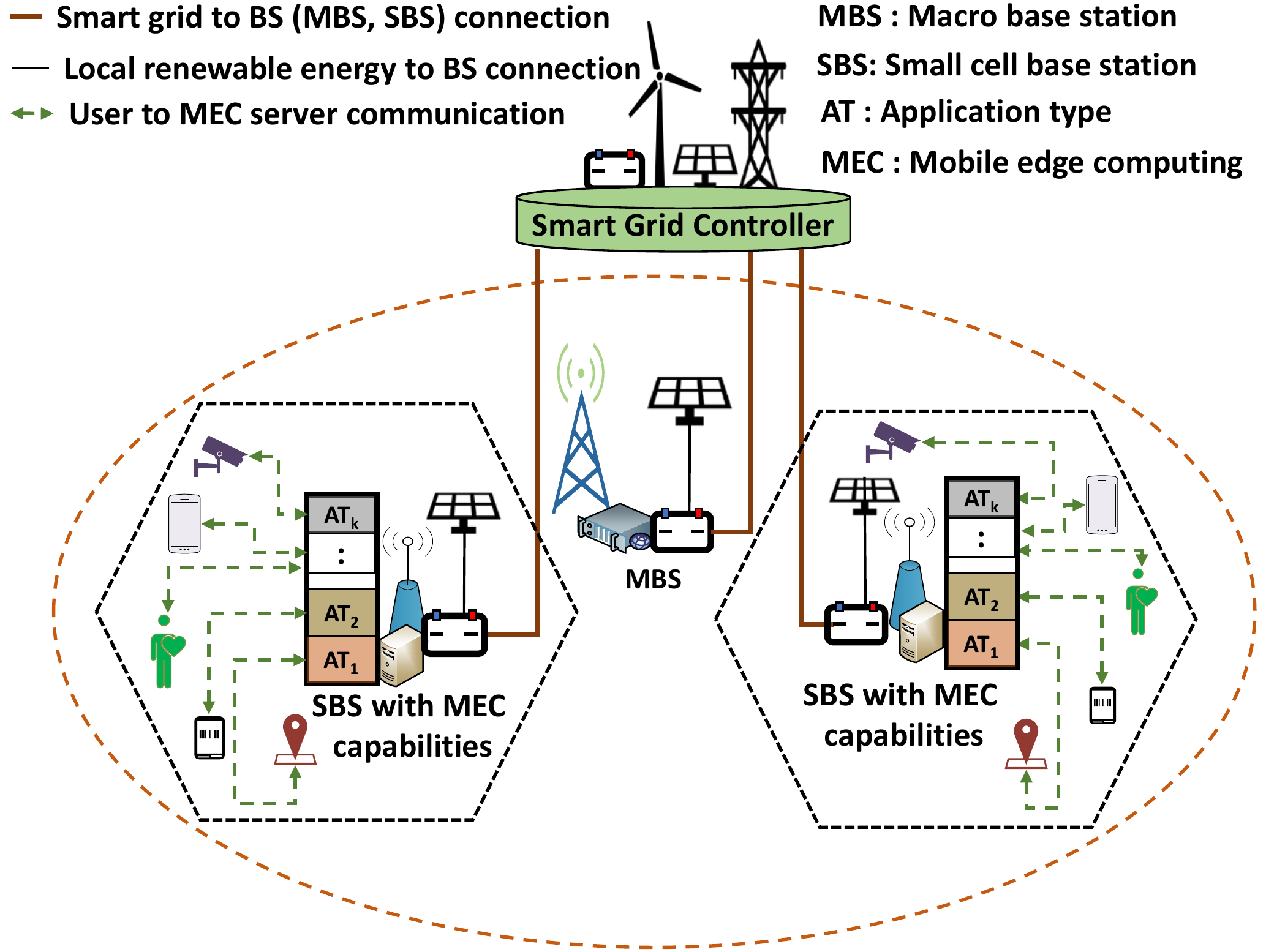}}
	\caption{System model for a self-powered wireless network with MEC capabilities.}
	\label{System_model}
	\vspace{-4mm}
\end{figure}
Consider a self-powered wireless network that is connected with a smart grid controller as shown in Fig. \ref{System_model}. Such a wireless network enables edge computing services for various MEC applications and services. The energy consumption of the network depends on network operations energy consumption along with the task loads of the MEC applications. Meanwhile, the energy supply of the network relies on the energy generation from renewable sources that are attached to the BSs, as well as both renewable and non-renewable sources of the smart grid. Furthermore, the smart grid controller is a representative of the main power grid (i.e, smart grid), where an additional amount of energy can be supplied via the smart grid controller to the network. Therefore, we will first discuss the energy demand model that includes MEC server energy consumption, and network communication energy consumption. We will then describe the energy generation model that consists of the non-renewable energy generation cost, surplus energy storage cost, and total energy generation cost. Table \ref{tab1} illustrates the summary of notations.  

\subsection{Energy Demand Model}
Consider a set $\mathcal{B} = \left\{{0,1,2,\dots, B}\right\}$ of $B+1$ ($0$ for MBS) BSs that encompass $B$ SBSs overlaid over a single MBS. Each BS $i \in \mathcal{B}$ includes a set $\mathcal{K}_i = \left\{{1,2,\dots,K_i}\right\}$ of $K_i$ MEC application servers. We consider a finite time horizon $\mathcal{T} = {1,2,\dots,T}$ with each time slot being indexed by $t$ and having a duration of 15 minutes \cite{IEEEhowto:Zhang_time_slot}. The observational period of each time slot $t$ ends at the $15$-th minute and is capable of capturing the changes of network dynamics \cite{IEEEhowto:Munir_Edge_Microgrid, IEEEhowto:Munir_GC_Multi_Agent, IEEEhowto:Han_time_10_min}. A set $\mathcal{J}_i$ of $J_i$ heterogeneous MEC application task requests from users will arrive to BS $i$ with an average task arrival rate $\lambda_i (t)$ (bits/s) at time $t$. The task arrival rate $\lambda_i (t)$ at BS $i \in \mathcal{B}$ follows a Poisson process at time slot $t$. BS $i$ integrates $K_i$ heterogeneous active MEC application servers that has $u_{k_i}(t)$ (bits/s) processing capacity. Thus, $J_i$ computational task requests will be accumulated into the service pool with an average traffic size $S_i(t)$ (bits) at time slot $t$. The average traffic arrival rate is defined as $\lambda_i (t) = \frac{1}{S_i(t)}$. Therefore, an $\emph{M/M/K}$ queuing model is suitable to model these $J_i$ user tasks using $K_i$ MEC servers at BS $i$ and time $t$ \cite{IEEEhowto:Abedin_Fog,IEEEhowto:Chang}. The task size of this queuing model is exponentially distributed since the average traffic size $S_i(t)$ is already known. Hence, the service rate of the BS $i$ is determined by $\mu_i(t) = \frac{1}{\mathbb{E}[\sum_{k_i \in \mathcal{K}_i}u_{k_i}(t)]}$. At any given time $t$, we assume that all of the tasks in $\mathcal{J}_i$ are uniformly distributed at each BS $i$. Thus, for a given MEC server task association indicator $\Upsilon_{jk_i}(t)=1$ if task $j$ is assigned to server $k$ at BS $i$, and $0$ otherwise, the average MEC server utilization is defined as follows \cite{IEEEhowto:Munir_Edge_Microgrid}: 
\begin{equation} \label{eq:edge_server_utilization}
\rho_i(t)  = 
\left\{
\begin{array}{ll}
\sum_{j \in \mathcal{J}_i}\sum_{k_i \in \mathcal{K}_i}\Upsilon_{jk_i}(t) \frac{\lambda_i (t)}{\mu_i(t)K_i}, \text{if }\Upsilon_{jk_i}(t)=1,\;\\
0,\;\;\;\;\;\;\;\;\;\;\text{otherwise.}
\end{array}
\right.
\end{equation}
\begin{table}[!t]
	\caption{Summary of notations.}
	\begin{center}
		\begin{tabular}{|c|c|}
			\hline
			\textbf{Notation}&{\textbf{Description}} \\
			\hline
			$\mathcal{B}$&Set of BSs (SBSs and MBS)\\
			\hline
			$\mathcal{K}_i$&Set of active servers under the BS $i \in \mathcal{B}$\\
			\hline
			$\mathcal{J}_i$&Set of user tasks at BS $i \in \mathcal{B}$\\
			\hline
			$\mathcal{R}$&Set of renewable energy sources \\
			\hline
			$\rho_i(t)$&Server utilization in BS $i \in \mathcal{B}$\\
			\hline
			$L$&No. of CPU cores \\
			\hline
			$R_i(t)$&Average downlink data of BS $i$ \\
			\hline
			$W_{ij}$&Fixed channel bandwidth of BS $i$ for user task $j$ \\
			\hline
			$P_i$&Transmission power of BS $i$ \\
			\hline
			$g_{ij}(t)$&Downlink channel gain between user task $j$ to BS $i$ \\
			\hline
			$I_{ij}(t)$&Co-channel interference for user task $j$ at BS $i$ \\
			\hline	
			$\delta_i$& Energy coefficient for BS $i \in \mathcal{B}$\\
			\hline
			$f$ & MEC server CPU frequency for a single core\\
			\hline
			$\tau$ & Server switching capacitance \\
			\hline
			$\eta^{\textrm{MEC}}_{\textrm{st}}(t)$ & MEC server static energy consumption\\
			\hline
			$\eta^{\textrm{MEC}}_{\textrm{idle}}(t)$ & MEC server idle state power consumption\\
			\hline
			$\varpi_{k}$ & Scaling factor of heterogeneous MEC CPU core\\
			\hline
			$\eta^{\textrm{net}}_{\textrm{st}}(t)$ & Static energy consumption of BS \\
			\hline
			$c^{\textrm{ren}}_t$&Renewable energy cost per unit\\
			\hline
			$c^{\textrm{non}}_t$&Non-renewable energy cost per unit\\
			\hline 
			$c^{\textrm{sto}}_t$&Storage energy cost per unit\\
			\hline	
			$\xi^{\textrm{ren}}_t$&Amount of renewable energy\\
			\hline
			$\xi^{\textrm{non}}_t$&Amount of non-renewable energy\\
			\hline
			$\xi^{\textrm{sto}}_t$&Amount of surplus energy\\
			\hline 
			$\xi^{\textrm{d}}_t$&Energy demand at time slot $t$\\
			\hline
			$\xi^{\textrm{D}}_t$&Random variable for energy demand\\
			\hline
			$\xi^{\textrm{ren}_{\textrm{max}}}_t$ & Maximum capacity of renewable energy at BS $i \in \mathcal{B}$\\
			\hline
			$\mathcal{O}_i$&Set of observation at BS $i \in \mathcal{B}$\\
			\hline
			$O(.)$&Big $O$ notation to represent complexity\\
			\hline
			$\beta$ & Entropy regularization coefficient\\
			\hline
			$\gamma$ & Discount factor\\
			\hline
			$\theta_{i}$ & Learning parameters for BS $i \in \mathcal{B}$\\
			\hline
			$\pi_{\theta_{i}}$ & Energy dispatch policy with parameters $\theta_{i}$ at BS $i \in \mathcal{B}$\\
			\hline
			$\phi$ & Meta-agent learning parameters\\
			\hline
		\end{tabular} 
		\label{tab1}
	\end{center}
	\vspace{-4mm}
\end{table}  

\subsubsection{MEC Server Energy Consumption}
In case of MEC server energy consumption, the computational energy consumption (dynamic energy) will be dependent on the CPU activity for executing computational tasks \cite{IEEEhowto:Tran_energy_MEC_no_downlink,IEEEhowto:Ndikumana_3C_Sim_para,IEEEhowto:Mao_Stochastic_Joint_Radio_Sim_para}. Further, such dynamic energy is also accounted with the thermal design power (TDP), memory, and disk I/O operations of the MEC server \cite{IEEEhowto:Tran_energy_MEC_no_downlink,IEEEhowto:Ndikumana_3C_Sim_para,IEEEhowto:Mao_Stochastic_Joint_Radio_Sim_para} and we denote as $\eta^{\textrm{MEC}}_{\textrm{st}}(t)$. Meanwhile, static energy $\eta^{\textrm{MEC}}_{\textrm{idle}}(t)$ includes the idle state power of CPU activities \cite{IEEEhowto:Mao_Stochastic_Joint_Radio_Sim_para, IEEEhowto:Chang_MEC_Power_Model}. We consider, a single core CPU with a processor frequency $f$ (cycles/s), an average server utilization $\rho_i(t)$ (using \eqref{eq:edge_server_utilization}) at time slot $t$, and a switching capacitance $\tau = 5 \times 10^{-27}$ (farad) \cite{IEEEhowto:Tran_energy_MEC_no_downlink}. The dynamic power consumption of such single core CPU can be calculated by applying a cubic formula $\tau \rho_i(t) f^3$  \cite{IEEEhowto:Chang_MEC_Power_Model, IEEEhowto:Rauber_MEC_Energy_Model}. Thus, energy consumption of $K_i$ MEC servers with $L$ CPU cores at BS $i$ is defined as follows:
\begin{equation} \label{eq:cpu_energy}
\xi^{\textrm{MEC}}_i(t)  = 
\left\{
\begin{array}{ll}
\sum_{k \in {K}_i} \sum_{l \in L} \tau \rho_i(t) f_{k_i}^3 \varpi_{k_{il}} + \; \eta^{\textrm{MEC}}_{\textrm{st}}(t), \text{if } \rho_i(t) > 0,\;\\
\eta^{\textrm{MEC}}_{\textrm{idle}}(t),\;\;\;\;\;\;\;\;\;\;\text{otherwise,}
\end{array}
\right.
\end{equation} 
where $\varpi_{k_{il}}$ denotes a scaling factor of heterogeneous CPU core of the MEC server. Thus, the value of $\varpi_{k_{il}}$ is dependent on the processor architecture \cite{IEEEhowto:Bertran_Multi_Core_CPU} that assures the heterogeneity of the MEC server.

\subsubsection{Base Station Energy Consumption}
The energy consumption needed for the operation of the network base stations (i.e., SBSs and MBS) includes two types of energy: dynamic and static energy consumption \cite{IEEEhowto:Total_Energy_Auer}. On one hand, a static energy consumption $\eta^{\textrm{net}}_{\textrm{st}}(t)$ includes the energy for maintaining the idle state of any BS, a constant power consumption for receiving packet from users, and the energy for wired transmission among the BSs. On the other hand, the dynamic energy consumption of the BSs depends on the amount of data transfer from BSs to users which essentially relates to the downlink \cite{IEEEhowto:Downlink_3GPP} transmit energy. Thus, we consider that each BS $i \in \mathcal{B}$ operates at a fixed channel bandwidth $W_{ij}$ and constant transmission power $P_{i}$ \cite{IEEEhowto:Downlink_3GPP}. Then the average downlink data of BS $i$ will be given by \cite{ IEEEhowto:Munir_Edge_Microgrid}: 
\begin{equation} \label{eq:data_rate}
R_{i}(t) = \sum_{i \in \mathcal{B}} \sum_{j \in \mathcal{J}_i} W_{ij} \\ \log_2\Big(1 + \frac{P_{i} g_{ij}(t)}{\sigma^2 + I_{ij}(t)}\Big)
\end{equation}
where $g_{ij}(t)$ represents downlink channel gain between user task $j$ to BS $i$, $\sigma^2$ determines a variance of an Additive White Gaussian Noise (AWGN), and $I_{ij}(t)$ denotes the co-channel interference \cite{IEEEhowto:Gu_Interference, IEEEhowto:Pantisano_Interference} among the BSs. Here, the co-channel interference $I_{ij}(t) = \sum_{i^\prime \in \mathcal{B}, i^\prime \ne i}P_{i^\prime} g_{{i^\prime}j}(t)$ relates to the transmissions from other BSs $i^\prime \in \mathcal{B}$ that use the same subchannels of $W_{ij}$. $P_{i^\prime}$ and $g_{{i^\prime}j}(t)$ represent, respectively, the transmit power and the channel gain of the BS $i^\prime \ne i \in \mathcal{B}$. Therefore, downlink energy consumption of the data transfer of BS $i \in \mathcal{B}$ is defined by $\frac{P_{i}S_i(t)}{R_{i}(t)}$ [watt-seconds or joule], where $\frac{S_i(t)}{R_{i}(t)}$ [seconds] determines the duration of transmit power $P_{i}$ [watt]. Thus, the network energy consumption for BS $i$ at time $t$ is defined as follows \cite{IEEEhowto:Total_Energy_Auer,IEEEhowto:Sun_EMM}:
\begin{equation} \label{eq:comm_energy} 
\xi^{\textrm{net}}_i(t) =  \sum_{j \in \mathcal{J}_i} \Big(\delta_i^{\textrm{net}} \frac{P_{i}S_i(t)}{R_{i}(t)}   + \eta^{\textrm{net}}_{\textrm{st}}(t)\Big),
\end{equation}
where $\delta_i^{\textrm{net}}$ determines the energy coefficient for transferring data through the network. In fact, the value of $\delta_i^{\textrm{net}}$ depends on the type of the network device (e.g., $\delta_i^{\textrm{net}}=2.8$ for a $6$ unit transceiver remote radio head \cite{IEEEhowto:Total_Energy_Auer}).

\subsubsection{Total Energy Demand}
The total energy consumption (demand) of the network consists of both MEC server computational energy (in \eqref{eq:cpu_energy}) consumption, and network the operational energy (i.e., BSs energy consumption in \eqref{eq:comm_energy}). Thus, the  overall energy demand of the network at time slot $t$ is given as follows:
\begin{equation} \label{eq:total_energy_demand}
\xi^{\textrm{d}}_t = \sum_{i \in \mathcal{B}} \Big(\xi^{\textrm{net}}_i(t) +\xi^{\textrm{MEC}}_i(t)\Big).
\end{equation}
The demand $\xi^{\textrm{d}}_t$ is random over time and completely depends on the computational tasks load of the MEC servers. 

\subsection{Energy Generation Model}
The energy supply of the self-powered wireless network with MEC capabilities relates to the network's own renewable (e.g., solar, wind, biofuels, etc.) sources as well as the main grid's non-renewable (e.g., diesel generator, coal power, and so on) energy sources \cite{IEEEhowto:Lee_self_powered, IEEEhowto:Wei_A3C_Hybrid_Power}. In this energy generation model, we consider a set $\mathcal{R} = \left\{{\mathcal{R}_0,\mathcal{R}_1,\dots, \mathcal{R}_B}\right\}$ of renewable energy sources of the network, with each element $\mathcal{R}_i$ representing the set of renewable energy sources of BS $i \in \mathcal{B}$. Each renewable energy source $q \in \mathcal{R}_i$ at BS $i \in \mathcal{B}$ can generate an amount $\xi^{\textrm{ren}}_{iq}(t)$ of renewable energy at time $t$. Therefore, the total amount of renewable energy generation $\xi^{\textrm{ren}}_i(t)$ at BS $i \in \mathcal{B}$ will be $\xi^{\textrm{ren}}_i(t) = \sum_{q \in \mathcal{R}_i} \xi^{\textrm{ren}}_{iq}(t)$ for time slot $t$. Thus, the total renewable energy generation for the considered network at time $t$ is defined as $\xi^{\textrm{ren}}_t= \sum_{i \in \mathcal{B}} \xi^{\textrm{ren}}_{i}(t)$. The maximum limit of this renewable energy $\xi^{\textrm{ren}}_t$ is less than or equal to the maximum capacity $\xi^{\textrm{ren}_{\textrm{max}}}_t$ of renewable energy generation at time period $t$. Thus, we consider a maximum storage limit that is equal to the maximum capacity $\xi^{\textrm{ren}_{\textrm{max}}}_t$ of the renewable energy generation \cite{IEEEhowto:Panwar_Role_of_Renewable,IEEEhowto:Energy_Storage_Chacra,IEEEhowto:Microgrid_Energy_Storage_Kanchev}.
Further, the self-powered wireless network is able to get an additional non-renewable energy amount $\xi^{\textrm{non}}_t$ from the main grid at time $t$. The per unit renewable and non-renewable energy cost are defined by ${c^{\textrm{ren}}_t}$ and ${c^{\textrm{non}}_t}$, respectively. In general, the renewable energy cost only depends on the maintenance cost of the renewable energy sources \cite{IEEEhowto:Panwar_Role_of_Renewable,IEEEhowto:Energy_Storage_Chacra,IEEEhowto:Microgrid_Energy_Storage_Kanchev}. Therefore, the per unit non-renewable energy cost is greater than the renewable energy cost ${c^{\textrm{non}}_t}>{c^{\textrm{ren}}_t}$. Additionally, the surplus amount of the energy $\xi^{\textrm{sto}}_t$ at time $t$ can be stored in energy storage medium for the future usages \cite{IEEEhowto:Energy_Storage_Chacra,IEEEhowto:Microgrid_Energy_Storage_Kanchev} and the energy storage cost of per unit energy store is denoted by ${c^{\textrm{sto}}_t}$.

\subsubsection{Non-renewable Energy Generation Cost}
In order to fulfill the energy demand ${\xi^{\textrm{d}}_t}$ when it is greater than the generated renewable energy $\xi^{\textrm{ren}}_t$, the main grid can provide an additional amount of energy $\xi^{\textrm{non}}_t$ from its non-renewable sources. Thus, the non-renewable energy generation cost ${C^{\textrm{non}}_t}$ of the network is determined as follows:
\begin{equation} \label{eq:non-renewable_cost}
C^{\textrm{non}}_t = 
\left\{
\begin{array}{ll}
c^{\textrm{non}}_t[{\xi^{\textrm{d}}_t}-\xi^{\textrm{ren}}_t] ,\;\text{if}\;\; {\xi^{\textrm{d}}_t} > \xi^{\textrm{ren}}_t,\;\\
0,\;\;\;\;\;\;\;\;\;\;\;\;\;\;\text{otherwise},
\end{array}
\right.
\end{equation}
where ${c^{\textrm{non}}_t}$ represents a unit energy cost.  

\subsubsection{Surplus Energy Storage Cost}  
The surplus amount of energy is stored in a storage medium when ${\xi^{\textrm{d}}_t}<\xi^{\textrm{ren}}_t$ (i.e., energy demand is smaller than the renewable energy generation) at time $t$. We consider the per unit energy storage cost $c^{\textrm{sto}}_t$. This storage cost depends on the storage medium and amount of the energy store at time $t$  \cite{IEEEhowto:Energy_Storage_Chacra,IEEEhowto:GreenCharge_Mishra,IEEEhowto:Incentivizing_Tran,IEEEhowto:Xu_storage_cost10}. With the per unit energy storage cost $c^{\textrm{sto}}_t$, the total storage cost at time $t$ is defined as follows:
\begin{equation} \label{eq:storage_cost}
C^{\textrm{sto}}_t = 
\left\{
\begin{array}{ll}
 c^{\textrm{sto}}_t[\xi^{\textrm{ren}}_t-{\xi^{\textrm{d}}_t}] ,\;\text{if}\;\; {\xi^{\textrm{d}}_t} < \xi^{\textrm{ren}}_t,\;\\
0,\;\;\;\;\;\;\;\;\;\;\;\;\;\;\text{otherwise}.
\end{array}
\right.
\end{equation}

\subsubsection{Total Energy Generation Cost}
The total energy generation cost includes renewable, non-renewable, and storage energy cost. Naturally, this total energy generation cost will depend on the energy demand $\xi^{\textrm{d}}_t$ of the network at time $t$. Therefore, the total energy generation cost at time $t$ is defined as follows:
\begin{equation} \label{eq:overall_gen_cost}
\begin{split}
Q(\xi^{\textrm{ren}}_t,\xi^{\textrm{d}}_t) = {c^{\textrm{ren}}_t}\xi^{\textrm{ren}}_t \;+ {c^{\textrm{non}}_t}{[{\xi^{\textrm{d}}_t}-\xi^{\textrm{ren}}_t]}_+ \\  +  {c^{\textrm{sto}}_t}{[\xi^{\textrm{ren}}_t-{\xi^{\textrm{d}}_t}]}_+,
\end{split}
\end{equation} 
where the energy cost of the renewable, non-renewable, and storage energy are given by ${c^{\textrm{ren}}_t}\xi^{\textrm{ren}}_t$, ${c^{\textrm{non}}_t}{[{\xi^{\textrm{d}}_t}-\xi^{\textrm{ren}}_t]}_+$, and  ${c^{\textrm{sto}}_t}{[\xi^{\textrm{ren}}_t-{\xi^{\textrm{d}}_t}]}_+$, respectively. In \eqref{eq:overall_gen_cost}, energy demand $\xi^{\textrm{d}}_t$ and renewable energy generation $\xi^{\textrm{ren}}_t$ are stochastic in nature. The energy cost of non-renewable energy \eqref{eq:non-renewable_cost} and storage energy \eqref{eq:storage_cost} completely rely on energy demand $\xi^{\textrm{d}}_t$ and renewable energy generation $\xi^{\textrm{ren}}_t$. Hence, to address the uncertainty of both energy demand and renewable energy generation in a self-powered wireless network, we formulate a two-stage stochastic programing problem. In particular, the first stage makes a decision of the energy dispatch without knowing the actual demand of the network. Then we make further energy dispatch decisions by analyzing the uncertainty of the network demand in the second stage. A detailed discussion of the problem formulation is given in the following section.
  

\section{Problem Formulation with a Two-Stage Stochastic Model}
We now consider the case in which the non-renewable energy cost is greater than the renewable energy cost, $c^{\textrm{non}}_t > c^{\textrm{ren}}_t$ that is often the case in a practical smart grid as discussed in \cite{IEEEhowto:Panwar_Role_of_Renewable}, \cite{IEEEhowto:Energy_Storage_Chacra}, \cite{IEEEhowto:Microgrid_Energy_Storage_Kanchev}, and \cite{IEEEhowto:Per_Unit_Energy_Generation_Cost}. Here, $\xi^{\textrm{ren}}_t$ and $\xi^{\textrm{d}}_t$ are the continuous variables over the observational duration $t$. The objective is to minimize the total energy consumption cost $Q(\xi^{\textrm{ren}}_t,\xi^{\textrm{d}}_t)$. $\xi^{\textrm{ren}}_t$ is the decision variable and the energy demand $\xi^{\textrm{d}}_t$ is a parameter. When the energy demand $\xi^{\textrm{d}}_t$ is known, the optimization problem will be:   
\begin{subequations}\label{Opt_1_1}
	\begin{align}
		\chi =
		\underset{{\xi^{\textrm{ren}}_t \ge 0}}\min
		&\; Q(\xi^{\textrm{ren}}_t,\xi^{\textrm{d}}_t).   \tag{\ref{Opt_1_1}} 
	\end{align}
\end{subequations}
In problem \eqref{Opt_1_1}, after removing the non-negativity constraints $\xi^{\textrm{ren}}_t \ge 0$, we can rewrite the objective function in the form of piecewise linear functions as follows:
\begin{equation}\label{Opt_1_2}
\begin{split}
Q(\xi^{\textrm{ren}}_t,\xi^{\textrm{d}}_t) = 
\underset{{\xi^{\textrm{ren}}_t}}\max\left\{
  \Big( (c^{\textrm{ren}}_t-c^{\textrm{non}}_t)\xi^{\textrm{ren}}_t+c^{\textrm{non}}_t\xi^{\textrm{d}}_t\Big),\right.\\ \left. \Big((c^{\textrm{ren}}_t+c^{\textrm{sto}}_t)\xi^{\textrm{ren}}_t-c^{\textrm{sto}}_t\xi^{\textrm{d}}_t \Big)\right\}. 
\end{split}
\end{equation}
Where $(c^{\textrm{ren}}_t-c^{\textrm{non}}_t)\xi^{\textrm{ren}}_t+c^{\textrm{non}}_t\xi^{\textrm{d}}_t$ and $(c^{\textrm{ren}}_t+c^{\textrm{sto}}_t)\xi^{\textrm{ren}}_t-c^{\textrm{sto}}_t\xi^{\textrm{d}}_t$ determine the cost of non-renewable (i.e., ${\xi^{\textrm{d}}_t} > \xi^{\textrm{ren}}_t$) and storage (i.e.,  ${\xi^{\textrm{d}}_t} < \xi^{\textrm{ren}}_t$) energy, respectively. Therefore, we have to choose one out of the two cases. In fact, if the energy demand $\xi^{\textrm{d}}_t$ is known and also the amount of renewable energy $\xi^{\textrm{ren}}_t$ is the same as the energy demand, then problem \eqref{Opt_1_2} provides the optimal decision in order to exact amount of demand $\xi^{\textrm{d}}_t$. However, the challenge here is to make a decision about the renewable energy $\xi^{\textrm{ren}}_t$ usage before the demand becomes known. To overcome this challenge, we consider the energy demand $\xi^{\textrm{D}}_t$ as a random variable whose probability distribution can be estimated from the previous history of the energy demand. We can re-write problem \eqref{Opt_1_1} using the expectation of the total cost as follows:  
\begin{subequations}\label{Opt_1_3}
	\begin{align}
		\underset{{\xi^{\textrm{ren}}_t \ge 0}}\min
		&\;  \mathbb{E}[Q(\xi^{\textrm{ren}}_t,\xi^{\textrm{D}}_t)].   \tag{\ref{Opt_1_3}} 
	\end{align}
\end{subequations}
%
%
The solution of problem \eqref{Opt_1_3} will provide an optimal result on average. However, in the practical scenario, we need to solve problem \eqref{Opt_1_3} repeatedly over the uncertain energy demand $\xi^{\textrm{D}}_t$. 
Thus, this solution approach does not significantly affect our model in terms of scalability while $B+1$ number of BSs generates a large variety of energy demand over the observational period of $t$. In fact, energy demand and generation can change over time for each BS $i \in \mathcal{B}$, and they can also induce large variations of demand-generation among the BSs. Hence, the solution to problem \eqref{Opt_1_3} cannot rely on an iterative scheme due to a lake of the adaptation for uncertain change of energy demand and generation over time.

We consider the moment of random variable $\xi^{\textrm{D}}_t$ that has a finitely supported distribution and takes values $\xi^{\textrm{D}}_{t0}, \dots, \xi^{\textrm{D}}_{tB}$ with respective probabilities $p_0, \dots, p_B$ of BSs $B+1$. The cumulative distribution function (CDF) $H(\xi^{\textrm{D}}_t)$ of energy demand $\xi^{\textrm{D}}_t$ is a step function and jumps of size $p_i$ at each demand $\xi^{\textrm{d}}_{ti}$. Therefore, the probability distribution of each BS energy demand $\xi^{\textrm{d}}_{ti}$ belongs to the CDF $H(\xi^{\textrm{D}}_t)$ of historical observation of energy demand $\xi^{\textrm{D}}_t$. In this case, we can convert problem \eqref{Opt_1_3} into a deterministic optimization problem and the expectation of energy usage cost $\mathbb{E}[Q(\xi^{\textrm{ren}}_t,\xi^{\textrm{D}}_t)]$ is determined by $\sum_{i \in \mathcal{B}} p_iQ(\xi^{\textrm{ren}}_t,\xi^{\textrm{d}}_t)$. Thus, we can rewrite the problem \eqref{Opt_1_1} as a linear programming problem using the representation in \eqref{Opt_1_2} as follows:
\begin{subequations}\label{Opt_1_4}
	\begin{align}
		\underset{{\xi^{\textrm{ren}}_t , \chi}}\min
		&\;  \chi \tag{\ref{Opt_1_4}} \\
		\text{s.t.} \quad & \label{Opt_1_4:const1} \chi \ge (c^{\textrm{ren}}_t-c^{\textrm{non}}_t)\xi^{\textrm{ren}}_t+c^{\textrm{non}}_t\xi^{\textrm{d}}_t,\\
		&\label{Opt_1_4:const2} \chi \ge (c^{\textrm{ren}}_t+c^{\textrm{sto}}_t)\xi^{\textrm{ren}}_t-c^{\textrm{sto}}_t\xi^{\textrm{d}}_t,\\
		&\label{Opt_1_4:const3} \xi^{\textrm{ren}_{\textrm{max}}}_t \ge \xi^{\textrm{ren}}_t \ge 0.
	\end{align}
\end{subequations}
For a fixed value of the renewable energy $\xi^{\textrm{ren}}_t$, problem \eqref{Opt_1_4} is an equivalent of problem \eqref{Opt_1_2}. Meanwhile, problem \eqref{Opt_1_4} is equal to $Q(\xi^{\textrm{ren}}_t,\xi^{\textrm{d}}_t)$. We have converted the piecewise linear function from problem  \eqref{Opt_1_2} into the inequality constraints \eqref{Opt_1_4:const1} and \eqref{Opt_1_4:const2}. Constraint \eqref{Opt_1_4:const3} ensures a limit on the maximum allowable renewable energy usage. We consider $p_i$ as a highest probability of energy demand at each BS $i\in \mathcal{B}$. Therefore, for $B+1$ BSs, we define $p_0, \dots, p_B$ as the probability of energy demand with respect to BSs $i = 0, \dots, B$. Thus, we can rewrite the problem \eqref{Opt_1_3} for $B+1$ BSs $\boldsymbol{\xi}^{\textrm{D}}_t = (\xi^{\textrm{D}}_{t0}, \dots, \xi^{\textrm{D}}_{tB})$ is as follows:
\begin{subequations}\label{Opt_1_5}
	\begin{align}
		\underset{{\xi^{\textrm{ren}}_t , {\chi_0}, \dots, \chi_B}}\min
		&\;  \sum_{i \in \mathcal{B}} p_i\chi_i, \tag{\ref{Opt_1_5}} \\
		\begin{split}
		\text{s.t.} \quad & \label{Opt_1_5:const1} \chi_i \ge (c^{\textrm{ren}}_t-c^{\textrm{non}}_t)\xi^{\textrm{ren}}_t  + c^{\textrm{non}}_t\xi^{\textrm{D}}_{ti}, \forall i \in \mathcal{B}, 
		\end{split}\\
		\begin{split}
		&\label{Opt_1_5:const2} \chi_i \ge (c^{\textrm{ren}}_t+c^{\textrm{sto}}_t)\xi^{\textrm{ren}}_t-c^{\textrm{sto}}_t\xi^{\textrm{D}}_{ti}, \forall i \in \mathcal{B},
		\end{split}\\
		&\label{Opt_1_5:const3}  \xi^{\textrm{ren}_{\textrm{max}}}_t \ge \xi^{\textrm{ren}}_t \ge 0,
	\end{align}
\end{subequations}
where $p_i$ represents the highest probability of the energy demand  $\xi^{\textrm{D}}_{ti}=\xi^{\textrm{d}}_{ti}$, in which $\xi^{\textrm{D}}_{ti}$ is a random variable and $\xi^{\textrm{d}}_{ti}$ denotes a realization of energy demand on BS $i\in \mathcal{B}$ at time $t$.
The value of $p_i$ belongs to the empirical CDF $H(\boldsymbol{\xi}^{\textrm{D}}_{ti})$ of the energy demand $\xi^{\textrm{D}}_{ti}$ for BS $i\in \mathcal{B}$. This CDF is calculated from the historical observation of the energy demand at BS $i\in \mathcal{B}$.
In fact, for a fixed value of non-renewable energy $\xi^{\textrm{ren}}_t$, problem (13) is separable. As a result, we can decompose this problem with a structure of two-stage linear stochastic programming problem \cite{IEEEhowto:Liu_Two_Stage, IEEEhowto:Zhou_two_stage_satellite}.


To find an approximation for a random variable with a finite probability distribution, we decompose problem \eqref{Opt_1_5} in a two-stage linear stochastic programming under uncertainty. The decision is made using historical data of energy demand, which is fully independent from the future observation. As a result, the first stage of \emph{self-powered energy dispatch} problem for sustainable edge computing is formulated as follows:
\begin{subequations}\label{Opt_1_6}
	\begin{align}
		\underset{{\xi^{\textrm{ren}}_t \ge 0}}\min
		&\;  (c^{\textrm{ren}}_t)^{\top}\xi^{\textrm{ren}}_t + \mathbb{E}[Q(\xi^{\textrm{ren}}_t,\boldsymbol{\xi}^{\textrm{D}}_t)],     \tag{\ref{Opt_1_6}}\\
		\text{s.t.} \quad & \label{Opt_1_6:const1}\xi^{\textrm{ren}_{\textrm{max}}}_t \ge \xi^{\textrm{ren}}_t \ge 0,
	\end{align}
\end{subequations}
where $Q(\xi^{\textrm{ren}}_t,\boldsymbol{\xi}^{\textrm{D}}_t)$ determines an optimal value of the second stage problem. In problem \eqref{Opt_1_6}, the decision variable $\xi^{\textrm{ren}}_t$ is calculated before the realization of uncertain energy demand $\boldsymbol{\xi}^{\textrm{D}}_t$. Meanwhile, at the first stage of the formulated problem \eqref{Opt_1_6}, the cost $(c^{\textrm{ren}}_t)^{\top}\xi^{\textrm{ren}}_t$ is minimized for the decision variable $\xi^{\textrm{ren}}_t$ which then allows us to estimate the expected energy cost $\mathbb{E}[Q(\xi^{\textrm{ren}}_t,\boldsymbol{\xi}^{\textrm{D}}_t)]$ for the second stage decision. Constraint \eqref{Opt_1_6:const1} provides a boundary for the maximum allowable renewable energy usage. Thus, based on the decision of the first stage problem, the second stage problem can be defined as follows:
\begin{subequations}\label{Opt_1_7}
	\begin{align}
		\underset{{\xi^{\textrm{non}}_t, \xi^{\textrm{sto}}_t}}\min
		&\;  (c^{\textrm{non}}_t)^{\top}\xi^{\textrm{non}}_t - (c^{\textrm{sto}}_t)^{\top}\xi^{\textrm{sto}}_t,      \tag{\ref{Opt_1_7}} \\
		\text{s.t.} \quad & \label{Opt_1_7:const1}  \xi^{\textrm{sto}}_t =  |\xi^{\textrm{ren}}_t - \xi^{\textrm{non}}_t|, \\
		&\label{Opt_1_7:const2} 0 \le \xi^{\textrm{non}}_t \le (\boldsymbol{\xi}^{\textrm{D}}_t)^{\top},\\
		&\label{Opt_1_7:const3}  \xi^{\textrm{non}}_t \ge 0.
	\end{align}
\end{subequations}
In the second stage problem $\eqref{Opt_1_7}$, the decision variables $\xi^{\textrm{non}}_t$ and $\xi^{\textrm{sto}}_t$ depend on the realization of the energy demand $\boldsymbol{\xi}^{\textrm{D}}_t$ of the first stage problem $\eqref{Opt_1_6}$, where, $\xi^{\textrm{ren}}_t$ determines the amount of renewable energy usage at time $t$. The first constraint  $\eqref{Opt_1_7:const1}$ is an equality constraint that determines the surplus amount of energy $\xi^{\textrm{sto}}_t$ must be equal to the absolute value difference between the usage of renewable $\xi^{\textrm{ren}}_t$ and non-renewable $\xi^{\textrm{non}}_t$ energy amount. The second constraint $\eqref{Opt_1_7:const2}$ is an inequality constraint that uses the optimal demand value from the first stage realization. In particular, the value of demand comes from $\eqref{eq:total_energy_demand}$ that is the historical observation of energy demand. Finally, the constraint  $\eqref{Opt_1_7:const3}$ protects from the non-negativity for the non-renewable energy $\xi^{\textrm{non}}_t$ usage.         

The formulated problems $\eqref{Opt_1_6}$ and $\eqref{Opt_1_7}$ can characterize the uncertainty between network energy demand and renewable energy generation. Particularly, the second stage problem $\eqref{Opt_1_7}$ contains random demand $\boldsymbol{\xi}^{\textrm{D}}_t$ that leads the optimal cost $\mathbb{E}[Q(\xi^{\textrm{ren}}_t,\boldsymbol{\xi}^{\textrm{D}}_t)]$ as a random variable. As a result, we can rewrite the problems $\eqref{Opt_1_6}$ and $\eqref{Opt_1_7}$ in a one large linear programming problem for $B+1$ BSs and the problem formulation is as follows:
\begin{subequations}\label{Opt_1_8}
	\begin{align}
	\begin{split}
	\underset{{{\xi}^{\textrm{ren}}_{t},{\xi}^{\textrm{non}}_{t}, {\xi}^{\textrm{sto}}_{t}}}\min
	  \sum_{t \in \mathcal{T}}  \Big((c^{\textrm{ren}}_{t})^{\top}\xi^{\textrm{ren}}_{t} \; + \;\;\;\;\;\;\;\;\;\;\;\;\;\;\;\;\;\;\;\;\; \\  \sum_{i \in \mathcal{B}} p_i [(c^{\textrm{non}}_t)^{\top}\xi^{\textrm{non}}_{ti} - (c^{\textrm{sto}}_t)^{\top}\xi^{\textrm{sto}}_{ti}]\Big), 
	\end{split} \tag{\ref{Opt_1_8}}
	\\
	\text{s.t.} \quad & \label{Opt_1_8:const1}  \xi^{\textrm{sto}}_{ti} =  |\xi^{\textrm{ren}}_{ti} - \xi^{\textrm{non}}_{ti}|, \forall i \in \mathcal{B}, \\
	&\label{Opt_1_8:const2} 0 \le \xi^{\textrm{non}}_{ti} \le \xi^{\textrm{D}}_{ti}, \forall i \in \mathcal{B},\\
	&\label{Opt_1_8:const3}  \xi^{\textrm{non}}_{ti} \ge 0, \forall i \in \mathcal{B},\\
	&\label{Opt_1_8:const4} \xi^{\textrm{ren}_{\textrm{max}}}_t \ge \xi^{\textrm{ren}}_{ti} \ge 0, \forall i \in \mathcal{B}.
	\end{align}
\end{subequations}
In problem $\eqref{Opt_1_8}$, for $B+1$ BSs, energy demand $\xi_{t0}^{\textrm{D}} \dots \xi_{tB}^{\textrm{D}}$ happens with positive probabilities $p_0 \dots p_B$ and $ \sum_{i \in \mathcal{B}}p_i = 1$. The decision variables are ${\xi}^{\textrm{ren}}_t$, ${\xi}^{\textrm{non}}_{t}$ and ${\xi}^{\textrm{sto}}_t$, which represent the amount of renewable, non-renewable, and storage energy, respectively. Constraint $\eqref{Opt_1_8:const1}$ defines a relationship among all of the decision variables ${\xi}^{\textrm{ren}}_t$, ${\xi}^{\textrm{non}}_t$ and ${\xi}^{\textrm{sto}}_t$. In essence, this constraint discretizes the surplus amount of energy for storage. Hence, constraint $\eqref{Opt_1_8:const2}$ ensures the utilization of non-renewable energy based on the energy demand of the network. Constraint $\eqref{Opt_1_8:const3}$ ensures that the decision variable ${\xi}^{\textrm{non}}_t$ will not be a negative value. Finally, constraint $\eqref{Opt_1_8:const4}$ restricts the renewable energy ${\xi}^{\textrm{ren}}_t$ usages in to maximum capacity $\xi^{\textrm{ren}_{\textrm{max}}}_t$ at time $t$. Problem $\eqref{Opt_1_8}$ is an integrated form of the first-stage problem in $\eqref{Opt_1_6}$ and the second-stage problem in $\eqref{Opt_1_7}$, where the solution of  $\xi^{\textrm{non}}_{t}$ and $\xi^{\textrm{sto}}_t$ completely depends on realization of demand $\xi^{\textrm{D}}_{ti}$ for all $B+1$ BSs. The decision of the $\xi^{\textrm{ren}}_t$ comes before the realization of demand $\xi^{\textrm{D}}_{ti}$ and, thus, the estimation of renewable energy generation $\xi^{\textrm{ren}}_t$ will be independent and random. Therefore, problem $\eqref{Opt_1_8}$ holds the property of relatively complete recourse. In problem $\eqref{Opt_1_8}$, the number of variables and constraints is proportional to the numbers of BSs, $B+1$. Additionally, the complexity of the decision problem $\eqref{Opt_1_8}$ leads to $\mathcal{O}(2^{|\mathcal{T}| \times |\mathcal{B}|})$ due to the combinatorial properties of the decisions and constraints \cite{IEEEhowto:Liu_Two_Stage, IEEEhowto:Zhou_two_stage_satellite, IEEEhowto:Abedin_MM1_Fog}.

The goal of the \emph{self-powered energy dispatch} problem $\eqref{Opt_1_8}$ is to find an optimal energy dispatch policy that includes amount of renewable ${\xi}^{\textrm{ren}}_{ti}$, non-renewable ${\xi}^{\textrm{non}}_{ti}$, and storage ${\xi}^{\textrm{sto}}_{ti}$ energy of each BS $i \in \mathcal{B}$ while minimizing the energy consumption cost. Meanwhile, such energy dispatch policy relies on an empirical probability distribution $H(\boldsymbol{\xi}^{\textrm{D}}_t)$ of historical demand at each BS $i \in \mathcal{B}$ at time $t$. In order to solve problem $\eqref{Opt_1_8}$, we choose an approach that does not rely on the conservativeness of a theoretical probability distribution of energy demand in problem $\eqref{Opt_1_8}$, and also will capture the uncertainty of renewable energy generation from the historical data. 
In contrast, we can construct a theoretical probability distribution of energy demand $\boldsymbol{\xi}^{\textrm{D}}_t$ when we know what the exact distribution is as well as what its parameters will be (e.g., mean, variance, and standard deviation).
In fact, in practice, the distribution of energy demand $\boldsymbol{\xi}^{\textrm{D}}_t$ is unknown and instead, a certain amount of historical energy demand data are available. As a result, we cannot rely on this distribution to measure uncertainty while the renewable energy generation and energy demand are random over time.
Hence, we can obtain time-variant features of both energy demand and generation by characterizing the Markovian properties from the historical observation over time. In particular, we capture the dynamics of Markovian by considering a data-driven approach. This approach can overcome the conservativeness of theoretical probability distribution as historical observation goes to finitely many.


To prevalence the aforementioned contemporary, we propose a multi-agent meta-reinforcement learning framework that can explore the Markovian behavior from historical energy demand and generation of each BS $i \in \mathcal{B}$. Meanwhile, meta-agent can cope with such time-varying features to a globally optimal energy dispatch policy for each BS $i \in \mathcal{B}$. 

We design an MAMRL framework by converting the cost minimization problem $\eqref{Opt_1_8}$ to a reward maximization problem that we then solve with a data-driven approach. In the MAMRL setting, each agent works as a local agent for each BS $i \in \mathcal{B}$ and determines an observation (i.e., exploration) for the decision variables, renewable $\xi^{\textrm{ren}}_{ti}$, non-renewable $\xi^{\textrm{non}}_{ti}$, and storage $\xi^{\textrm{sto}}_{ti}$ energy. The goal of this exploration is to find time-varying features from the local historical data so that the energy demand $\xi^{\textrm{d}}_{ti}$ of the network is satisfied. Furthermore, using these observations and current state information, a meta-agent is used to determine a stochastic energy dispatch policy. Thus, to obtain such dispatch policy, the meta-agent only requires the observations (behavior) from each local agent. Then, the meta-agent can evaluate (exploit) behavior toward an optimal decision for dispatching energy. Further, the MAMRL approach is capable of capturing the exploration-exploitation tradeoff in a way that the meta-agent optimizes decisions of the each self-powered BS under uncertainty. A detailed discussion of the MAMRL framework is given in the following section.
\begin{figure}[!t]
	\centerline{\includegraphics[width=8.6cm, height=8.6cm]{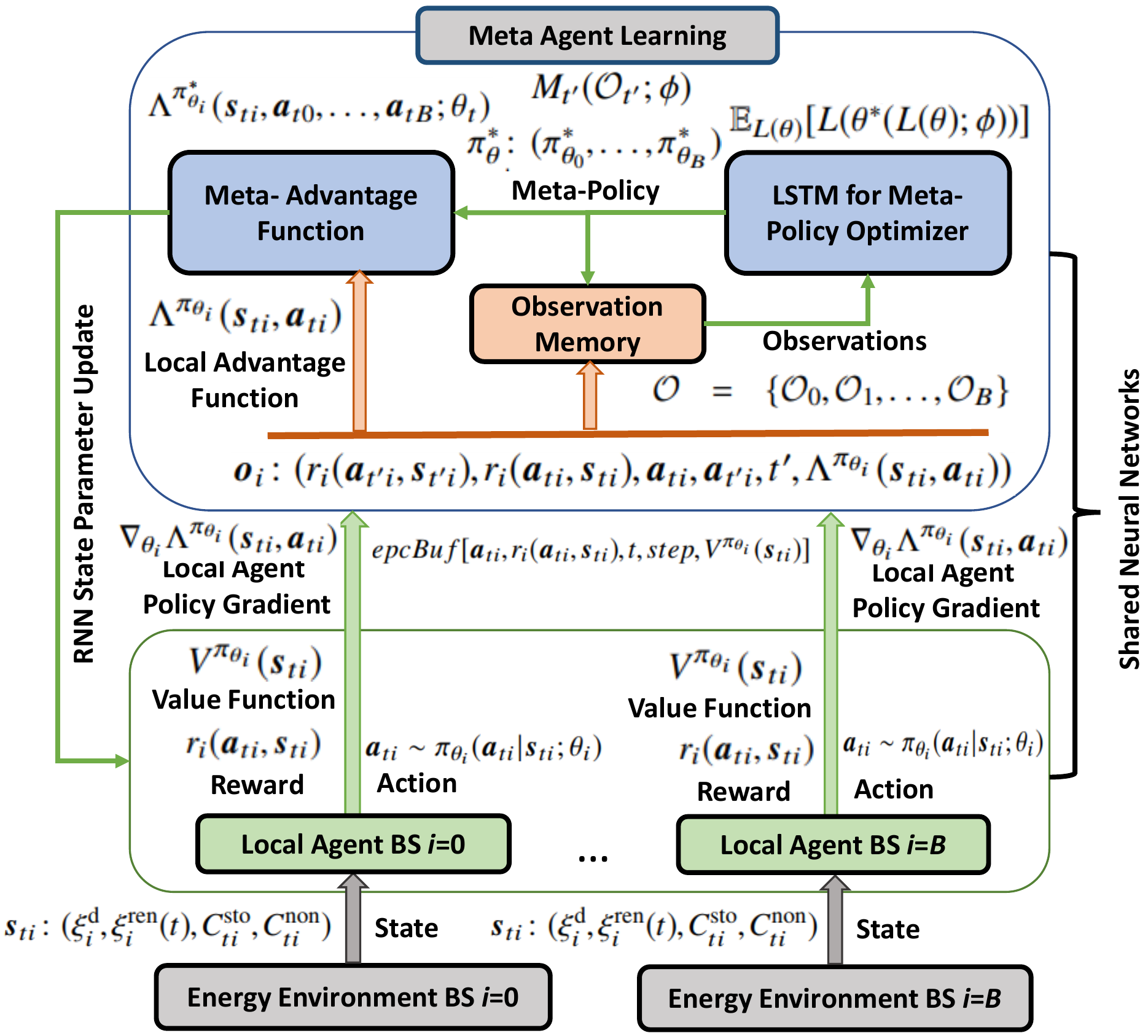}}
	\caption{Multi-agent meta-reinforcement learning framework.}
	\label{MAMRL_Training_Framework_Sol}
	\vspace{-4mm}
\end{figure}
\section{Energy Dispatch with Multi-Agent Meta-Reinforcement Learning Framework}
In this section, we developed our proposed multi-agent meta-reinforcement learning framework (as seen in Fig. \ref{MAMRL_Training_Framework_Sol}) for energy dispatch in the considered network. 
The proposed MAMRL framework includes two types of agents: A local agent that acts as a local learner at each self-powered with MEC capabilities BS and a meta-agent that learns the global energy dispatch policy. In particular, each local BS agent can discretize the Markovian dynamics for energy demand-generation of each BS (i.e., both SBSs and MBS) separately by applying deep-reinforcement learning. Meanwhile, we train a long short-term memory (LSTM) \cite{IEEEhowto:Hochreiter_LSTM,IEEEhowto:Fadlullah_DeepLearning} as a meta-agent at the MBS that optimizes \cite{IEEEhowto:Andrychowicz_L2L_by_GD} the accumulated energy dispatch of the local agents. As a result, the meta-agent can handle the non-i.i.d. energy demand-generation of the each local agent with own state information of the LSTM. To this end, MAMRL mitigates the curse of dimensionality for the uncertainty of energy demand and generation while providing an energy dispatch solution with a less computational and communication complexity (i.e., less message passing between the local agents and meta-agent).

\subsection{Preliminary Setup}
\begin{algorithm}[t!]
	\caption{State Space Generation of BS $i \in \mathcal{B}$ in MAMRL Framework}
	\label{alg:preprocessing_state_space_generation}
	\begin{algorithmic}[1]
		\renewcommand{\algorithmicrequire}{\textbf{Input:}}
		\renewcommand{\algorithmicensure}{\textbf{Output:}}
		\REQUIRE $W_{ij}, P_{i}, g_{ij}(t), \sigma^2,  I_{ij}(t), \Upsilon_{jk_i}(t), \tau, f_{k_i},\varpi_{k_{il}}, \eta^{\textrm{MEC}}_{\textrm{st}}(t), S_i(t)$
		\REQUIRE  $\delta_i^{\textrm{net}}, \eta^{\textrm{net}}_{\textrm{st}}(t), c^{\textrm{non}}_t, c^{\textrm{sto}}_t$ 
		\ENSURE $\boldsymbol{s}_{ti} \colon ( \xi_i^{\textrm{d}}, \xi_i^{\textrm{ren}}(t), C^{\textrm{sto}}_{ti}, C^{\textrm{non}}_{ti} ), \forall \boldsymbol{s}_{ti} \in \mathcal{S}_i \in \mathcal{S}, \forall t \in \mathcal{T}$
		\\ \textit{Initialization}: $\mathcal{R}_i, \mathcal{K}_i, \mathcal{J}_i$,$\mathcal{S}_i, \lambda_i (t), \mu_i(t), \rho_i(t), R_{i}(t)$
		\FOR {\textit{each} $t \in \mathcal{T}$} 
		\FOR {\textit{each} $i \in \mathcal{B}$} 
		\FOR {\textit{each} $j \in \mathcal{J}_i$}
		\STATE \textit{Calculate:} $\xi^{\textrm{MEC}}_i(t)$ using eq. \eqref{eq:cpu_energy}
		\STATE \textit{Calculate:} $\xi^{\textrm{net}}_i(t)$ using eq. \eqref{eq:comm_energy}
		\ENDFOR
		\STATE \textit{Calculate:} $\xi^{\textrm{d}}_t = \xi^{\textrm{net}}_i(t) +\xi^{\textrm{MEC}}_i(t)$ using eq. \eqref{eq:total_energy_demand}
		\STATE \textit{Calculate:} $\xi^{\textrm{ren}}_t= \sum_{q \in \mathcal{R}} \xi^{\textrm{ren}}_{iq}(t)$
		\STATE \textit{Calculate:}
		$C^{\textrm{non}}_t$ using eq. \eqref{eq:non-renewable_cost}
		\STATE \textit{Calculate:} $C^{\textrm{sto}}_t$ using eq. \eqref{eq:storage_cost}
		\STATE \textit{Assign:} $\boldsymbol{s}_{ti} \colon ( \xi_i^{\textrm{d}}, \xi_i^{\textrm{ren}}(t), C^{\textrm{sto}}_{ti}, C^{\textrm{non}}_{ti}) $
		\ENDFOR
		\STATE \textit{Append:} $\boldsymbol{s}_{ti} \in \mathcal{S}_i $
		\ENDFOR
		\RETURN $\forall \mathcal{S}_i \in \mathcal{S}$ 
	\end{algorithmic} 
\end{algorithm}
In the MAMRL setting, each BS $i \in \mathcal{B}$ acts as a local agent and the number of local agents is equal to $B+1$ BSs (i.e., $1$ MBS and $B$ SBSs). We define a set $\mathcal{S} = \left\{{\mathcal{S}_0,\mathcal{S}_1,\dots,\mathcal{S}_B}\right\}$ of state spaces and a set $\mathcal{A} = \left\{{\mathcal{A}_0,\mathcal{A}_1,\dots,\mathcal{A}_B}\right\}$ of actions for the $B+1$ agents. The state space of a local agent $i$ is defined by $\boldsymbol{s}_{ti} \colon ( \xi_i^{\textrm{d}}, \xi_i^{\textrm{ren}}(t), C^{\textrm{sto}}_{ti}, C^{\textrm{non}}_{ti}) \in \mathcal{S}_i$, where $\xi_i^{\textrm{d}}, \xi_i^{\textrm{ren}}(t),C^{\textrm{sto}}_{ti}$, and $C^{\textrm{non}}_{ti}$ represent the amount of energy demand, renewable generation, storage cost, and non-renewable energy cost, respectively, at time $t$. We execute Algorithm \ref{alg:preprocessing_state_space_generation} to generate the state space for every BSs $i \in \mathcal{B}$, individually. In Algorithm \ref{alg:preprocessing_state_space_generation}, lines $3$ to $6$ calculate the individual energy consumption of the MEC computation and network operation using \eqref{eq:cpu_energy} and \eqref{eq:comm_energy}, respectively. Overall, the energy demand of the BS $i$ is computed in line $7$ and the self-powered energy generation is estimated by line $8$ in Algorithm \ref{alg:preprocessing_state_space_generation}. Non-renewable and storage energy costs are calculated in lines $9$ and $10$ for time slot $t$. Finally, line $11$ creates state space tuple (i.e., $\boldsymbol{s}_{ti}$) for time $t$ in Algorithm \ref{alg:preprocessing_state_space_generation}. 

\subsection{Local Agent Design}
Consider each local BS agent $i \in \mathcal{B}$ that can take two types of actions $\xi^{\textrm{sto}}_i(t)$ and $\xi^{\textrm{non}}_i(t)$ which are the amount of storage energy $\xi^{\textrm{sto}}_i(t)$, and the amount of non-renewable energy $\xi^{\textrm{non}}_i(t)$ at time $t$.
We consider a discrete set of actions that consists of two actions $\boldsymbol{a}_{ti} \colon ( \xi^{\textrm{sto}}_i(t), \xi^{\textrm{non}}_i(t) ) \in \mathcal{A}_i$ for each BS unit $i \in \mathcal{B}$. Since the state $\boldsymbol{s}_{ti}$ and action $\boldsymbol{a}_{ti}$ both contain a time varying information of the agent $i \in \mathcal{B}$, we consider the dynamics of Markovian and represent problem $\eqref{Opt_1_8}$ as a discounted reward maximization problem for each agent $i$ (i.e., each BS). Thus, the objective function of the discounted reward maximization problem of agent $i$ is defined as follows \cite{IEEEhowto:Lowe_Multi_agent_actor_critic}:
\begin{equation} \label{eq:each_reward}
r_{i}(\boldsymbol{a}_{ti}, \boldsymbol{s}_{ti}) = \underset{\boldsymbol{a}_{ti} \in \mathcal{A}_i} \max \; \mathbb{E}_{\boldsymbol{a}_{ti} \sim \boldsymbol{s}_{ti}}[\sum_{t'=t}^{\infty} \gamma^{t'-t} \Upsilon_{t}(\boldsymbol{a}_{ti}, \boldsymbol{s}_{ti})],
\end{equation}
where $\gamma \in (0,1)$ is a discount factor and each reward $ \Upsilon_{t}\boldsymbol{a}_{ti}, \boldsymbol{s}_{ti})$ is considered as,
\begin{equation} \label{eq:each_reward_fn}
\Upsilon_{t}(\boldsymbol{a}_{ti}, \boldsymbol{s}_{ti}) = 
\left\{
\begin{array}{ll}
1 ,\;\text{if}\;\; \frac{\xi^{\textrm{ren}}_{ti}}{\xi^{\textrm{d}}_{ti}} > 1,\;\\
0,\;\;\;\;\;\;\;\;\;\;\;\;\;\;\text{otherwise}.
\end{array}
\right.
\end{equation}
In \eqref{eq:each_reward_fn}, $\frac{\xi^{\textrm{ren}}_{ti}}{\xi^{\textrm{d}}_{ti}}$ determines a ratio between renewable energy generation and energy demand (supply-demand ratio) of the BS agent $i \in \mathcal{B}$ at time $t$. When renewable energy generation-demand ratio $\frac{\xi^{\textrm{ren}}_{ti}}{\xi^{\textrm{d}}_{ti}}$ is larger than $1$ then the BS agent $i$ achieves a reward of $1$ because the amount of renewable energy exceeds the demand that can be stored in the storage unit.

Each action $\boldsymbol{a}_{ti}$ of BS agent $i \in \mathcal{B}$ determines a stochastic policy $\pi_{\theta_{i}}$. $\theta_{i}$ is a parameter of $\pi_{\theta_{i}}$ and the energy dispatch policy is defined by $\pi_{\theta_{i}} \colon \mathcal{S}_i \times \mathcal{A}_i \mapsto [0,1]$. Policy $\pi_{\theta_{i}}$ decides a state transition function $\Gamma \colon \mathcal{S}_i \times \mathcal{A}_B \mapsto \mathcal{S}_i$ for the next state $\boldsymbol{s}_{t'i} \in \mathcal{S}_i$. Thus, the state transition function $\Gamma$ of BS agent $i \in \mathcal{B}$ is determined by a reward function $\eqref{eq:each_reward_fn}$, where $\Upsilon_{t}(\boldsymbol{a}_{ti}, \boldsymbol{s}_{ti}) \colon \mathcal{S}_i \times \mathcal{A}_i \mapsto \mathbb{R}$. Further, each BS agent $i \in \mathcal{B}$ chooses an action $\boldsymbol{a}_{ti}$ from a parametrized energy dispatch policy $\pi_{\theta_{i}}(\boldsymbol{a}_{ti}| \boldsymbol{s}_{ti} ; \theta_{i})$. Therefore, for a given state $\boldsymbol{s}_{ti}$, the state value function with a cumulative discounted reward will be:
\begin{equation} \label{eq:each_state_value}
V^{\pi_{\theta_{i}}}(\boldsymbol{s}_{ti}) = \mathbb{E}_{\boldsymbol{a}_{ti} \sim \pi_{\theta_{i}}(\boldsymbol{a}_{ti}| \boldsymbol{s}_{ti} ; \theta_{i})}[\sum_{t'=t}^{\infty} \gamma^{t'-t} \Upsilon_{t+t'}(\boldsymbol{a}_{ti}, \boldsymbol{s}_{ti}) |\boldsymbol{s}_{ti}, \boldsymbol{a}_{ti}],
\end{equation}
where $\gamma^{t'-t}$ is a discount factor and ensures the convergence of state value function $V^{\pi_{\theta_{i}}}(\boldsymbol{s}_{ti})$ over the infinity time horizon. Thus, for a given state $\boldsymbol{s}_{ti}$, the optimal policy $\pi_{\theta_{i}}^*(\boldsymbol{a}_{ti}|\boldsymbol{s}_{ti})$ for the next state $\boldsymbol{s}_{t'i}$ can be determined by an optimal state value function while a Markovian property is imposed. Therefore, the optimal value function is given as follows:  
\begin{equation} \label{eq:opt_val_action_fn}
\begin{split}
V^{\pi_{\theta_{i}}^*}(\boldsymbol{s}_{ti})  = \;\;\;\;\;\;\;\;\;\;\;\;\;\;\;\;\;\;\;\;\;\;\;\;\;\;\;\;\;\;\;\;\;\;\;\;\;\;\;\;\;\;\;\;\;\;\;\;\;\;\;\;\;\;\;\;\;\;\;\; \\ \underset{a_{t} \in \mathcal{A}} \max \;\mathbb{E}_{\pi_{\theta_{i}}^*} [\sum_{i \in \mathcal{B}}r_{i}(\boldsymbol{a}_{t'i}, \boldsymbol{s}_{t'i}) + \sum_{t'=t}^{\infty} \gamma^{t'-t} V_{t'}^{\pi_{\theta_{i}}}(\boldsymbol{s}_{t'i}) |\boldsymbol{s}_{ti} ;\theta_{i}, \boldsymbol{a}_{ti}].
\end{split}
\end{equation}
Here, the optimal value function \eqref{eq:opt_val_action_fn} learns a parameterized policy $\pi_{\theta_{i}}(\boldsymbol{a}_{ti}| \boldsymbol{s}_{ti} ; \theta_{i})$ by using an LSTM-based Q-networks for the parameters $\theta_{i}$. Thus, each BS agent $i \in \mathcal{B}$ determines its parametrized energy dispatch policy $\pi_{\theta_{i}}(\boldsymbol{a}_{ti}| \boldsymbol{s}_{ti} ; \theta_{i}) = P(\boldsymbol{a}_{ti}| \boldsymbol{s}_{ti} ; \theta_{i})$, where $P(\xi^{\textrm{sto}}_i(t)) = P(\boldsymbol{a}_{ti} = \xi^{\textrm{sto}}_i(t)| \boldsymbol{s}_{ti} ; \theta_{i})$ and $P(\xi^{\textrm{non}}_i(t)) = 1-P(\xi^{\textrm{sto}}_i(t))$ for the parameters $\theta_{i}$. The decision of each BS agent $i \in \mathcal{B}$ relies on $\theta_{i}$. In particular, energy dispatch policy $\pi_{\theta_{i}}$ is the probability of taking action $\boldsymbol{a}_{ti}$ for a given state $\boldsymbol{s}_{ti}$ with parameters $\theta_{i}$.
In this setting, each local agent $i \in \mathcal{B}$ is comprised of an actor and a critic \cite{IEEEhowto:Mnih_async_a3c, IEEEhowto:single_agent_Sutton_RL_book}. The policy of energy dispatch is determined by choosing an action in (20) that can be seen as an actor of BS agent $i$. Meanwhile, the value function (19) is estimated by a critic of each local BS agent $i$. The critic can criticize actions that are made by the actor of each BS agent $i$. Therefore, each BS agent $i \in \mathcal{B}$ can determine a temporal difference (TD) error \cite{IEEEhowto:single_agent_Sutton_RL_book} based on the current energy dispatch policy of the actor and value estimation by the critic.
The TD error is considered as an advantage function and the advantage function of agent $i$ is defined as follows:
\begin{equation} \label{eq:TD_fn}
\begin{split}
\Lambda^{\pi_{\theta_{i}}}(\boldsymbol{s}_{ti},\boldsymbol{a}_{ti}) = \;\;\;\;\;\;\;\;\;\;\;\;\;\;\;\;\;\;\;\;\;\;\;\;\;\;\;\;\;\;\;\;\;\;\;\;\;\;\;\;\;\;\;\;\;\;\;\;\;\;\;\;\;\;\;\;\; \\  \Big(r_{i}(\boldsymbol{a}_{ti}, \boldsymbol{s}_{ti}) + \sum_{t'=t}^{\infty} \gamma^{t'-t} V^{\pi_{\theta_{i}}}(\boldsymbol{s}_{t'i}) \Big)-  V^{\pi_{\theta_{i}}}(\boldsymbol{s}_{ti}).
\end{split}
\end{equation}
Thus, the policy gradient of each BS agent $i \in \mathcal{B}$ is determined as,
\begin{equation} \label{eq:TD_fn_policy_gradient}
\begin{split}
\nabla_{\theta_{i}}\Lambda^{\pi_{\theta_{i}}}(\boldsymbol{s}_{ti},\boldsymbol{a}_{ti}) = \;\;\;\;\;\;\;\;\;\;\;\;\;\;\;\;\;\;\;\;\;\;\;\;\;\;\;\;\;\;\;\;\;\;\;\;\;\;\;\;\; \\  \mathbb{E}_{\pi_{\theta_{i}}}[\sum_{t'=t}^{\infty} \gamma^{t'-t} \nabla_{\theta_{i}} \log  \pi_{\theta_{i}}(\boldsymbol{a}_{ti}| \boldsymbol{s}_{ti} ; \theta_{i}) \Lambda^{\pi_{\theta_{i}}}(\boldsymbol{s}_{ti},\boldsymbol{a}_{ti}) ],
\end{split}
\end{equation}
where $\log \pi_{\theta_{i}}(\boldsymbol{a}_{ti}| \boldsymbol{s}_{ti} ; \theta_{i})$, and $\Lambda^{\pi_{\theta_{i}}}(\boldsymbol{s}_{ti},\boldsymbol{a}_{ti})$ represent the actor and critic, respectively, for each local BS agent $i \in \mathcal{B}$. 

Using \eqref{eq:TD_fn_policy_gradient}, we can discretize the energy dispatch decision $\boldsymbol{a}_{ti} \colon ( \xi^{\textrm{sto}}_i(t), \xi^{\textrm{non}}_i(t) )$ for each self-powered BS $i \in \mathcal{B}$ in the network.  
In fact, we can achieve a centralized solution for $\forall i \in \mathcal{B}$ when all of the BSs state information (i.e., demand and generation) are known. However, the space complexity for computation increases as $O(2^{|\mathcal{S}_i| \times |\mathcal{A}_i| \times |\mathcal{B}| \times T})$ and also the computational complexity becomes $O({|\mathcal{S}_i| \times |\mathcal{A}_i|\times |\mathcal{B}|^2}\times T)$ \cite{IEEEhowto:Zhang_Calibrated_Lr_Power_SBS}. Further, the solution does not meet the exploration-exploitation dilemma since the centralized (i.e., single agent) method ignores the interactions and energy dispatch decision strategies of other agents (i.e., BSs) which creates an imbalance between exploration and exploitation. In other words, this learning approach optimizes the action policy by exploring its own state information. 
Therefore, when we change the energy environment (i.e., demand and generation), this method cannot cope with an unknown environment due to the lack of diverse state information during the training.
Next, we propose an approach that not only reduces the complexity but also explores alternative energy dispatch decision to achieve the highest expected reward in \eqref{eq:each_reward}. 

\subsection{Multi-Agent Meta-Reinforcement Learning Modeling}
We consider a set $\mathcal{O} = \left\{{\mathcal{O}_0,\mathcal{O}_1,\dots,\mathcal{O}_B}\right\}$ of $B+1$ observations   \cite{IEEEhowto:Wang_Meta_RL, IEEEhowto:Littman_Markov_games} and for an BS agent $i \in \mathcal{B}$, a single observation tuple is given by $\boldsymbol{o}_{i} \in \mathcal{O}_i$. For a given state $\boldsymbol{s}_{ti}$, the observation $\boldsymbol{o}_{i}$ of the next state $\boldsymbol{s}_{t'i}$ consists of ${\boldsymbol{o}_{i} \colon ( r_{i}(\boldsymbol{a}_{t'i},\boldsymbol{s}_{t'i}), r_{i}(\boldsymbol{a}_{ti},\boldsymbol{s}_{ti}), \boldsymbol{a}_{ti}, \boldsymbol{a}_{t'i}, t',\Lambda^{\pi_{\theta_{i}}}(\boldsymbol{s}_{ti},\boldsymbol{a}_{ti}))}$, where $r_{i}(\boldsymbol{a}_{t'i},\boldsymbol{s}_{t'i})$, $r_{i}(\boldsymbol{a}_{ti},\boldsymbol{s}_{ti})$, $\boldsymbol{a}_{ti}, \boldsymbol{a}_{t'i}$, $t'$ and $\Lambda^{\pi_{\theta_{i}}}(\boldsymbol{s}_{ti},\boldsymbol{a}_{ti})$ are next-state discounted rewards, current state discounted rewards, next action, current action, time slot, and TD error, respectively. Here, a complete information of the observation $\boldsymbol{o}_{i}$ is correlated with the state space $\boldsymbol{o}_{i} \colon \mathcal{S}_i \mapsto \mathcal{O}_i$ while observation $\boldsymbol{o}_{i}$ does not require the complete state information of the previous states. 

Thus, the space complexity for computation at each BS agent $i \in \mathcal{B}$ leads to $O((|\mathcal{S}_i| + |\mathcal{A}_i|)^2 \times T)$. Meanwhile, the computational complexity for each time slot $t$ becomes $O( |\mathcal{S}_i|^2 \times \mathcal{A}_i \times \theta_{t} +H)$, where $\theta_{t}$ is the learning parameter and $H$ represents the numbers of LSTM units. Each BS agent $i \in \mathcal{B}$ requires to send an amount of $|\mathcal{O}_i|$ observational data (i.e., payload) to the meta-agent. Therefore, the communication overhead for each BS agent $i \in \mathcal{B}$ leads to $O(\frac{|\mathcal{O}| \times T}{B+1} )$. On the other hand, the computational complexity of the meta-agent leads to $O(|\mathcal{O}|^2 \times \phi + H)$ while $\phi$ represents learning parameter at meta-agent. In particular, for a fixed number of output memory $\phi$, the meta-agent's update complexity at each time slot $t$ becomes $O(\phi^2)$ \cite{IEEEhowto:Complexity_RNN}. Further, when transferring the learned parameters $\theta_{t'}$ from the meta-agent to all local agents $\forall i \in \mathcal{B}$, the communication overhead goes to the $O(\theta_{t'} \times (B+1))$ at each time slot $t$. Here, the size of $\theta_{t'}$ depends on the memory size of the LSTM cell at the meta-agent [see Appendix \ref{apd:example_information_exchange}].

In the MAMRL framework, the local agents work as an optimizee and the meta-agent performs the role of optimizer \cite{IEEEhowto:Andrychowicz_L2L_by_GD}. To model our meta-agent, we consider an LSTM architecture \cite{IEEEhowto:Hochreiter_LSTM, IEEEhowto:Fadlullah_DeepLearning} that stores its own state information (i.e., parameters) and the local agent (i.e., optimizee) only provides the observation of a current state. In the proposed MAMRL framework, a policy $\pi_{\theta_{i}}$ is determined by updating the parameters \footnote{We consider recurrent neural networks (RNNs) state parameters for the parameterization of energy dispatch policy. In particular, we consider a long short-term memory (LSTM) for RNN, in which cell state and hidden state are considered as parameters.} $\theta_{i}$. 
Therefore, we can represent the state value function \eqref{eq:opt_val_action_fn} for time $t$ is as follows: $V^{\pi_{\theta_{i}}^*}(\boldsymbol{s}_{ti})  \approx V^{\pi_{\theta_{i}}}(\boldsymbol{s}_{ti};\theta_{t})$, and the advantage (temporal difference) function \eqref{eq:TD_fn} is presented by, $\Lambda^{\pi_{\theta_{i}}}(\boldsymbol{s}_{ti},\boldsymbol{a}_{ti})  \approx \Lambda^{\pi_{\theta_{i}}}(\boldsymbol{s}_{ti},\boldsymbol{a}_{ti}; \theta_{t})$. As a result, the parameterized policy is defined by, $\pi_{\theta_{i}}(\boldsymbol{a}_{ti}|\boldsymbol{s}_{ti}) \approx \pi_{\theta_{i}}(\boldsymbol{a}_{ti}|\boldsymbol{s}_{ti};\theta_{t})$.
Considering all of the BS agents $B+1$ and the advantage function $\eqref{eq:TD_fn}$ is rewritten as,
\begin{equation} \label{eq:game_A_function}
\begin{split}
\Lambda^{\pi_{\theta_{i}}^*}(\boldsymbol{s}_{ti}, \boldsymbol{a}_{t0}, \dots,\boldsymbol{a}_{tB} ; \theta_{t})  = 
r_{i}(\boldsymbol{s}_{ti}, \boldsymbol{a}_{t0}, \dots, \boldsymbol{a}_{tB}) \; + \;\;\;\;\;\;\;\;\;\;\;\;\;\;\;\; \\ \sum_{\boldsymbol{s}_{t'i} \in \mathcal{S}_i, t'=t}^{\infty} \gamma^{t'-t} \Gamma(\boldsymbol{s}_{t'i}|\boldsymbol{s}_{ti}, \boldsymbol{a}_{t0}, \dots,\boldsymbol{a}_{tB}) V^{\pi_{\theta_{i}}}(\boldsymbol{s}_{t'i}, \pi_{\theta_{0}}^*,\dots, \pi_{\theta_{B}}^* )\;-\\
V^{\pi_{\theta_{i}}}(\boldsymbol{s}_{ti}, \pi_{\theta_{0}}^*,\dots, \pi_{\theta_{B}}^* ), \;\;\;\;\;\;\;\;\;\;\;\;\;\;\;\;\;\;\;\;\;\;\;\;\;\;\;\;\;\;\;\; \;\;\;\;\;\;\;\;\;\;\;\;\;\;\;\;
\end{split}
\end{equation}      
where $\pi_{\theta}^* \colon (\pi_{\theta_{0}}^*,\dots, \pi_{\theta_{B}}^*)$ is a joint energy dispatch policy and  $ \Gamma(\boldsymbol{s}_{t'i}|\boldsymbol{s}_{ti}, \boldsymbol{a}_{t0}, \dots,\boldsymbol{a}_{tB}) \mapsto [0,1]$ represents state transition probability. Using \eqref{eq:game_A_function}, we can get the value loss function for agent $i$ and the objective is to minimize the temporal difference \cite{IEEEhowto:Mnih_async_a3c},
\begin{equation} \label{eq:loss_fn_dqn}
\begin{split}
L(\theta_{i}) = \;\;\;\;\;\;\;\;\;\;\;\;\;\;\;\;\;\;\;\;\;\;\;\; \;\;\;\;\;\;\;\;\;\;\;\;\;\;\;\;\;\;\;\;\;\;\;\;\;\;\;\;\;\;\; \;\;\;\;\;\;\;\;\;\;\;\;\;\;\;\;\;\;\;\;\;\;\;\;\;\;\;\;\;\\ \underset{\pi_{\theta_{i}}}\min \; \frac{1}{|\mathcal{B}|} \sum_{i \in \mathcal{B}}\frac{1}{2} \bigg(\Big(r_{i}(\boldsymbol{a}_{ti}, \boldsymbol{s}_{ti}) + \sum_{t'=t}^{\infty} \gamma^{t'-t} V^{\pi_{\theta_{i}}^*}(\boldsymbol{s}_{t'i}|\theta_{t}) \Big)-  V^{\pi_{\theta_{i}}^*}(\boldsymbol{s}_{ti})\bigg)^2.
\end{split}
\end{equation}

To improve the exploration with a low bias, we consider an entropy regularization \footnote{Entropy \cite{IEEEhowto:entropy_Uncertainty_Birula,IEEEhowto:entropy_Seidenfeld_uncertainty, IEEEhowto:entropy_Feyzmahdavian_Asynchronous, IEEEhowto:entropy_Agarwal_Online} can allow us to manage non-i.i.d. datasets when changes in the environment over time lead to an uncertainty. Therefore, we use entropy regularization to handle the non-i.i.d. energy demand and generation over time by managing with the uncertainty for each BS agent $i \in \mathcal{B}$.} $\beta h(\pi_{\theta_{i}}(\boldsymbol{a}_{ti}|\boldsymbol{s}_{ti};\theta_{t}))$ that cope with the non-i.i.d. energy demand and generation for all of the BS agents $\forall i \in \mathcal{B}$. Here, $\beta$ is a coefficient for the magnitude of regularization and $h(\pi_{\theta_{i}}(\boldsymbol{a}_{ti}|\boldsymbol{s}_{ti};\theta_{t}))$ determines the entropy of the policy $\pi_{\theta_{i}}$ for the parameter $\theta_{i}$. Additionally, a larger value of $\beta h(\pi_{\theta_{i}}(\boldsymbol{a}_{ti}|\boldsymbol{s}_{ti};\theta_{t}))$ encourages the agents to have a more diverse exploration to estimate the energy dispatch policy. Thus, we can redefine the policy loss function as follows:
\begin{equation} \label{eq:entroyp_centralized_action_value}
\begin{split}
L(\theta_{i})  =  -\mathbb{E}_{\boldsymbol{s}_{ti}, \boldsymbol{a}_{ti}}[\pi_{\theta_{i}}(\boldsymbol{a}_{ti}|\boldsymbol{s}_{ti}) + \beta h(\pi_{\theta_{i}}(\boldsymbol{a}_{ti}|\boldsymbol{s}_{ti};\theta_{t}))].
\end{split}
\end{equation}
Therefore, the policy gradient of the loss function \eqref{eq:entroyp_centralized_action_value} is defined in terms of temporal difference and entropy. The policy gradient of the loss function is defined as follows:
\begin{equation} \label{eq:loss_gradient_TD}
\begin{split}
\nabla_{\theta_{i}}L(\theta_{i}) =  \frac{1}{|\mathcal{B}|} \sum_{i \in \mathcal{B}} \sum_{t'=t}^{\infty} \nabla_{\theta_{i}} \log  \pi_{\theta_{i}}(\boldsymbol{a}_{ti}|\boldsymbol{s}_{ti}) \Lambda^{\pi_{\theta_{i}}}(\boldsymbol{s}_{ti},\boldsymbol{a}_{ti}|\theta_{t}) \\ + \; \beta h(\pi_{\theta_{i}}(\boldsymbol{a}_{ti}|\boldsymbol{s}_{ti};\theta_{t})). \;\;\;\;\;\;\;\;\;\;\;\;\;\;\;\;\;\;\;\;\;\;\;\;\;\;\;\;\;\;\;\;\;\;\;\;\;\;\;\
\end{split}
\end{equation}

To design our meta-agent, we consider meta-agent parameters $\phi$ and optimized parameters $\theta^*$ of the optimizee (i.e., local agent). The meta-agent is defined as $M_t(\mathcal{O}_t; \phi) \coloneqq M_t(\nabla_{\theta_{t}}L(\theta_{t}); \phi)$, where $M_t(.)$ is modeled by an LSTM. 
Consider an observational vector $\mathcal{O}_{it'} \in \mathcal{O}$ of a local BS agent $i \in \mathcal{B}$ at time $t'$ and each observation is ${\boldsymbol{o}_{i} \colon ( r_{i}(\boldsymbol{a}_{t'i},\boldsymbol{s}_{t'i}), r_{i}(\boldsymbol{a}_{ti},\boldsymbol{s}_{ti}), \boldsymbol{a}_{ti}, \boldsymbol{a}_{t'i}, t',\Lambda^{\pi_{\theta_{i}}}(\boldsymbol{s}_{ti},\boldsymbol{a}_{ti}))} \in \mathcal{O}_{it'} $. The LSTM-based meta-agent takes the observational vector $\mathcal{O}_{it'}$ as an input. Meanwhile, the meta-agent holds long-term dependencies by generating its own state with parameters $\phi$. To do this, the LSTM model creates several gates to determine an optimal policy $\pi_{\theta_{i}}^*$ and advantage function $\Lambda^{\pi_{\theta_{i}}^*}(\boldsymbol{s}_{ti}, \boldsymbol{a}_{t0}, \dots,\boldsymbol{a}_{tB} ; \theta_{t})$ for the next state $\boldsymbol{s}_{t'i}$. 
As a result, the structure of the recurrent neural network for the meta-agent is the same as the LSTM model \cite{IEEEhowto:Hochreiter_LSTM, IEEEhowto:Fadlullah_DeepLearning}.
In particular, each LSTM unit for the meta-agent consists of four gate layers such as forget gate $\boldsymbol{F}_{t'}$, input gate $\boldsymbol{I}_{t'}$, cell state $\boldsymbol{\hat{E}}_{t'}$, and output $\boldsymbol{Z}_{t'}$ layer. The cell state gate $\boldsymbol{\hat{E}}_{t'}$ usages a $\tanh$ activation function and other gates are used sigmoid $\sigma(.)$ as an activation function. Thus, the outcome of the meta policy for a single unit LSTM cell is presented as follows:
\begin{subequations}\label{meta_agent_model}
	\begin{align}
	M_{t'}(\mathcal{O}_{t'}; \phi) = \text{softmax}\Big( (\boldsymbol{H}_{t'})^{\top} \Big),      \tag{\ref{meta_agent_model}} \;\;\;\;\;\;\;\;\;\;\;\;\;\;\;\;\;\;\\
	\text{where} \quad \label{meta_agent_model:forget}  \boldsymbol{F}_{t'} = \sigma\Big(\phi_{FO}(\mathcal{O}_{it'})^{\top} +\phi_{FH}(\boldsymbol{H}_{t})^{\top} + \boldsymbol{b}_F\Big) , \;\;\;\;\\
	\label{meta_agent_model:input}  \boldsymbol{I_{t'}} = \sigma\Big(\phi_{IO}(\mathcal{O}_{it'})^{\top} +\phi_{IH}(\boldsymbol{H}_{t})^{\top} + \boldsymbol{b}_I\Big) , \;\;\;\;\;\;\;\;\\
	\label{meta_agent_model:state}  \boldsymbol{\hat{E_{t'}}} = \tanh\Big(\phi_{EO}(\mathcal{O}_{it'})^{\top} +\phi_{EH}(\boldsymbol{H}_{t})^{\top} + \boldsymbol{b}_E\Big) , \\
	\label{meta_agent_model:state_main} \boldsymbol{E}_{t'} = \boldsymbol{\hat{E}_{t'}} \odot \boldsymbol{I}_{t'} + \boldsymbol{F}_{t'} \odot \boldsymbol{E}_{t}, \;\;\;\;\;\;\;\;\;\;\;\;\;\;\;\;\;\;\;\;\;\;\;\;\;\\
	\label{meta_agent_model:output}  \boldsymbol{Z}_{t'} = \sigma \Big(\phi_{ZO}(\mathcal{O}_{it'})^{\top} +\phi_{ZH}(\boldsymbol{H}_{t})^{\top} + \boldsymbol{b}_Z\Big) , \;\;\;\;\;\\
	\label{meta_agent_model:H}  \boldsymbol{H}_{t'} = \tanh(\boldsymbol{E}_{t'}) \odot \boldsymbol{Z}_{t'}.\;\;\;\;\;\;\;\;\;\;\;\;\;\;\;\;\;\;\;\;\;\;\;\;\;\;\;\;\;\;\;
	\end{align}
\end{subequations}
\begin{figure*}[!t]
	\centerline{\includegraphics[width=\textwidth]{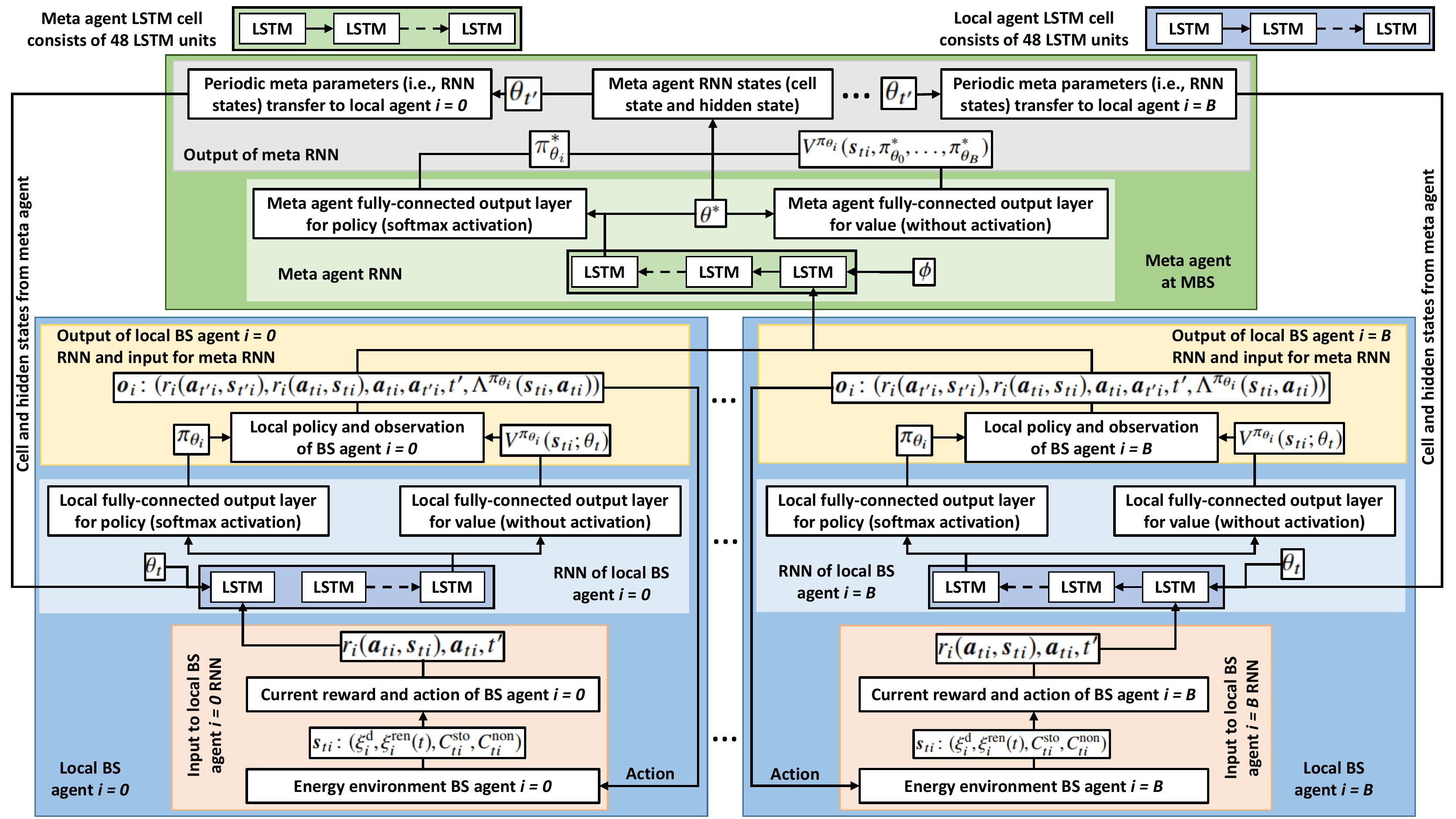}}
	\caption{Recurrent neural network architecture for the proposed multi-agent meta-reinforcement learning framework.}
	\label{architecture_of_the_neural_network}
	\vspace{-4mm}
\end{figure*}
In the meta-agent policy formulation \eqref{meta_agent_model}, the forget gate vector \eqref{meta_agent_model:forget} determines what information is needed to throw away. Input gate vector \eqref{meta_agent_model:input} helps to decide which information is needed to update, the cell state \eqref{meta_agent_model:state} creates a vector of new candidate values using $\tanh(\cdot)$ function, and updates the cell state information by applying \eqref{meta_agent_model:state_main}. The output layer \eqref{meta_agent_model:output} that determines what parts of the cell state are going to output and calculate the cell outputs using the equation $\eqref{meta_agent_model:H}$. Further, the cell state through the $\tanh(\cdot)$ will restrict the values between $-1$ and $+1$. This entire process is followed for each LSTM block and finally, $\eqref{meta_agent_model}$ determines the meta-policy for $\pi_{\theta_{i}}^*$ of the state $s_{t'}$. In addition, optimized RNN state parameters $\theta^*$ are obtained from the cell state \eqref{meta_agent_model:state_main} and hidden state  $\eqref{meta_agent_model:H}$ of an LSTM unit. Thus, the loss function $L(\phi) =  \mathbb{E}_{L(\theta)}[L(\theta^*(L(\theta);\phi))]$ of meta-agent depends on the distribution of $L(\theta_{t})$ and the expectation of the meta-agent loss function is defined as follows \cite{IEEEhowto:Andrychowicz_L2L_by_GD}:    
\begin{equation} \label{eq:loss_fn_meta}
\begin{split}
L(\phi) =  \mathbb{E}_{L(\theta)}[\sum_{t = 1}^T L(\theta_t)].
\end{split}
\end{equation}
In the proposed MAMRL framework, we transfer the learned parameters (i.e., cell state and hidden state) of meta-agent to the local agents so that each local agent will be estimated an optimal energy dispatch policy by updating its own learning parameters.
Thus, the parameters of each agent (i.e., BS) is updated with $\theta_{t'}= \theta^*$ while $\pi_{\theta_{i}}^*= M_t(\nabla_{\theta_{t}}L(\theta_{t}) ; \phi)$ to decide the energy dispatch policy.


We consider an LSTM-based recurrent neural network (RNN) for the both local agents and the meta-agent. This LSTM RNN consists of $48$ LSTM units for each LSTM cell as shown in Fig. \ref{architecture_of_the_neural_network}. In particular, the configuration of the LSTM for the meta-agent and each local agent is the same while the objective of the loss functions differ from local agent to meta-agent. In which, local BS agent determines its own energy dispatch policy by exploring its own environmental state information for reducing the TD error. Meanwhile, meta-agent deals with the observations of each local BS agent by exploiting its own RNN states information using entropy based loss function to capture non-i.i.d. energy demand and generation of each local BS. Therefore, having different loss functions for local and meta agent leads the proposed MAMRL model to learn a domain specific generalized model so that it can cope with an unknown environment. 
Further, this RNN consists of a branch of two fully connected output layers on top of the LSTM cell. In particular, fully connected layer with a softmax activation is considered for energy dispatch policy determination, and another fully connected output layer without activation function is deployed for value function estimation. Thus, the advantage is calculated based on value function estimation from the second fully connected layer.
Each local LSTM-based RNN receives a current reward $r_{i}(\boldsymbol{a}_{ti}$, $\boldsymbol{s}_{ti})$, current action $\boldsymbol{a}_{ti}$, and next time slot $t'$ as an input for each BS agent $i \in \mathcal{B}$. Meanwhile, this local LSTM model estimates a policy $\pi_{\theta_{i}}$ and value $ V^{\pi_{\theta_{i}}}(\boldsymbol{s}_{ti})$ for BS agent $i \in \mathcal{B}$.  On the other hand, meta agent LSTM-based RNN feeds input as an observational tuple ${\boldsymbol{o}_{i} \colon (r_{i}(\boldsymbol{a}_{t'i},\boldsymbol{s}_{t'i}), r_{i}(\boldsymbol{a}_{ti},\boldsymbol{s}_{ti}), \boldsymbol{a}_{ti}, \boldsymbol{a}_{t'i}, t',\Lambda^{\pi_{\theta_{i}}}(\boldsymbol{s}_{ti},\boldsymbol{a}_{ti})) }$ from each BS agent $i \in \mathcal{B}$. This observation consists of the current and next reward, current and next action, next time slot, and TD error for each BS agent $i$. Thus, this meta agent estimates parameters $\theta_{t'}$ to find a globally optimal energy dispatch policy $\pi_{\theta_{i}}^*$ for each BS $i \in \mathcal{B}$.
The learned parameters of the meta-agent are transferred to each local BS agent $i \in \mathcal{B}$ asynchronously while this local agent updates its own parameters for estimating the globally optimal energy dispatch policy via the local LSTM-based RNN. In particular, the learned parameters (i.e., RNN states) are transfered from meta-agent to each local agent $i \in \mathcal{B}$. Additionally, these RNN state parameters include cell state and hidden state of the LSTM cell, which do not depend on any of the fully connected out layers of the proposed RNN architecture. Meanwhile, each local agent $i \in \mathcal{B}$ updates its own RNN states using the transferred parameters by the meta-agent. We consider a cellular network for exchanging observations and parameters between local BS agent and meta-agent. 

\begin{algorithm}[t!]
	\caption{Local Agent Training of Energy Dispatch of BS $i \in \mathcal{B}$ in MAMRL Framework}
	\label{alg:Local_agent_algo}
	\begin{algorithmic}[1]
		\renewcommand{\algorithmicrequire}{\textbf{Input:}}
		\renewcommand{\algorithmicensure}{\textbf{Output:}}
		\REQUIRE  $\boldsymbol{s}_{ti} \colon ( \xi_i^{\textrm{d}}, \xi_i^{\textrm{ren}}(t), C^{\textrm{sto}}_{ti}, C^{\textrm{non}}_{ti} ), \forall \boldsymbol{s}_{ti} \in \mathcal{S}_i, \forall t \in T$ 
		\ENSURE  ${\boldsymbol{o}_{i} \colon ( r_{i}(\boldsymbol{a}_{t'i},\boldsymbol{s}_{t'i}), r_{i}(\boldsymbol{a}_{ti},\boldsymbol{s}_{ti}), \boldsymbol{a}_{ti}, \boldsymbol{a}_{t'i}, t',\Lambda^{\pi_{\theta_{i}}}(\boldsymbol{s}_{ti},\boldsymbol{a}_{ti}))}$, $\boldsymbol{o}_{i} \in \mathcal{O}_i$, $\nabla_{\theta_{t}}L(\theta_{t})$
		\\ \textit{Initialization:} $LocalLSTM(.)$, $\theta_{i}, i \in \mathcal{B}, \gamma, \mathcal{O}_i$
		\FOR {\textit{episode = 1} to \textit{maximum episodes}} 
		\STATE \textit{Initialization:} $epcBuf[]$
		\FOR {\textit{each} $t \in \mathcal{T}$}
		\FOR {$step = 1$ to $MaxStep$}
		\STATE \textit{Calculate:} $r_{i}(\boldsymbol{a}_{ti}, \boldsymbol{s}_{ti})$ using eq. \eqref{eq:each_reward}
		\STATE \textit{Calculate:} $V^{\pi_{\theta_{i}}}(\boldsymbol{s}_{ti})$ using eq. \eqref{eq:each_state_value}
		\STATE \textit{Choose Action:}  $\boldsymbol{a}_{ti} \sim \pi_{\theta_{i}}(\boldsymbol{a}_{ti}| \boldsymbol{s}_{ti})$
		\STATE \textit{Append:} $epcBuf[\boldsymbol{a}_{ti}, r_{i}(\boldsymbol{a}_{ti}, \boldsymbol{s}_{ti}), t, step, V^{\pi_{\theta_{i}}}(\boldsymbol{s}_{ti}) ]$
		\ENDFOR
		\STATE $LocalLSTM(r_{i}(\boldsymbol{a}_{ti},\boldsymbol{s}_{ti}), \boldsymbol{a}_{ti}, t' = t + 1)$ \\ \algorithmiccomment{LSTM-based local RNN block}
		\STATE\{
		\STATE \textit{Evaluate:} $\Lambda^{\pi_{\theta_{i}}}(\boldsymbol{s}_{ti},\boldsymbol{a}_{ti})$ using eq. \eqref{eq:TD_fn}
		\STATE \textit{Local agent policy gradient:} $\nabla_{\theta_{i}}\Lambda^{\pi_{\theta_{i}}}(\boldsymbol{s}_{ti},\boldsymbol{a}_{ti})$ using eq. \eqref{eq:TD_fn_policy_gradient} \\	\algorithmiccomment{In \eqref{eq:TD_fn_policy_gradient}, $\pi_{\theta_{i}}(\boldsymbol{a}_{ti}| \boldsymbol{s}_{ti} ; \theta_{i})$ is determined by a fully connected output layer with a softmax activation function and  $\Lambda^{\pi_{\theta_{i}}}(\boldsymbol{s}_{ti},\boldsymbol{a}_{ti})$ is calculated through a fully connected output layer without activation function}
		\STATE\}
		\STATE \textit{Append:} \\$\boldsymbol{o}_{i} \colon ( r_{i}(\boldsymbol{a}_{t'i},\boldsymbol{s}_{t'i}), r_{i}(\boldsymbol{a}_{ti},\boldsymbol{s}_{ti}), \boldsymbol{a}_{ti}, \boldsymbol{a}_{t'i}, t',\Lambda^{\pi_{\theta_{i}}}(\boldsymbol{s}_{ti},\boldsymbol{a}_{ti}))$,
		$\boldsymbol{o}_{i} \in \mathcal{O}_i$
		\STATE \textit{Get Meta-agent policy $\pi_{\theta_{i}}^*$ and RNN states $\theta^*$:} $M_t(\mathcal{O}_t; \phi)$ using Algorithm \ref{alg:MAMRL_algo}
		\STATE Update: $\theta_{t'}= \theta^*$ \algorithmiccomment{RNN states update}
		\ENDFOR
		\ENDFOR
		\RETURN \textit{new\_state}$(\boldsymbol{s}_{t'i} = \argmax_{\pi_{\theta_{i}}^*} (\boldsymbol{a}_{ti})), i \in \mathcal{B}$ 
	\end{algorithmic} 
\end{algorithm}  

We run the proposed Algorithm \ref{alg:Local_agent_algo} at each self-powered BS $i \in \mathcal{B}$ with MEC capabilities as local agent $i$. The input of Algorithm \ref{alg:Local_agent_algo} is the state information $\mathcal{S}_i$ of local agent $i$, which is the output from  Algorithm \ref{alg:preprocessing_state_space_generation}. The cumulative discounted reward \eqref{eq:each_reward} and state value in \eqref{eq:each_state_value} are calculated in lines $5$ and $6$, respectively (in Algorithm \ref{alg:Local_agent_algo}) for each step (until the maximum step size \footnote{To capture the heterogeneity for energy demand and generation of each BS separately, we consider the same number of user tasks that are executed by each BS agent $i \in \mathcal{B}$ during one observational period $t$ as the steps size.} for time step $t$). Consequently, based on a chosen action $\boldsymbol{a}_{ti}$ from the estimated policy $\pi_{\theta_{i}}(\boldsymbol{a}_{ti}| \boldsymbol{s}_{ti})$ (in line $7$), episode buffer is generated and appended in line $8$. Advantage function \eqref{eq:TD_fn} of local agent $i$ is evaluated in line $12$ and the policy gradient \eqref{eq:TD_fn_policy_gradient} is calculated in line $13$ using an LSTM-based local neural network. Algorithm \ref{alg:Local_agent_algo} generates observational tuple $\boldsymbol{o}_{i} \colon ( r_{i}(\boldsymbol{a}_{t'i},\boldsymbol{s}_{t'i}), r_{i}(\boldsymbol{a}_{ti},\boldsymbol{s}_{ti}), \boldsymbol{a}_{ti}, \boldsymbol{a}_{t'i}, t',\Lambda^{\pi_{\theta_{i}}}(\boldsymbol{s}_{ti},\boldsymbol{a}_{ti}))$ in line $15$. Here, we transfer the knowledge of local BS agent $i \in \mathcal{B}$ to the meta-agent learner (deployed in MBS) in Algorithm \ref{alg:MAMRL_algo} so as to optimize the energy dispatch decision (in Algorithm \ref{alg:Local_agent_algo} line $16$). Hence, the observation tuple $\boldsymbol{o}_{i}$ of local BS agent $i$ consists of only the decision from BS $i$, where does not require to send all of the state information to meta-agent learner. Employing the meta-agent policy gradient, each local agent is capable of updating the energy dispatch decision policy in line $17$ in Algorithm \ref{alg:Local_agent_algo}. Finally, the energy dispatch policy is executed in line $20$ at the BS $i\in \mathcal{B}$ by local agent $i$.  
\begin{algorithm}[t!]
	\caption{Meta-Agent Learner of Energy Dispatch in MAMRL Framework}
	\label{alg:MAMRL_algo}
	\begin{algorithmic}[1]
		\renewcommand{\algorithmicrequire}{\textbf{Input:}}
		\renewcommand{\algorithmicensure}{\textbf{Output:}}
		\REQUIRE  ${\boldsymbol{o}_{i} \colon ( r_{i}(\boldsymbol{a}_{t'i},\boldsymbol{s}_{t'i}), r_{i}(\boldsymbol{a}_{ti},\boldsymbol{s}_{ti}), \boldsymbol{a}_{ti}, \boldsymbol{a}_{t'i}, t',\Lambda^{\pi_{\theta_{i}}}(\boldsymbol{s}_{ti},\boldsymbol{a}_{ti}))},$
		$\forall \boldsymbol{o}_{i} \in \mathcal{O}_i$, $t \in \mathcal{T}$, $i \in \mathcal{B}$ 
		\ENSURE  $\phi$
		\\ \textit{Initialization:} $MetaLSTM(.)$, $\phi$, $\pi_{\theta_{i}}$, $\gamma$
		\FOR {\textit{each} $t \in \mathcal{T}$}
		\FOR {\textit{each} $i \in \mathcal{B}$}
		\STATE ${\boldsymbol{o}_{i} \colon ( r_{i}(\boldsymbol{a}_{t'i},\boldsymbol{s}_{t'i}), r_{i}(\boldsymbol{a}_{ti},\boldsymbol{s}_{ti}), \boldsymbol{a}_{ti}, \boldsymbol{a}_{t'i}, t',\Lambda^{\pi_{\theta_{i}}}(\boldsymbol{s}_{ti},\boldsymbol{a}_{ti}))},$
		$\boldsymbol{o}_{i} \in \mathcal{O}_i$
		\STATE $MetaLSTM(\mathcal{O}_i,\pi_{\theta_{i}})$ \algorithmiccomment{LSTM-based RNN block}
		\STATE\{ \\
		\algorithmiccomment{Lines from $6$ to $7$ using fully connected output layer without activation function}
		\STATE \textit{Entropy loss:} $L(\theta_{i})$ using eq. \eqref{eq:entroyp_centralized_action_value}
		\STATE \textit{Gradient of the loss:} $\nabla_{\theta_{t}}L(\theta_{t})$ using eq. \eqref{eq:loss_gradient_TD} \\
		\algorithmiccomment{Policy is estimated using a fully connected output layer with softmax activation function}
		\STATE \textit{Calculate:} $\pi_{\theta_{i}} = M_t(\nabla_{\theta_{t}}L(\theta_{t}) ; \phi)$ using eq. \eqref{meta_agent_model}
		\STATE \textit{Get meta policy loss} $L(\phi)$ using eq. \eqref{eq:loss_fn_meta}
		\STATE \textit{Update:} $\pi_{\theta_{i}}^* = \pi_{\theta_{i}}$
		\STATE \textit{Get RNN states:} $\theta^*$ 	\\	\algorithmiccomment{cell state and hidden state from the LSTM cell}
		\STATE\}
		\ENDFOR
		\STATE \textit{Send:} Meta-agent policy $\pi_{\theta_{i}}^*$ and RNN states $\theta^*$
		\ENDFOR
		\RETURN 
	\end{algorithmic} 
\end{algorithm}

The meta-agent learner (Algorithm \ref{alg:MAMRL_algo} in MBS) receives the observations $ \mathcal{O}_i \in \mathcal{O}$ from each local BS agent $i \in \mathcal{B}$ asynchronously. Then the meta-agent asynchronously updates the meta policy gradient of the each BS agent $i \in \mathcal{B}$. Lines from $4$ to $12$ of Algorithm \ref{alg:MAMRL_algo} represent the LSTM block for the meta-agent. In Algorithm \ref{alg:MAMRL_algo}, entropy loss \eqref{eq:entroyp_centralized_action_value} and gradient of the loss \eqref{eq:loss_gradient_TD} are estimated in lines $6$ and $7$, respectively. In order to estimate this, Algorithm \ref{alg:MAMRL_algo} deploys a fully connected output layer without activation function, so that advantage loss can be calculated without affecting the value that is calculated by the value function of the proposed MAMRL framework. The meta-agent energy dispatch policy is updated in line $10$ of Algorithm \ref{alg:MAMRL_algo}. Before that, a fully connected output layer with a softmax activation function of the LSTM cell assists to determine the energy dispatch policy and meta policy loss in lines $8$ and $9$ (in Algorithm \ref{alg:MAMRL_algo}), respectively, for the meta-agent. Additionally, the meta-agent utilizes the observations of the local agents and determines its own state information that helps to estimate the energy dispatch policy of the meta-agent. In line $11$, the meta-agent RNN states $\theta^*$ (i.e., cell and hidden states) are received from the considered LSTM cell in Algorithm \ref{alg:MAMRL_algo}. Finally, the meta-agent policy and RNN states are transfered to each BS agent for updating the parameters (i.e., RNN states) of each local BS agent. To this end, a meta-agent learner deployed at center node (i.e., MBS) in the considered network and sends the learning parameters of the optimal energy dispatch policy to each local BS (i.e., MBS and SBS) through the network.   

The proposed MAMRL framework established a guarantee to converge with an optimal energy dispatch policy. In fact, the MAMRL framework can be reduced to a $|\mathcal{B}|$-player Markovian game \cite{IEEEhowto: Fink_Equilibrium_Th1,IEEEhowto:Herings_Equilibrium_Th1_proof} as a base problem that establishes more insight into convergence and optimality. The proposed MAMRL model has at least one Nash equilibrium point that ensures an optimal energy dispatch policy. This argument is similar from the previous studies of $|\mathcal{B}|$-player Markovian game \cite{IEEEhowto: Fink_Equilibrium_Th1,IEEEhowto:Herings_Equilibrium_Th1_proof}. Hence, we can conclude with the following proposition: 
\begin{proposition}
	\label{pro:discounted_reward_game_Nash_equilibrium_point}
	$\pi_{\theta_{i}}^*$ is an optimal energy dispatch policy that is an equilibrium point with an equilibrium value $V^{\pi_{\theta_{i}}}(\boldsymbol{s}_{ti}, \pi_{\theta_{0}}^*,\dots, \pi_{\theta_{B}}^* )$ for BS $i$  [see Appendix \ref{apd:discounted_reward_game_Nash_equilibrium_point_proof}].
\end{proposition}
We can justify the convergence of MAMRL framework via the following Proposition: 
\begin{proposition}
	\label{pro:Convergence_Multi_Agent_Meta_RL}
	Consider a stochastic environment with a state space $\boldsymbol{s}_{ti} \in \mathcal{S}, i \in \forall \mathcal{B}$ of $|\mathcal{B}|$ BS agents such that all BS agents are initialized with an equal probability of $0.5$ for a binary actions, $P(\xi^{\textrm{sto}}_i(t)) = P(\xi^{\textrm{non}}_i(t)) = \theta_{i} \approx 0.5$, where $\boldsymbol{a}_{ti} \colon ( \xi^{\textrm{sto}}_i(t), \xi^{\textrm{non}}_i(t) ) \in \mathcal{A}_i, \forall i \in \mathcal{B}$, and  $r_{i}(\boldsymbol{s}_{ti}, \boldsymbol{a}_{t0}, \dots, \boldsymbol{a}_{tB})$. Therefore, to estimate the gradient of loss function \eqref{eq:loss_fn_dqn}, we can establish a relationship among the gradient of approximation $\hat{\nabla}_{\theta_{i}}L(\theta_{i})$ and true gradient  $\nabla_{\theta_{i}}L(\theta_{i})$,    
	\begin{equation} 
	\label{eq:Convergence_relation_policy_grdient}
	\begin{split}
	P\left(\hat{\nabla}_{\theta_{i}}L(\theta_{i}), \nabla_{\theta_{i}}L(\theta_{i}) > 0\right) \propto \left(0.5\right)^{|\mathcal{B}|}.
	\end{split}
	\end{equation}
	[See Appendix \ref{apd:Convergence_of_Proposed_Model}]. 
\end{proposition}

Propositions \ref{pro:discounted_reward_game_Nash_equilibrium_point} and \ref{pro:Convergence_Multi_Agent_Meta_RL} validate the optimality and convergence, respectively for the proposed MAMRL framework. Proposition \ref{pro:discounted_reward_game_Nash_equilibrium_point} guarantees an optimal energy dispatch policy. Meanwhile, Proposition \ref{pro:Convergence_Multi_Agent_Meta_RL} ensures that the proposed MAMRL model can meet the convergence for a single state $\boldsymbol{s}_{ti} \in \mathcal{S}, i \in \forall \mathcal{B}$. That implies this model is also able to converge for $\forall \boldsymbol{s}_{ti} \in \mathcal{S}, i \in \forall \mathcal{B}$.

The significance of the proposed MAMRL model are explained as follows:
\begin{itemize}
	\item First, each BS (i.e., local agent) can explore its own energy dispatch policy based on individual requirements for the energy generation and consumption. Meanwhile, the meta-agent exploits each BS energy dispatch decision from its own recurrent neural networks state information. As a result, individual BS anticipates its own energy demand and generation while meta-agent handles the non-i.i.d. energy demand and generation for all BS agents to efficiently meet the exploration-exploitation tradeoff of the proposed MAMRL.
	\item Second, the proposed MAMRL model can effectively handle distinct environment dynamics for non-i.i.d. energy demand and generation among the agents.
	\item Third, the proposed MAMRL model ensures less information exchange between the local agents and meta-agent. In particular, each local BS agent only sends an observational vector to meta-agent and received neural network parameters at the end of $15$ minutes observation period. Additionally, the proposed MAMRL model does not require sending an entire environment state from each local agent to the meta-agent.
	\item Finally, the meta-agent can learn a generalized model toward the energy dispatch decision and transfer its skill to each local BS agent. This, in turn, can significantly increase the learning accuracy as well as reduce the computational time for each local BS agent thus enhancing the robustness of the energy dispatch decision.
\end{itemize}

We benchmark the proposed MAMRL framework by performing an extensive experimental analysis, and the experimental analysis and discussion are given in the later section.

\section{Experimental Results and Analysis}
\begin{table}
	\caption{Summary of experimental setup}
	\begin{center}
		\begin{tabular}{|c|c|}
			\hline
			\textbf{Description}&{\textbf{Value}} \\
			\hline
			No. of SBSs (no. of local agents) & $9$ \\
			\hline
			No. of MEC servers in each SBS & $2$ \\
			\hline
			No. of MBS (meta-agent)& $1$ \\
			\hline
			Channel bandwidth & $180$ kHz \cite{IEEEhowto:Bairagi_Coexistence}\\
			\hline
			System bandwidth & $20$ MHz \cite{IEEEhowto:Tran_energy_MEC_no_downlink}\\
			\hline
			Transmission power & $27$ dB \cite{IEEEhowto:Mao_Stochastic_Joint_Radio_Sim_para}\\
			\hline
			Channel gain & $140.7 + 36.7\log d$ \cite{IEEEhowto:Tran_energy_MEC_no_downlink}\\
			\hline	
			A variance of an AWGN & -114 dBm/Hz \cite{IEEEhowto:Bairagi_Coexistence}\\
			\hline		
			Energy coefficient for data transfer $\delta_i^{\textrm{net}}$ &$2.8$ \cite{IEEEhowto:Total_Energy_Auer}\\
			\hline
			MEC server CPU frequency $f$ & 2.5 GHz \cite{IEEEhowto:Mao_Stochastic_Joint_Radio_Sim_para}\\
			\hline
			Server switching capacitance $\tau$ & $5 \times 10^{-27}$ (farad) \cite{IEEEhowto:Tran_energy_MEC_no_downlink}\\
			\hline
			MEC static energy $\eta^{\textrm{MEC}}_{\textrm{st}}(t)$ & $[7.5, 25]$ Watts \cite{IEEEhowto:TPD_Intel}\\
			\hline
			Task sizes (uniformly distributed)& $[31,1546060]$ bytes \cite{IEEEhowto:CRAWDAD_dataset_packet_delivery}\\
			\hline
			No. of task requests at BS $i$ & $[1,10,000]$ \cite{IEEEhowto:Munir_Edge_Microgrid}\\
			\hline
			Unit cost renewal energy $c^{\textrm{ren}}_t$&$\$50$ per MW-hour \cite{IEEEhowto:Per_Unit_Energy_Generation_Cost} \\
			\hline
			Unit cost non-renewal energy $c^{\textrm{non}}_t$&$\$102$ per MW-hour \cite{IEEEhowto:Per_Unit_Energy_Generation_Cost}\\
			\hline
			Unit cost storage energy $c^{\textrm{sto}}_t$& $10\%$ additional \cite{IEEEhowto:Xu_storage_cost10}\\
			\hline	
			Initial discount factor $\gamma$& $0.9$\\
			\hline
			Initial action selection probability & $[0.5, 0.5]$\\
			\hline
			One observation period $t$ & $15$ minutes\\
			\hline
			No. of episodes & $800$\\
			\hline
			No. of epochs $T$ for each day & $96$\\
			\hline
			No. of steps for each epoch at each agent $i$ & $J_i$ =  $[1,10,000]$ \cite{IEEEhowto:CRAWDAD_dataset_packet_delivery}\\
			\hline
			No. of actions & $2$ (i.e., $\xi^{\textrm{sto}}_i(t)$, $\xi^{\textrm{non}}_i(t))$\\
			\hline
			No. of LSTM units in one LSTM cell & $48$\\
			\hline
			No. of LSTM cells & $10$ (i.e., B+1)\\
			\hline
			LSTM cell API BasicLSTMCell(.)  & tf.contrib.rnn \cite{IEEEhowto:TensorFlow_BasicLSTMCell}\\
			\hline
			Entropy regularization coefficient $\beta$ & $0.05$\\
			\hline
			Learning rate & $0.001$\\
			\hline
			Optimizer & Adam \cite{IEEEhowto:adam_Kingma}\\
			\hline
			Output layer activation function & Softmax \cite{IEEEhowto:single_agent_Sutton_RL_book}\\
			\hline
		\end{tabular}
		\label{tab2_sim_param}
	\end{center}
	\vspace{-6mm}
\end{table}

\begin{figure}[!t]
	\centering
	\subfigure[MEC network energy demand]
	{
		\includegraphics[width=4.1cm]{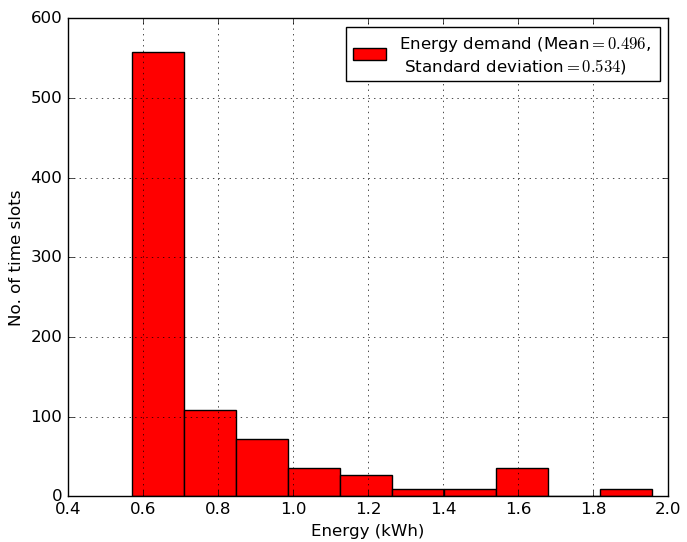}
		\label{fig:Hist_Demand}
	}
	\subfigure[Renewable energy generation]
	{
		\includegraphics[width=4.1cm]{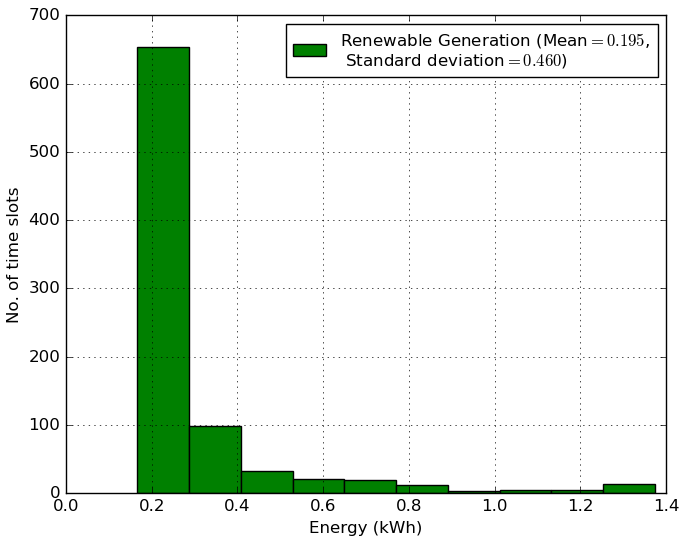}
		\label{fig:Hist_Renewable}
	}
	\caption{Histogram of energy demand and renewable energy generation for $9$ SBSs and each SBS consists of $96$ time slots after preprocessing using Algorithm \ref{alg:preprocessing_state_space_generation}.}
	\label{fig:J1_histo}
\end{figure}

In our experiment, we use the CRAWDAD nyupoly/video packet delivery dataset \cite{IEEEhowto:CRAWDAD_dataset_packet_delivery} to discretize the self-powered SBS network's energy consumption. Further, we choose a state-of-the-art UMass solar panel dataset \cite{IEEEhowto:UMass_Solar_panel_dataset} to evaluate renewable energy generation. We create deterministic, asymmetric, and stochastic environments by selecting different days of the same solar unit for the generation. Meanwhile, usage several session from the network packet delivery dataset. We train our proposed meta-reinforcement learning (Meta-RL)-based MAMRL framework using deterministic environment and evaluate the testing performance for the three environments. Three environments\footnote{For example, we train and test the MAMRL model using the known (i.e., deterministic environment) network energy consumption, and renewable generation data of day $1$. Then we have tested the trained model using day $2$ data, where network energy consumption is known, and renewable generation is unknown which represents an asymmetric environment. In a stochastic environment, let us consider day $3$ data, where both energy consumption and renewable generation are unknown to the trained model.} are as follows: 1) In the deterministic environment, both network energy consumption and renewable generation are known, 2) network energy consumption is known but renewable generation is unknown in the asymmetric environment, and 3) the stochastic environment contains both energy consumption and renewable generation are unknown. To benchmark the proposed MAMRL framework intuitively, we have considered a centralized single-agent deep-RL, multi-agent centralized A3C deep-RL with a same neural networks configuration as the proposed MAMRL, and a pure greedy model as baselines. These are as follows:
\begin{itemize}
	\item  We consider the neural advantage actor-critic (A2C) \cite{IEEEhowto:single_agent_Sutton_RL_book, IEEEhowto:AC_Base_Takahashi} method as a centralized single-agent deep-RL. In particular, the learning environment encompasses the state information of all BSs $\forall i \in \mathcal{B}$ and is learned by a neural A2C \cite{IEEEhowto:single_agent_Sutton_RL_book, IEEEhowto:AC_Base_Takahashi} scheme with the same configuration as the MAMRL model.
		
	\item An asynchronous advantage actor-critic (A3C) based multi-agent RL framework \cite{IEEEhowto:Lowe_Multi_agent_actor_critic} is considered a second benchmark in a cooperative environment \cite{IEEEhowto:Mnih_async_a3c}. In particular, each local actor can find its own policy in a decentralized manner while a centralized critic is augmented with additional policy information. Therefore, this model is learned by a centralized training with decentralized execution \cite{IEEEhowto:Lowe_Multi_agent_actor_critic}. We call this model a multi-agent centralized A3C deep-RL \cite{IEEEhowto:Lowe_Multi_agent_actor_critic}. The environment (i.e., state information) of this model remains the same for all of the local actor agents. To ensure a meaningful comparison with the proposed MAMRL model, we employ this joint energy dispatch policy using the same advantage function \eqref{eq:game_A_function} as the MAMRL model.
	
	\item We deploy a pure greedy-based algorithm \cite{IEEEhowto:single_agent_Sutton_RL_book} to find the best action-value mapping. In particular, this algorithm never takes the risk to choose an unknown action. Meanwhile, it explores other strategies and learns from them so as to infer more reasonable decisions. Thus, we choose this upper confidence bounded action selection mechanism \cite{IEEEhowto:single_agent_Sutton_RL_book} as one of the baselines used for benchmarking our proposed MAMRL model.
\end{itemize}

We implement our MAMRL framework using multi-threading programming in Python platform \footnote{MAMRL}, along with TensorFlow APIs \cite{IEEEhowto:TensorFlow_python}. Table \ref{tab2_sim_param} shows the key parameters of this experiment setup. 


\begin{figure}[!t]
	\centering
	\includegraphics[width=8.6cm]{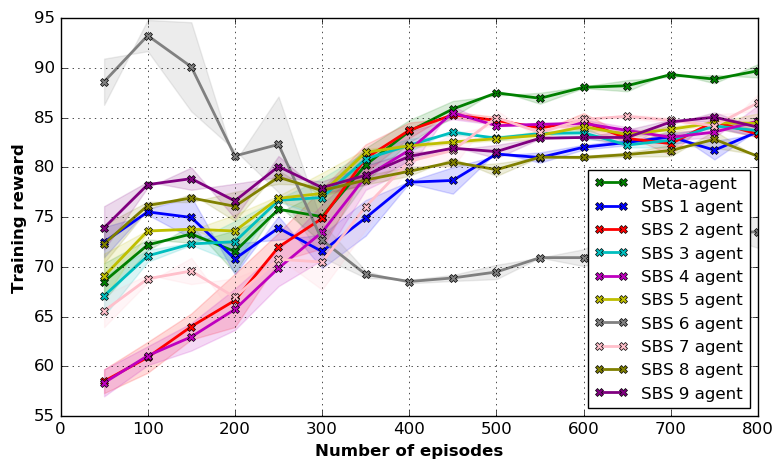}
	\caption{Reward value achieved for proposed Meta-RL training of the meta-agent alone with other SBS agents.}
	\label{fig:J1_Meta_RL_All_Agent_Reward}
	\vspace{-4mm}
\end{figure}
\begin{figure}[!t]
	\centering
	\includegraphics[width=8.6cm]{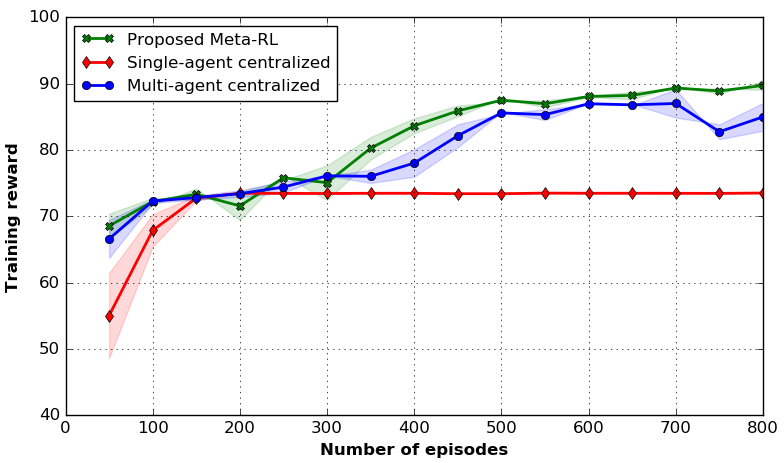}
	\caption{Reward value achieved of proposed Meta-RL, single-agent centralized, and multi-agent centralized methods.}
	\label{fig:J1_Reward}
	\vspace{-4mm}
\end{figure}

We prepossess both of the datasets (\cite{IEEEhowto:CRAWDAD_dataset_packet_delivery} and \cite{IEEEhowto:UMass_Solar_panel_dataset}) using Algorithm \ref{alg:preprocessing_state_space_generation} that generates the state space information. The histograms of the network energy demand (in \ref{fig:Hist_Demand}) and a renewable energy generation (in \ref{fig:Hist_Renewable}) of the deterministic environment are shown in Fig. \ref{fig:J1_histo}. To the best of our knowledge, there are no publicly available datasets that comprises both of energy consumption and generation of a self-powered network with MEC capabilities. Additionally, if we change the environment using other datasets, the proposed MAMRL framework can deal with the new, unknown environment by using the skill transfer feature from the meta-agent to each local BS agent. In particular, the MAMRL approach can readily deal with the case in which the BS agent achieves a much lower reward due to more variability in consumption and generation. As a result, the above experiment setup is reasonable for the benchmarking of the proposed MAMRL framework.           

\begin{figure}
	\centering
	\subfigure[Proposed Meta-RL]
	{
		\includegraphics[width=8.6cm]{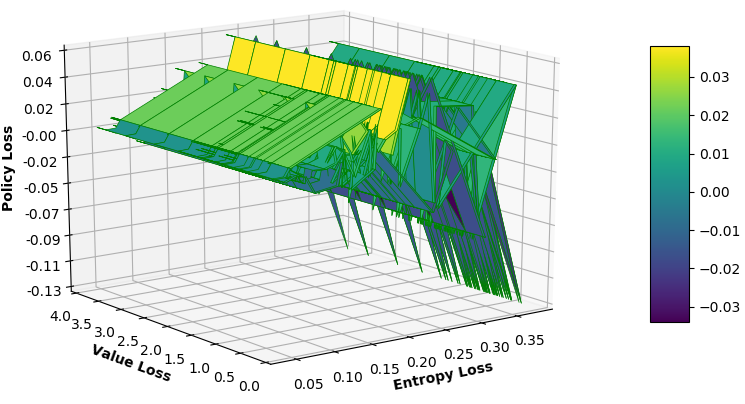}
		\label{fig:first_sub_Meta_RL}
	}
	\\
	\subfigure[Single-agent centralized]
	{
		\includegraphics[width=8.6cm]{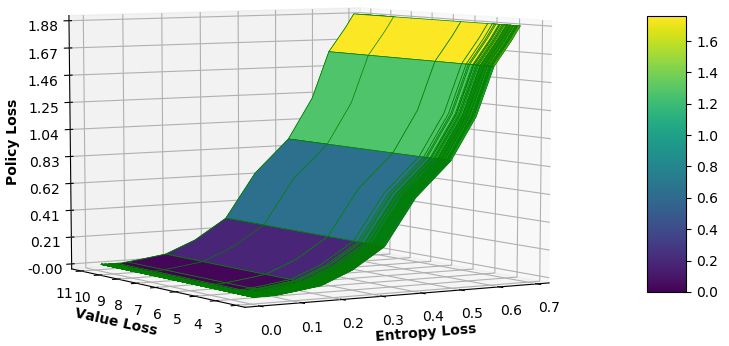}
		\label{fig:second_sub_Single_agent}
	}
	\\
	\subfigure[Multi-agent centralized]
	{
		\includegraphics[width=8.6cm]{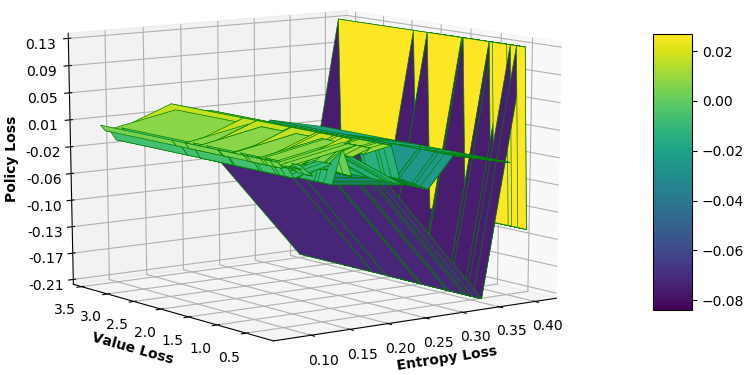}
		\label{fig:third_sub_Multi_agent}
	}
	\caption{Relationship among the entropy loss, value loss, and policy loss in the training phase of proposed Meta-RL, single-agent centralized, and multi-agent centralized methods.}
	\label{fig:J1_all_loss}
	\vspace{-4mm}
\end{figure}

Fig. \ref{fig:J1_Meta_RL_All_Agent_Reward} illustrates the reward achieved by each local SBS along with a meta-agent, where we take an average reward for each $50$ episodes. In the MAMRL setting, we design a maximum reward of $96$ ($15$ minute slot for $24$ hours), where meta-agent converges with a high reward value (around $90$). Hence, all of the local agents converge with around $80-85$ reward value except the SBS $6$ that achieves a reward of $70$ at convergence because its energy consumption and generation vary more than the others. In fact, this variation of reward among the BSs is leading to anticipate the non-i.i.d. energy demand and generation of the considered network as well as densification of the exploration and exploitation tradeoff for energy dispatch. 
The proposed approach can adapt the uncertain energy demand and generation over time by characterizing the expected amount of uncertainty in an energy dispatch decision for each BS $i \in \mathcal{B}$ individually. Meanwhile, the meta-agent exploits the energy dispatch decision by employing a joint policy toward the globally optimal energy dispatch for each BS $i \in \mathcal{B}$. Therefore, the challenges of distinct energy demand and generation of the state space among the BSs can be efficiently handled by applying learned parameters from the meta-agent to each BS $i \in \mathcal{B}$ during the training that establishes a balance between exploration and exploitation.

We compare the achieved reward of proposed MAMRL model with single-agent centralized and multi-agent centralized models in Fig. \ref{fig:J1_Reward}. The single agent centralized (diamond mark with red line) model converges faster than the other two models but it achieves the lowest reward due to the lack of exploitation as it has only one agent. Further, the multi-agent centralized (circle mark with blue line) model converges with a higher reward than the single agent method. The proposed MAMRL (cross mark with green line) model outperforms the other two models while converges with the highest reward value. In addition, multi-agent centralized needs the entire state information.
In contrast, the meta-agent requires only the observation from local agents, and it can optimize the neural network parameters by using its own state information.

We analyze the relationship among the value loss, entropy loss, and policy loss in Fig. \ref{fig:J1_all_loss}, where the maximum policy loss of the proposed MAMRL (in \ref{fig:first_sub_Meta_RL}) model is around $0.06$ whereas single-agent centralized (in \ref{fig:second_sub_Single_agent}) and multi-agent centralized (in \ref{fig:third_sub_Multi_agent}) methods gain about $1.88$ and $0.12$, respectively. Therefore, the training accuracy increases due to more variation between exploration and exploitation. Thus, our MAMRL model is capable of incorporating the decision of each local BS agent that solves the challenge of non-i.i.d. demand-generation for the other BSs.
\begin{figure}[!t]
	\centering
	\includegraphics[width=8.0cm]{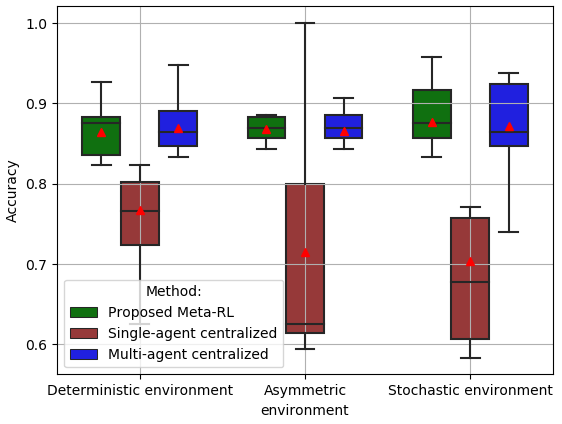}
	\caption{Testing accuracy of the proposed Meta-RL, single-agent centralized, and multi-agent centralized methods with deterministic, asymmetric, and stochastic environments of the $9$ SBSs.}
	\label{fig:J1_Accuracy_Box}
	\vspace{-4mm}
\end{figure}

In Fig. \ref{fig:J1_Accuracy_Box}, we examine the testing accuracy  \cite{IEEEhowto:Model_evaluation_scikit} of the storage energy $\xi^{\textrm{sto}}_i(t)$ and the non-renewable energy generation decision \footnote{Each BS agent $i \in \mathcal{B}$ can calculate its action from a globally optimal energy dispatch policy $\pi_{\theta_{i}}^*$ by using $\argmax(.)$ (i.e., $\argmax_{\pi_{\theta_{i}}^*} (\boldsymbol{a}_{ti})$). In which, at the end of $15$ minutes duration of each time slot $t$, the each BS agent $i \in \mathcal{B}$ can choose one action (i.e., storage or non-renewable) from the energy dispatch policy $\pi_{\theta_{i}}^*$.} $\xi^{\textrm{non}}_i(t)$ for $96$ time slots ($1$ days) of $9$ SBSs under the deterministic, asymmetric, and stochastic environments. In the experiment, we have used the well-known UMass solar panel dataset \cite{IEEEhowto:UMass_Solar_panel_dataset} for renewable energy generation information as well as, the CRAWDAD nyupoly/video dataset\cite{IEEEhowto:CRAWDAD_dataset_packet_delivery}, for estimating the energy consumption of the self-powered network. Further, we preprocess both of the datasets (\cite{IEEEhowto:CRAWDAD_dataset_packet_delivery} and \cite{IEEEhowto:UMass_Solar_panel_dataset}) using Algorithm \ref{alg:preprocessing_state_space_generation} that generates the state space information. Thus, the \emph{ground truth} comes from this state-space information of the considered datasets, where the actions are depended on the renewable energy generation and consumption of a particular BS $i \in \mathcal{B}$. The proposed MAMRL (green box) and multi-agent centralized (blue box) methods achieve a maximum accuracy of around $95\%$ and $92\%$, respectively, under the stochastic environment (in Fig. \ref{fig:J1_Accuracy_Box}). Further, Fig. \ref{fig:J1_Accuracy_Box} shows that the mean accuracy ($88\%$) of the proposed method is also higher than the centralized solution ($86\%$). Similarly, in the deterministic and asymmetric environment, the average accuracy (around $87\%$) of the proposed low complexity semi-distributed solution is almost the same as the baseline method.
\begin{figure}[!t]
	\centering
	\includegraphics[width=8.6cm]{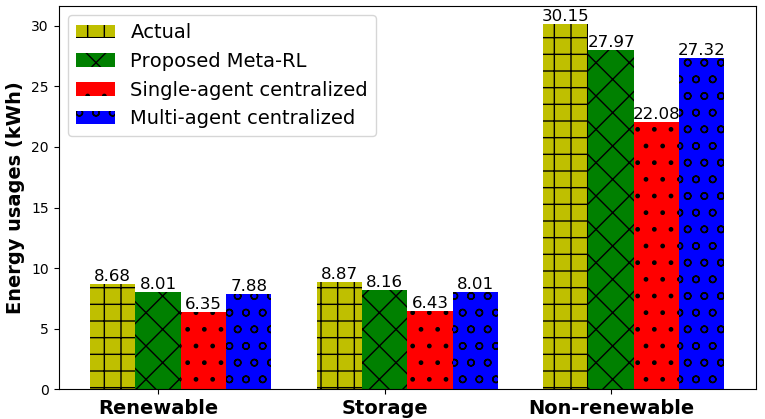}
	\caption{Prediction result of renewable, storage, and non-renewable energy usages of a single SBS (SBS $2$) for $24$ hours ($96$ time slots) under the stochastic environment.}
	\label{fig:J1_Energy_Usages_SBS_2_Unknown}
	\vspace{-4mm}
\end{figure}
\begin{figure}[!t]
	\centering
	\includegraphics[width=8.6cm]{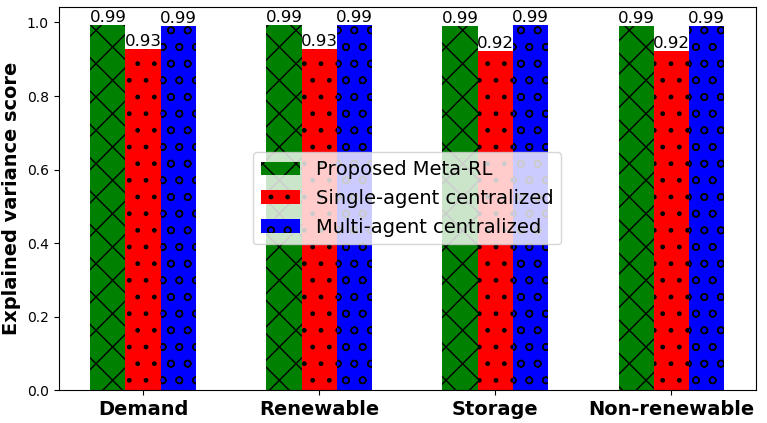}
	\caption{Explained variance score of a single SBS (SBS 2) for $24$ hours ($96$ time slots) under the stochastic environment.}
	\label{fig:J1_Explained_variance_score}
	\vspace{-4mm}
\end{figure}
\begin{figure}[!t]
	\centering
	\includegraphics[width=8.6cm]{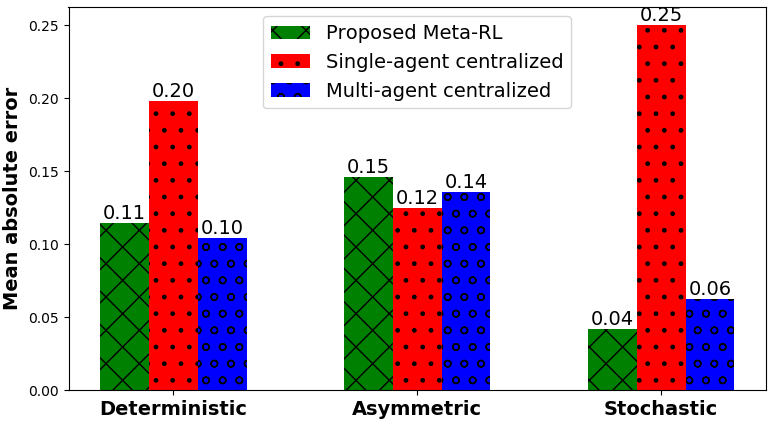}
	\caption{Mean absolute error of a single SBS (SBS 2) for $24$ hours ($96$ time slots) under the stochastic environment.}
	\label{fig:J1_MEA_Decision}
	\vspace{-4mm}
\end{figure}
\begin{figure}[!t]
	\centering
	\includegraphics[width=8.6cm]{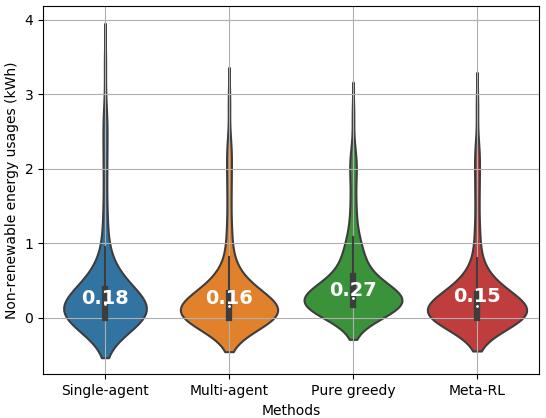}
	\caption{Kernel density analysis of non-renewable energy usages for $24$ hours ($96$ time slots) under the stochastic environment.}
	\label{fig:J1_Non_renewable_energy_usages}
	\vspace{-4mm}
\end{figure}
The prediction results of renewable, storage and non-renewable energy usage for a single SBS (SBS $2$) for $24$ hours ($96$ time slots) under the stochastic environment are shown in Fig. \ref{fig:J1_Energy_Usages_SBS_2_Unknown}. The proposed MAMRL outperforms all other baselines and achieves an accuracy of around $95.8\%$. In contrast, the accuracy of the other two methods is $75\%$ and $93.7\%$ for the single-agent centralized and multi-agent centralized, respectively. 

In Figs. \ref{fig:J1_Explained_variance_score} and \ref{fig:J1_MEA_Decision}, we validate our approach with two standard regression model evaluation metric, explained variance \footnote{We measure the discrepancy for energy dispatch decisions between the proposed and baseline models on the ground truth of the datasets (\cite{IEEEhowto:CRAWDAD_dataset_packet_delivery} and \cite{IEEEhowto:UMass_Solar_panel_dataset}). We deploy the explained variance regression score function using \emph{sklearn} API \cite{IEEEhowto:Model_evaluation_EV_scikit} to measure and compare this discrepancy. } (i.e., explained variation) and mean absolute error (MAE) \cite{IEEEhowto:Model_evaluation_scikit}, respectively. Fig. \ref{fig:J1_Explained_variance_score} shows that the explained variance score of the proposed MAMRL method almost the same as the multi-agent centralized. However, in the case of renewable energy generation, MAMRL method significantly performs better (i.e., $1\%$ more score) than the multi-agent centralized solution. In particular, the proposed MAMRL model has pursued the uncertainty of renewable energy generation by the dynamics of Markovian for each BS. Further, meta-agent anticipates the energy dispatch by other BSs decisions and its own state information. We analyze the MAE \footnote{This performance metric provides us with the average magnitude of errors for the energy dispatch decision of a single SBS (SBS 2) for $24$ hours ($96$ time slots). Particularly, we analyze the average error over the $96$ time slots of the absolute differences between prediction and actual observation. To evaluate this metric, we have used the mean absolute error regression loss function of \emph{sklearn} API \cite{IEEEhowto:Model_evaluation_MAE_scikit}. } for the three environments (i.e.,  deterministic, asymmetric, and stochastic) among the proposed MAMRL, single-agent centralized, and multi-agent centralized methods in Fig. \ref{fig:J1_MEA_Decision}. The MAE of the proposed MAMRL is $11\%$, $15\%$, and $4\%$ for deterministic, asymmetric, and stochastic, respectively since meta-agent has the capability to adopt the uncertain environment very fast. This adaptability is enhanced by the exploration mechanism that is taken into account at each BS, and exploitation that performs by capitalizing the non-i.i.d. explored information of all BSs.  

\begin{figure}[!t]
	\centering
	\includegraphics[width=8.6cm]{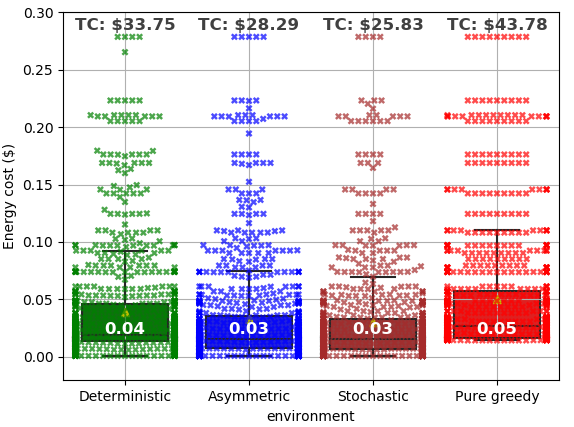}
	\caption{Energy consumption cost analysis of $9$ SBSs for $24$ hours ($96$ time slots) under deterministic, asymmetric, and stochastic environments using the proposed Meta-RL method over pure greedy method.}
	\label{fig:J1_Meta_RL_Energy_Cost}
	\vspace{-4mm}
\end{figure}

Fig. \ref{fig:J1_Non_renewable_energy_usages} illustrates the efficacy of the proposed MAMRL model in terms of the non-renewable energy usages into a stochastic environment with other benchmarks. This figure considers a kernel density analysis for $24$ hours ($96$ time slots) under a stochastic environment, where the median of the non-renewable energy usages $0.15$ (kWh), and $0.27$ (kWh) for the proposed MAMRL, and pure greedy, respectively, at each $15$ minutes time slot. Further, the proposed MAMRL can significantly reduce the usages of non-renewable energy for the considered self-powered wireless network, where the MAMRL can save up to $13.3\%$ of the non-renewable energy usages. Here, the meta agent of the MAMRL model can discretize uncertainty from each local BS agent and transfer the knowledge (i.e., learning parameters) to each local agent that can take a globally optimal energy dispatch decision.

\begin{figure}[!h]
	\centering
	\includegraphics[width=8.6cm]{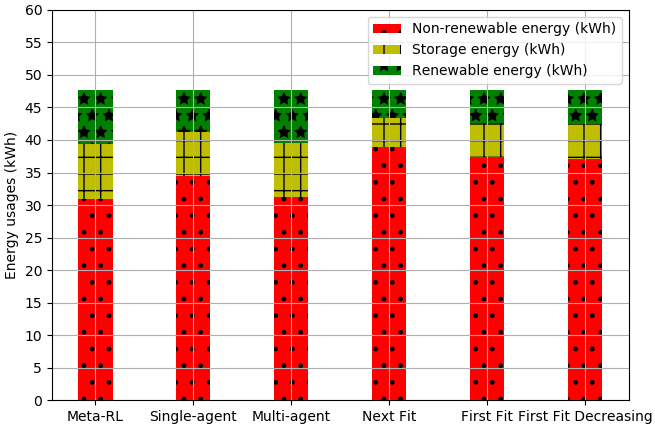}
	\caption{Amount of renewable, non-renewable, and storage energy estimation for $24$ hours ($96$ time slots) for proposed meta-RL, single-agent RL, multi-agent RL, next fit, first fit, and first fit decreasing methods.}
	\label{fig:J1_comp_with_others}
	\vspace{-4mm}
\end{figure}

\begin{figure}[!t]
	\centering
	\includegraphics[width=8.6cm]{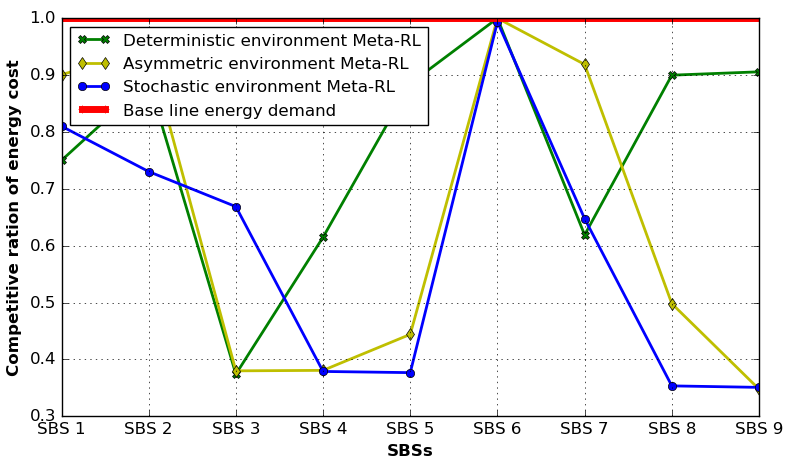}
	\caption{Competitive cost ratio of the proposed Meta-RL method for $24$ hours ($96$ time slots) under the deterministic, asymmetric, and stochastic environments.}
	\label{fig:J1_Competitive_ratio}
	\vspace{-4mm}
\end{figure}
\begin{table*}[!t]
	\caption{Comparison between the proposed method and other methods with ground truth for a single SBS (SBS 2) for $24$ hours ($96$ time slots) under the stochastic environment.}
	\centering
	\begin{tabular}{|p{3.6cm}|p{1.2cm}|p{1.2cm}|p{1.6cm}|p{1.2cm}|p{1.2cm}|p{1.6cm}|p{1.2cm}|p{1.2cm}|}
		\hline
		\textbf{\small Method} & \textbf{\small Non-renewable energy usage (kWh)} & \textbf{\small Storage energy usage (kWh)} & 
		\textbf{\small Renewable energy usage (kWh)} & 
		\textbf{\small Non-renewable energy usage cost (\$)}   & 
		\textbf{\small Storage energy usage cost (\$)} & 
		\textbf{\small Renewable energy usage cost (\$)} & 
		\textbf{\small Total energy usage cost (\$)} & 
		\textbf{\small Cost difference with ground truth (\%)}\\ 
		\hline
		\small Ground truth (i.e., optimal) & \small $30.15$ & \small $8.87$ & \small $8.67$  & \small $3.07$ & \small $0.49$ & \small $0.43$ & \small $3.99$  & \small  NA \\ 
		\hline
		\textbf{\small MAMRL (proposed)} & \small $\boldsymbol{30.88}$ & \small $\boldsymbol{8.50}$ & \small $\boldsymbol{8.32}$  & \small $\boldsymbol{3.14}$ & \small $\boldsymbol{0.47}$ & \small  $\boldsymbol{0.42}$ & \small $\boldsymbol{4.03}$ & \small  $\boldsymbol{0.90}$ \\ 
		\hline
		\small Single-agent RL & \small $34.53$ & \small $6.65$ & \small $6.50$  & \small $3.52$ & \small $0.37$ & \small  $0.33$ & \small $4.21$   & \small  $5.43$ \\ 
		\hline
		\small Multi-agent RL & \small $31.24$ & \small $8.31$ & \small $8.14$  & \small $3.19$ & \small $0.46$ & \small  $0.41$ & \small $4.05$ & \small  $1.36$ \\ 
		\hline
		\small Next Fit & \small $38.92$ & \small $4.44$ & \small $4.34$  & \small $3.97$ & \small $0.24$ & \small  $0.21$ & \small $4.43$ & \small  $10.86$ \\ 
		\hline
		\small First Fit & \small $37.37$ & \small $5.22$ & \small $5.10$  & \small $3.81$ & \small $0.29$ & \small  $0.26$ & \small $4.35$ & \small  $8.94$ \\ 
		\hline
		\small First Fit Decreasing & \small $37.12$ & \small $5.34$ & \small $5.23$  & \small $3.79$ & \small $0.30$ & \small  $0.26$ & \small $4.34$ & \small  $8.63$\\ 
		\hline
		\small Without renewable & \small $47.69$ & \small NA & \small NA & \small $4.86$ & \small NA & \small  NA & \small $4.86$ & \small  $21.72$ \\ 
		\hline
	\end{tabular}
	\label{tab:comparison_energy_cost}
\end{table*}

Fig. \ref{fig:J1_Meta_RL_Energy_Cost} presents the energy consumption cost analysis for $9$ SBSs over $24$ hours ($96$ time slots) under deterministic, asymmetric, and stochastic environments using the proposed Meta-RL method while comparing it to the pure greedy method. The total energy cost achieved by the proposed approach for a particular day will be $\$33.75$, $\$28.29$, and $\$25.83$ for deterministic, asymmetric, and stochastic environments, respectively. Fig. \ref{fig:J1_Meta_RL_Energy_Cost} also shows that the proposed method significantly reduces the energy consumption cost (by at least $22.4\%$) for all three environments over the pure greedy method. The median of the energy cost at each time slot is $\$0.04$, $\$0.03$, and $\$0.03$ for the deterministic, asymmetric, and stochastic environments, respectively. In contrast, Fig. \ref{fig:J1_Meta_RL_Energy_Cost} has shown that a median energy cost for the pure greedy baseline is $\$0.05$ at each time slot that is due to a lack of the competence to cope with an unknown environment for energy consumption and generation. Therefore, the proposed MAMRL model can overcome the challenges of an unknown environment as well as non-i.i.d. characteristics for energy consumption and generation of a self-powered MEC network.

In Fig. \ref{fig:J1_comp_with_others}, we compare our proposed meta-RL model with single-agent RL, multi-agent RL, next fit, first fit, and first fit decreasing methods in terms of amount of renewable, non-renewable, and storage energy usages for $24$ hours ($96$ time slots). Fig. \ref{fig:J1_comp_with_others} shows that the proposed MAMRL model outperforms the others that achieves around $22\%$ less non-renewable energy usages than the next fit scheduling algorithm. Additionally, next fit, first fit, and first fit decreasing scheduling methods \cite{IEEEhowto:Johnson_near_optimal_bin} cannot capture the uncertainty of energy generation and consumption, as well as provide a near optimal solution. Further, a comparison between the proposed method and other methods with the ground truth for a single SBS (SBS 2) for $24$ hours ($96$ time slots) under the stochastic environment is illustrated in Table \ref{tab:comparison_energy_cost}. The proposed method can achieve significant outcomes with respect to energy cost as compared with the ground truth. In particular, the experiment shows that the energy usage cost difference between the proposed method and ground truth is around $1\%$ for a single BS (in Table \ref{tab:comparison_energy_cost}). This leads to one of the evidence that the proposed MAMRL can adopt the unknown environment and can utilize it during the execution for each BS energy dispatch.

Finally, in Fig. \ref{fig:J1_Competitive_ratio}, we examine the competitive cost ratio \cite{IEEEhowto:Zhang_time_slot} of the proposed MAMRL framework. From this figure, we observe that the proposed MAMRL framework effectively minimizes the energy consumption cost for each BS under deterministic, asymmetric, and stochastic environments. In fact, Fig. \ref{fig:J1_Competitive_ratio} ensures the robustness of the proposed MAMRL framework that is performed a tremendous performance gain by coping with non-i.i.d. energy consumption and generation under the uncertainty. Furthermore, in MAMRL training, each local agent has captured the time-variant features of energy demand and generation from the historical data while meta-agent optimizes energy dispatch decisions by obtaining those features with its own parameters of LSTM. In the case of testing, a generalized MAMRL trained model is employed that makes a fully independent and unbiased energy dispatch from an unknown environment. To this end, the proposed MAMRL framework shows the efficacy of solving the energy dispatch of a self-powered wireless network with MEC capabilities with a higher degree of reliability.

\section{Conclusions}
In this paper, we have investigated an energy dispatch problem of a self-powered wireless network with MEC capabilities. We have formulated a two-stage stochastic linear programming energy dispatch problem for the considered network. To solve the energy dispatch problem in a semi-distributed manner, we have proposed a novel multi-agent meta-reinforcement learning framework. In particular, each local BS agent obtains the time-varying features by capturing the Markovian properties of the network's energy consumption and renewable generation for each BS unit, and predict its own energy dispatch policy. Meanwhile, a meta-agent optimizes each BS agent's energy dispatch policy from its own state information, and it transfers global learning parameters to each BS agent so that they can update their energy dispatch policy into an optimal policy. We have shown that the proposed MAMRL framework can capture the uncertainty of non-i.i.d. energy demand and generation for the self-powered wireless network with MEC capabilities. Our experimental results have shown that the proposed MAMRL framework can save a significant amount of non-renewable energy with higher accuracy prediction that ensures the energy sustainability of the network. In particular, the performance of energy dispatch over deterministic, asymmetric, and stochastic environments outperform other baseline approaches, where average accuracy achieves up to $95.8\%$ and reduces the energy cost about $22.4\%$ of the self-powered wireless network. To this end, the proposed MAMRL model can reduce by at least $11\%$ of the non-renewable energy usage for the self-powered wireless network.


%

\appendices
\appendices
\section{Example of Information Exchange between Local BS Agent and Meta Agent in MAMRL Framework} \label{apd:example_information_exchange}
For example, consider an LSTM cell with $48$ LSTM units \cite{IEEEhowto:Hochreiter_LSTM,IEEEhowto:TensorFlow_BasicLSTMCell}. Thus, the dimension of forget gate, input gate, gate/memory/activation gate, and output gate will be $48$ for each gate. Now consider a local BS agent $i \in \mathcal{B}$ that embedding a dimension of $3$ inputs $(r_{i}(\boldsymbol{a}_{ti},\boldsymbol{s}_{ti}), \boldsymbol{a}_{ti}, t')$ to a local LSTM cell. This input comes from the state information $\boldsymbol{s}_{ti} \colon ( \xi_i^{\textrm{d}}, \xi_i^{\textrm{ren}}(t), C^{\textrm{sto}}_{ti}, C^{\textrm{non}}_{ti} )$ of a local BS agent $i$. As a result, inputs are appended to all gates during the training. Therefore, the number of learning parameters will be $4 \times (48(48+3)+48)$ (i.e., $gates \times [units (units+input)+units]$). Additionally, the size of hidden state and cell state parameters remain $48$ for each due to an LSTM cell with $48$ LSTM units. Further, on top of the LSTM cell, we have two fully connected output layers, a fully connected output layer with a Softmax activation to determine the local energy dispatch policy. Meanwhile, advantage is determined from another fully connected output layer without activation function by value function estimation. The hidden and cell state of each local agent are updated by receiving the state parameters with a $48 \times 2$ dimensional data from the meta-agent. In case of the meta-agent, the configuration of LSTM cell is the same as each local LSTM cell. Therefore, at the end of each time slot duration, the  meta-agent sends a $48 \times 2$ dimensional state information to each local BS agent $i \in \mathcal{B}$. Subsequently, the meta-agent receives a $6$ dimensional observation  ${\boldsymbol{o}_{i} \colon ( r_{i}(\boldsymbol{a}_{t'i},\boldsymbol{s}_{t'i}), r_{i}(\boldsymbol{a}_{ti},\boldsymbol{s}_{ti}), \boldsymbol{a}_{ti}, \boldsymbol{a}_{t'i}, t',\Lambda^{\pi_{\theta_{i}}}(\boldsymbol{s}_{ti},\boldsymbol{a}_{ti}))}$ as an input from each local BS agent, where the number of learning parameters at the meta-agent will be $4\times (48(48+6)+48)$ for each iteration. The output layer of the meta-agent also consists of two fully connected output layers for determining meta-policy (i.e., joint policy) and meta advantage. Thus, these output layers do not affect the dimension of hidden and cell states for the meta agent's LSTM cell. In fact, these RNN states are used as an input to these fully connected layers. As a result, for each epoch (i.e., end of a time slot duration), the meta-agent sends $48 \times 2$ dimensional RNN states to each local agent along with an energy dispatch policy, and each local agent sends $6$ dimensional observation to the meta-agent.

\section{Proof of Proposition \ref{pro:discounted_reward_game_Nash_equilibrium_point} } \label{apd:discounted_reward_game_Nash_equilibrium_point_proof}
\begin{proof}
	For a BS agent $i$, energy dispatch policy $\pi_{\theta_{i}}^*$ is the best response for the equilibrium responses from all other BS agents. Thus, the BS agent $i$ can not be improved the value $V^{\pi_{\theta_{i}}^*}(\boldsymbol{s}_{ti})$ any more by deviating of policy  $\pi_{\theta_{i}}^*$. Therefore, \eqref{eq:loss_fn_dqn} holds the following property,
	\begin{equation} \label{eq:game_optimal_function}
	\begin{split}
	V^{\pi_{\theta_{i}}^*}(\boldsymbol{s}_{ti}) \ge 
	r_{i}(\boldsymbol{s}_{ti}, \boldsymbol{a}_{t0}, \dots, \boldsymbol{a}_{tB}) \; + \;\;\;\;\;\;\;\;\;\;\;\;\;\;\;\;\;\;\;\;\;\;\;\;\;\;\;\;\;\;\;\;\;\;\; \\ \sum_{\boldsymbol{s}_{t'i} \in \mathcal{S}_i, t'=t}^{\infty} \gamma^{t'-t} \Gamma(\boldsymbol{s}_{t'i}|\boldsymbol{s}_{ti}, \boldsymbol{a}_{t0}, \dots,\boldsymbol{a}_{tB}) V^{\pi_{\theta_{i}}}(\boldsymbol{s}_{t'i}, \pi_{\theta_{0}}^*,\dots, \pi_{\theta_{B}}^* ).\\
	\end{split}
	\end{equation}
Hence, the meta-agent $M_t(\mathcal{O}_t; \phi)$ of the $|\mathcal{B}|$-agent energy dispatch model (i.e., MAMRL) reaches a Nash equilibrium point for policy $\pi_{\theta_{i}}^*$ with parameters $\theta_{i}$. As a result, the optimal value of BS agent $i\in \mathcal{B}$ can be as follows:
\begin{equation} \label{eq:meta_equilibrium}
\begin{split}
V^{\pi_{\theta_{i}}^*}(\boldsymbol{s}_{ti}) = M_t(\nabla_{\theta_{t}}L(\theta_{t}); \phi).
\end{split}
\end{equation}
\eqref{eq:meta_equilibrium} implies that $\pi_{\theta_{i}}^*$ is an optimal policy of energy dispatch decisions. Thus, the optimal policy $\pi_{\theta_{i}}^*$ belongs to a Nash equilibrium point and holds the following inequality,
\begin{equation} \label{eq:meta_equilibrium_loss}
\begin{split}
V^{\pi_{\theta_{i}}^*}(\boldsymbol{s}_{ti}) \ge \mathbb{E}_{L(\theta)}[L(\theta^*(L(\theta);\phi))]
\end{split}
\end{equation}
\end{proof}
\section{Proof of Proposition \ref{pro:Convergence_Multi_Agent_Meta_RL}} \label{apd:Convergence_of_Proposed_Model}
\begin{proof}
	A probability of action $\boldsymbol{a}_{ti}$ of BS agent $i \in \mathcal{B}$ at time $t$ can be presented as follows: 
	\begin{equation} 
	\label{eq:Convergence_action_probability}
	\begin{split}
	P(\boldsymbol{a}_{ti}) = \theta_{i}^{\boldsymbol{a}_{ti}} (1-\theta_{i})^{1-\boldsymbol{a}_{ti}} \;\;\;\;\;\;\;\;\;\;\;\;\;\;\;\;\;\;\;\;\;\;\;\\
	\;\;\;\;\;\;\;\;\;\;\;\;\;\;\;	= {\boldsymbol{a}_{ti}} \log \theta_{i} + ({1-\boldsymbol{a}_{ti}}) \log (1-\theta_{i}).
	\end{split}
	\end{equation} 
	We consider a single state, and a policy gradient estimator can be defined as,
	\begin{equation} 
	\label{eq:Convergence_policy_gradient_estimator}
	\begin{split}
	\frac{\hat{\partial}}{\partial \theta_{i}} L(\theta_{i}) = r_{i}(\boldsymbol{s}_{ti}, \boldsymbol{a}_{t0}, \dots, \boldsymbol{a}_{tB}) \frac{\partial}{\partial \theta_{i}} \log P(\boldsymbol{a}_{t0}, \dots, \boldsymbol{a}_{tB}) \;\;\;\;\;\;\;\;\;\;\;\;\;\\
	= r_{i}(\boldsymbol{s}_{ti}, \boldsymbol{a}_{t0}, \dots, \boldsymbol{a}_{tB}) \frac{\partial}{\partial \theta_{i}} \sum_{\forall i \in \mathcal{B}} {\boldsymbol{a}_{ti}} \log \theta_{i} + ({1-\boldsymbol{a}_{ti}}) \log (1-\theta_{i}) \\
	= r_{i}(\boldsymbol{s}_{ti}, \boldsymbol{a}_{t0}, \dots, \boldsymbol{a}_{tB}) \frac{\partial}{\partial \theta_{i}} ( {\boldsymbol{a}_{ti}} \log \theta_{i} + ({1-\boldsymbol{a}_{ti}}) \log (1-\theta_{i})) \;\;\;\;\;\;\;\\
	= r_{i}(\boldsymbol{s}_{ti}, \boldsymbol{a}_{t0}, \dots, \boldsymbol{a}_{tB}) (\frac{\boldsymbol{a}_{ti}}{\theta_{i}} - \frac{(1-\boldsymbol{a}_{ti})}{(1-\theta_{i})}) \;\;\;\;\;\;\;\;\;\;\;\;\;\;\;\;\;\;\;\;\;\;\;\;\;\;\;\;\;\;\;\;\;\;\;\;\\
	= r_{i}(\boldsymbol{s}_{ti}, \boldsymbol{a}_{t0}, \dots, \boldsymbol{a}_{tB}) (2\boldsymbol{a}_{ti}-1),\;
	\text{for $\theta_{i} = 0.5$}. \;\;\;\;\;\;\;\;\;\;\;\;\;\;\;\;\;\;\;\;\;\;\;\;\;\;
	\end{split}
	\end{equation}
	Thus, an expected reward for $|\mathcal{B}|$ BS agents can be represented as, $\mathbb{E}[r_{i}] = \sum_{\forall i \in \mathcal{B}} r_{i}(\boldsymbol{s}_{ti}, \boldsymbol{a}_{t0}, \dots, \boldsymbol{a}_{tB}) (0.5)^{|\mathcal{B}|}$, where by applying $r_{i}(\boldsymbol{s}_{ti}, \boldsymbol{a}_{t0}, \dots, \boldsymbol{a}_{tB})= \boldsymbol{1} |r_{i}(\boldsymbol{s}_{ti}, \boldsymbol{a}_{t0}, \dots, \boldsymbol{a}_{tB})$, we can get $\mathbb{E}[r_{i}] = (0.5)^{|\mathcal{B}|}$. Now, we can define an expectation of a gradient estimation as, $ \mathbb{E}[\frac{\hat{\partial}}{\partial \theta_{i}} L(\theta_{i})] = \frac{\partial}{\partial \theta_{i}}L(\theta_{i}) = (0.5)^{|\mathcal{B}|}$. Therefore, a variance of the estimated gradient can be defined as,
	\begin{equation} 
	\label{eq:Convergence_variance}
	\begin{split}
	\mathbb{V}\big[\frac{\hat{\partial}}{\partial \theta_{i}} L(\theta_{i})\big] = \mathbb{E}\big[\frac{\hat{\partial}}{\partial \theta_{i}} L^2(\theta_{i})\big] - \left(\mathbb{E}\big[\frac{\hat{\partial}}{\partial \theta_{i}} L(\theta_{i})\big]\right)^2 \\ = \left(0.5\right)^{|\mathcal{B}|} - \left(0.5\right)^{2|\mathcal{B}|}. \;\;\;\;\;\;\;\;\;\;\;\;\;\;\;\;\;\;\;\;\;\;\;\;\;\;
	\end{split}
	\end{equation}
	Now, we can analyze the step of gradient for $P((\hat{\nabla}_{\theta_{i}} L(\theta_{i}), \nabla_{\theta_{i}} L(\theta_{i})) > 0)$ (in \eqref{eq:Convergence_relation_policy_grdient}), where 
	\begin{equation}
	\label{eq:Convergence_final}
	\begin{split}
	P\left(\hat{\nabla}_{\theta_{i}} L(\theta_{i}), \nabla_{\theta_{i}} L(\theta_{i})\right) = \left(0.5\right)^{|\mathcal{B}|} \sum_{\forall i \in \mathcal{B}} \frac{\hat{\partial}}{\partial \theta_{i}} L(\theta_{i}).
	\end{split}
	\end{equation} 
	As a result, $P((\hat{\nabla}_{\theta_{i}} L(\theta_{i}), \nabla_{\theta_{i}} L(\theta_{i})) > 0) = (0.5)^{|\mathcal{B}|}$ implies that the gradient step not only moves in the correct direction but also decreases exponentially with an increasing number of BS agents.     	
\end{proof}

\ifCLASSOPTIONcaptionsoff
  \newpage
\fi



%

\begin{IEEEbiography}[{\includegraphics[width=1in,height=1.25in,clip,keepaspectratio]{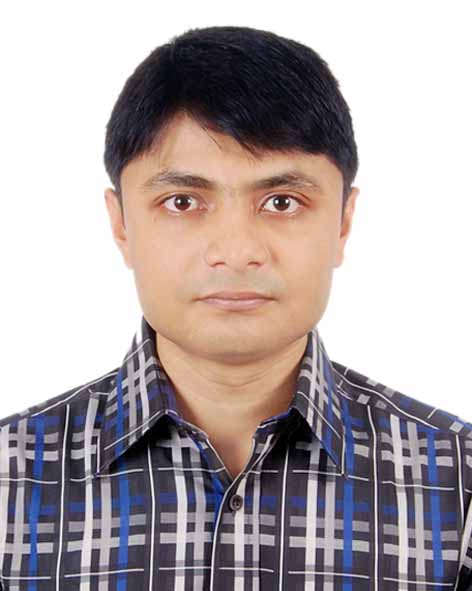}}]{Md.~Shirajum~Munir}
	(Graduate Student Member, IEEE) received the B.S. degree in computer science and engineering from Khulna	University, Khulna, Bangladesh, in 2010. He is currently pursuing the Ph.D. degree in computer science and engineering at Kyung Hee University, Seoul, South Korea. He served as a Lead Engineer with the Solution	Laboratory, Samsung Research and Development Institute, Dhaka, Bangladesh, from 2010 to 2016. His current research interests include IoT network management, fog computing, mobile edge computing, software-defined networking, smart grid, and machine learning.
\end{IEEEbiography}
\begin{IEEEbiography}[{\includegraphics[width=1in,height=1.25in,clip,keepaspectratio]{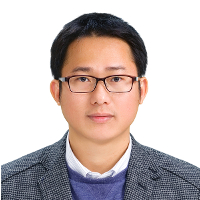}}]{Nguyen~H.~Tran}
	(S'10-M'11-SM'18)  received the	B.S. degree from the Ho Chi Minh City University
	of Technology, Ho Chi Minh City, Vietnam, in 2005, and the Ph.D. degree in electrical and computer engineering from Kyung Hee University, Seoul, South Korea, in 2011. Since 2018, he has been with the School of Computer Science, University of Sydney, Sydney,	NSW, Australia, where he is currently a Senior Lecturer. He was an Assistant Professor with the Department of Computer Science and Engineering, Kyung Hee University, from 2012 to 2017. His current research interests include applying analytic techniques of optimization, game theory, and stochastic modeling to cutting-edge applications, such as cloud and mobile edge computing, data centers, heterogeneous wireless networks, and big data for networks. Dr. Tran was a recipient of the Best KHU Thesis Award in Engineering	in 2011 and the Best Paper Award of IEEE ICC 2016. He has been an Editor of the IEEE TRANSACTIONS ON GREEN COMMUNICATIONS AND	NETWORKING since 2016 and served as the Editor of the 2017 Newsletter of Technical Committee on Cognitive Networks on Internet of Things.
\end{IEEEbiography}

\begin{IEEEbiography}[{\includegraphics[width=1in,height=1.25in,clip,keepaspectratio]{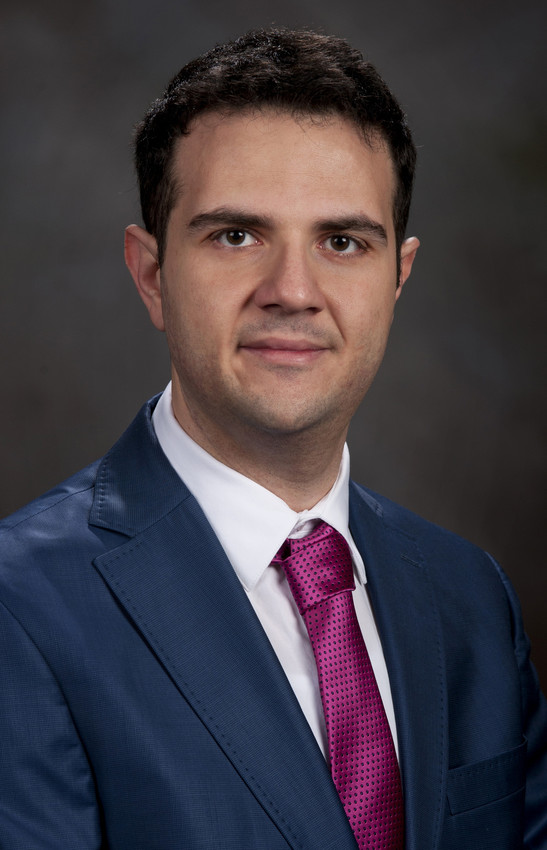}}]{Walid~Saad}
	d (S'07, M'10, SM'15, F'19)
	received his Ph.D degree from the University of Oslo in	2010. He is currently a Professor at the Department	of Electrical and Computer Engineering at Virginia Tech, where he leads the Network sciEnce, Wireless, and Security (NEWS) laboratory. His research interests include wireless networks, machine learning, game theory, security, unmanned aerial vehicles, cyber-physical systems, and network science. Dr. Saad is a Fellow of the IEEE and an IEEE Distinguished Lecturer. He is also the recipient of the NSF CAREER award in 2013, the AFOSR summer faculty fellowship in 2014, and the Young Investigator Award from the Office of Naval Research (ONR) in 2015. He was the author/co-author of eight conference best paper	awards at WiOpt in 2009, ICIMP in 2010, IEEE WCNC in 2012, IEEE PIMRC in 2015, IEEE SmartGridComm in 2015, EuCNC in 2017, IEEE GLOBECOM in 2018, and IFIP NTMS in 2019. He is the recipient of the 2015 Fred W. Ellersick Prize from the IEEE Communications Society, of the 2017 IEEE ComSoc Best Young Professional in Academia award, of the 2018 IEEE ComSoc Radio Communications Committee Early Achievement Award, and of the 2019 IEEE ComSoc Communication Theory Technical Committee. From 2015-2017, Dr. Saad was named the Stephen O. Lane Junior Faculty Fellow at Virginia Tech and, in 2017, he was named College of Engineering Faculty Fellow. He received the Dean's award for Research Excellence from Virginia Tech in 2019. He currently serves as an editor for the IEEE Transactions on Wireless Communications, IEEE Transactions on Mobile Computing, IEEE Transactions on Cognitive Communications and	Networking, and IEEE Transactions on Information Forensics and Security. He is an Editor-at-Large for the IEEE Transactions on Communications.
\end{IEEEbiography}

\begin{IEEEbiography}[{\includegraphics[width=1in,height=1.25in,clip,keepaspectratio]{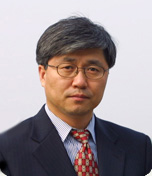}}]{Choong~Seon~Hong}
	(S'95-M'97-SM'11)
	received the B.S. and M.S. degrees in electronic engineering from Kyung Hee University, Seoul, South Korea, in 1983 and 1985, respectively, and the Ph.D. degree from Keio University, Japan, in 1997. In 1988, he joined KT, where he was involved in broadband networks as a Member of Technical Staff. Since 1993, he has been with Keio University. He was with the Telecommunications Network Laboratory, KT, as a Senior Member of Technical Staff and as the Director of the Networking Research Team until 1999. Since 1999, he has been a Professor with the Department of Computer Science and Engineering, Kyung Hee University. His research interests include future Internet, ad hoc networks, network management, and network security. He is a member of the ACM, the IEICE, the IPSJ, the KIISE, the KICS, the KIPS, and the OSIA. Dr. Hong has served as the General Chair, the TPC Chair/Member, or an Organizing Committee Member of international conferences such as NOMS, IM, APNOMS, E2EMON, CCNC, ADSN, ICPP, DIM, WISA, BcN, TINA, SAINT, and ICOIN. He was an Associate Editor of the IEEE TRANSACTIONS ON NETWORK AND SERVICE MANAGEMENT, and the IEEE JOURNAL OF COMMUNICATIONS AND NETWORKS. He currently serves as an Associate Editor of the International Journal of Network Management, and an Associate Technical Editor of the IEEE Communications Magazine.
\end{IEEEbiography}

\end{document}